\documentclass{article}
\usepackage{microtype}
\usepackage{graphicx}
\usepackage{subfigure}
\usepackage{booktabs} 

\usepackage{amsmath,amsfonts,bm}

\def\1{\bm{1}}

\DeclareMathAlphabet{\mathsfit}{\encodingdefault}{\sfdefault}{m}{sl}
\SetMathAlphabet{\mathsfit}{bold}{\encodingdefault}{\sfdefault}{bx}{n}

\usepackage{todonotes}
\usepackage{hyperref}
\usepackage{symbols}
\usepackage{multirow}
\usepackage{makecell}
\usepackage[multiple]{footmisc}

\usepackage{xcolor}

\usepackage{hyperref}
\usepackage[linewidth=1pt]{mdframed}
\usepackage{tikz-dependency}
\NewEnviron{elaboration}{
\par
\begin{tikzpicture}
\node[rectangle,minimum width=\textwidth] (m) {\begin{minipage}{.98\textwidth}\BODY\end{minipage}};
\draw[dashed] (m.south west) rectangle (m.north east);
\end{tikzpicture}
}

\usepackage[accepted]{icml2022}
\newcommand*\colourcheck[1]{%
  \expandafter\newcommand\csname #1check\endcsname{\textcolor{#1}{\ding{52}}}%
}

\usepackage{fdsymbol}
\newcommand{\qbai}{\emph{Q-BabyAI}\xspace}
\newcommand{\qtw}{\emph{Q-TextWorld}\xspace}
\newcommand{\afk}{AFK\xspace}

\newcommand{\oib}{\textbf{Object in Box}\xspace}
\newcommand{\danger}{\textbf{Danger}\xspace}
\newcommand{\gtfav}{\textbf{Go to Favorite}\xspace}
\newcommand{\od}{\textbf{Open Door}\xspace}

\newcommand{\noquery}{No Query\xspace}
\newcommand{\query}{Query Baseline\xspace}
\newcommand{\ours}{AFK\xspace}

\definecolor{green1}{HTML}{0b5400}
\definecolor{orange1}{HTML}{f3905c}
\definecolor{purple1}{HTML}{9258cc}
\definecolor{blue1}{HTML}{027db5}
\definecolor{pink1}{HTML}{ff7a7a}
\definecolor{darkgreen}{HTML}{005e19}
\definecolor{darkblue}{HTML}{240394}
\definecolor{darkred}{HTML}{C00000}
\definecolor{lightblue2}{HTML}{DEEBF7}
\definecolor{lightgreen2}{HTML}{E2F0D9}
\definecolor{lightgray2}{HTML}{767171}

\newcommand{\code}[1]{\texttt{#1}}
\newcommand{\cmd}[1]{\textcolor{darkgreen}{\textbf{\small{\code{#1}}}}}
\newcommand{\token}[1]{\textcolor{darkblue}{\textbf{\small{\code{#1}}}}}
\newcommand{\instruction}[1]{\textit{\code{#1}}}

\newcommand{\sclub}{$\textcolor{blue1}{\clubsuit}$}
\newcommand{\sspade}{$\textcolor{green1}{\spadesuit}$}
\newcommand{\sdiamond}{$\textcolor{orange1}{\vardiamondsuit}$}
\newcommand{\sheart}{$\textcolor{pink1}{\varheartsuit}$}

\icmltitlerunning{Asking for Knowledge (AFK)}

\begin{document}

\twocolumn[
\icmltitle{\vspace{-0.4cm}Asking for Knowledge
\includegraphics[width=0.06\textwidth]{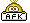}:
\\ Training RL Agents to Query External Knowledge Using Language
}
\hypersetup{pdftitle={Asking for Knowledge: Training RL Agents to Query External Knowledge Using Language}}

\icmlsetsymbol{equal}{*}
\icmlsetsymbol{note}{\dag}

\begin{icmlauthorlist}
\icmlauthor{Iou-Jen Liu}{equal,note,uiuc}
\icmlauthor{Xingdi Yuan}{equal,msr}
\icmlauthor{Marc-Alexandre C\^{o}t\'{e}}{equal,msr}
\icmlauthor{Pierre-Yves Oudeyer}{note,msr,inria}
\icmlauthor{Alexander G. Schwing}{uiuc}
\end{icmlauthorlist}

\icmlaffiliation{uiuc}{University of Illinois at Urbana-Champaign, IL, U.S.A.}
\icmlaffiliation{msr}{Microsoft Research, Montr\'{e}al, Canada}
\icmlaffiliation{inria}{Inria, France}

\icmlcorrespondingauthor{Iou-Jen Liu}{iliu3@illinois.edu}
\icmlcorrespondingauthor{Xingdi Yuan}{eric.yuan@microsoft.com} 

\vskip 0.3in
]

\printAffiliationsAndNotice{\icmlEqualContribution \dag Work partially done while visiting MSR} 


\begin{abstract}
To solve difficult tasks, humans ask questions to acquire knowledge from external sources. In contrast, classical reinforcement learning agents lack such an ability and often resort to exploratory behavior. 
This is exacerbated as few present-day environments support querying for knowledge. 
In order to study how agents can be taught to query external knowledge via language, we first introduce two new environments: the grid-world-based \qbai and the text-based \qtw. In addition to physical interactions, an agent can query an external knowledge source specialized for these environments to gather information. 
Second, we propose the `Asking for Knowledge' (\afk) agent, which learns to generate language commands to query for meaningful knowledge that helps solve the tasks. 
\afk leverages a non-parametric memory, a pointer mechanism and an episodic exploration bonus to tackle (1) irrelevant information, (2) a large query language space, (3) delayed reward for making meaningful queries.
Extensive experiments demonstrate that the \afk agent 
outperforms recent baselines on the challenging \qbai and \qtw environments. The code of the environments and agents are available at \url{https://ioujenliu.github.io/AFK}.
\end{abstract}

\section{Introduction}
\label{sec:intro}

To solve challenging tasks, humans query external knowledge sources, \ie, we ask for help. We also constantly create knowledge sources (\eg, manuals), as it is often more economical in the long term than users exploring via  trial and error. Moreover, cognitive science research~\cite{maratsos2007children, mills2010preschoolers, ronfard2018question} showed that learning to ask questions and to interpret answers is key in a child's development of problem-solving skills. 
Consequently, we hypothesize that autonomous agents can address more complicated tasks if they can successfully learn to query external knowledge sources. For querying, it seems desirable to use some form of language. Not only does this allow to leverage existing knowledge sources built for humans, but  also does it enable us to interpret the queries.


\begin{figure}[t]
\centering
\includegraphics[width=0.5\textwidth]{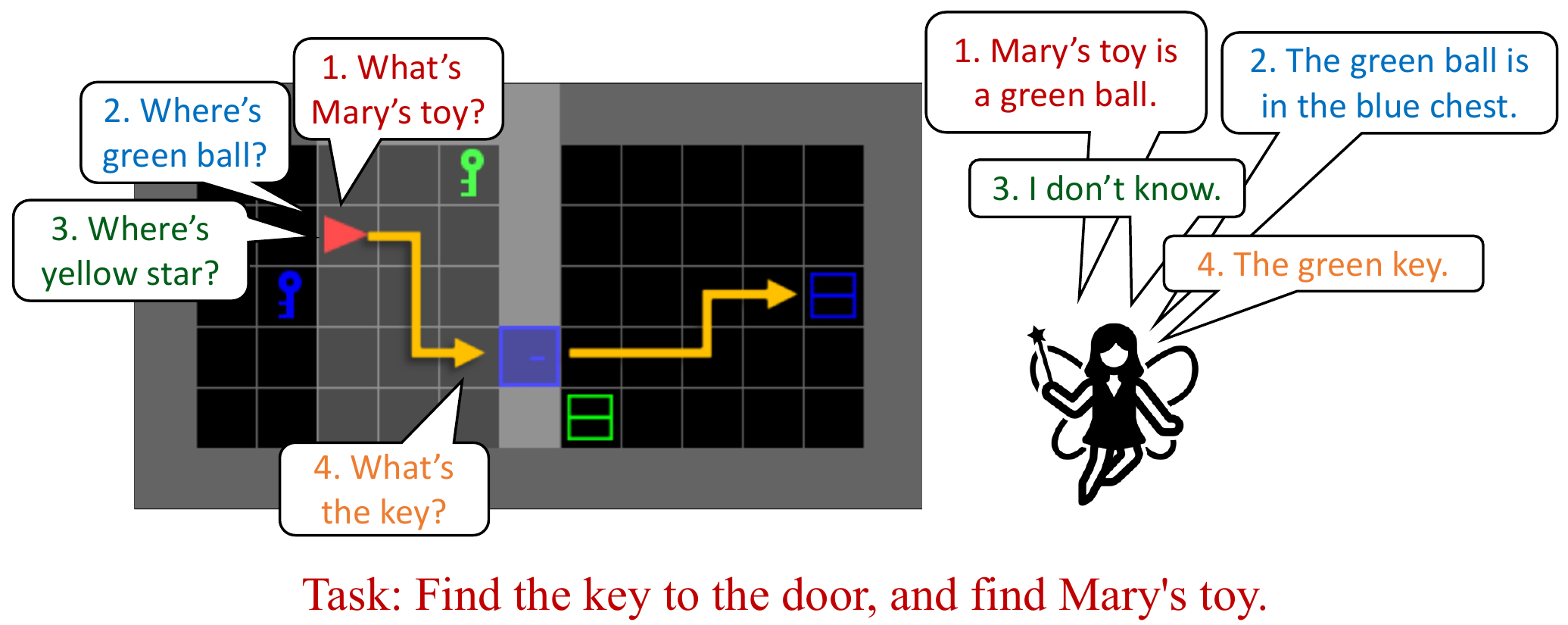}
\caption{Proposed \qbai. Here the agent has to query the knowledge source to succeed.}
\label{fig:teaser}
\end{figure}

\begin{figure*}[t!]
\centering
\includegraphics[width=1.0\textwidth]{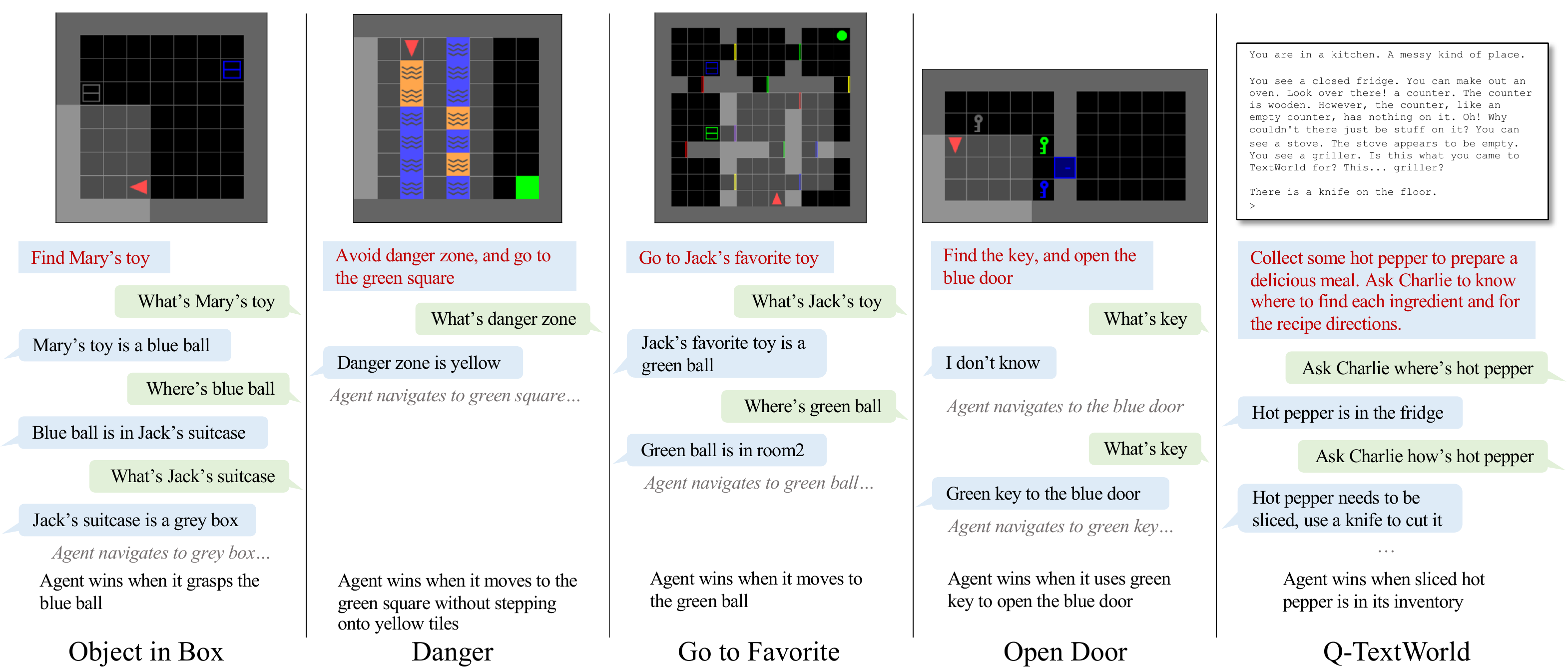}
\caption{Querying interactions in \qbai and \qtw. We  illustrate standard physical interactions as {\color{lightgray2}gray colored} stage directions and highlight the \colorbox{lightgreen2}{questions (green)} and \colorbox{lightblue2}{oracle replies (blue)} upon receiving the {\color{darkred}instruction (red)}.}
\label{fig:tasks}
\end{figure*}

However, the literature to teach agents to query external knowledge sources via language is scarce.
\citet{anna} consider agents that can request help in visual navigation tasks. The agent can issue a `help' signal, and expects the environment to provide an instruction leading it to a place towards the goal. Hence, agents learn when to query, but do not have control of what to query.
\citet{rtfm} show that agents can better address novel tasks when a manual is available. However, the manual contains all relevant information and agents don't need to learn to query. 
\citet{socialai} discuss the open challenge of building agents that learn social interaction skills mixing physical action and language. They show that state-of-the-art deep reinforcement learning systems cannot learn several kinds of social interaction skills. Instead of social skills, we focus on the open challenge of learning to ask for knowledge using language and propose an effective approach.

To deliberately study how agents can be taught to query, we introduce two environments: the grid-world-based \qbai, illustrated in \figref{fig:teaser} and inspired by BabyAI~\cite{babyai}, and the text-based \qtw inspired by TextWorld~\cite{textworld}.
In addition to physical interactions, an agent can use a query language to gather information related to a task. 
Importantly, in \qbai and \qtw, the knowledge source is designed to be task-agnostic, \ie, it replies to all queries, even if irrelevant to the task at hand.
This mimics many real-world knowledge sources, \eg, search engines, which return results based on a user's query, regardless of relevance.  

When training agents to query external knowledge via language, three main challenges arise: 
(1) The action space for generating a language query is large. Even with a template language, the action space grows combinatorially and large action spaces remain a challenge for reinforcement learning~\cite{arnold16,ammanabrolu20}. 
(2) Irrelevant knowledge queried from a task-agnostic knowledge source can confuse agents. As a result, learning to ask meaningful questions is critical. This challenge is in line with the cognitive science finding~\cite{mills2010preschoolers} that children must learn to ask questions that result in answers with useful information.
(3) Rewards for queries are often significantly delayed and sparse. Since the knowledge source  provides specific information rather than a solution, agents have to also understand how to use the acquired information before receiving a significant reward. This mimics the discovery of~\citet{mills2010preschoolers} that children must learn to use the received information.

To address the three challenges, we propose the `asking for knowledge' (AFK) agent. The AFK agent is equipped with a pointer mechanism and a non-parametric memory, which we refer to as a  `notebook,' while using an episodic exploration strategy. The pointer mechanism addresses the challenge of a combinatorially growing action space by restricting the available actions based on the current context. The notebook keeps track of all the information related to the task at hand. The episodic exploration strategy deals with delayed and sparse rewards by issuing an exploration bonus when the agent makes novel and meaningful queries, inspired by information-seeking and epistemic curiosity observed in children~\cite{engel2011children,gottlieb2013information,kidd2015psychology}. 

Comparing this AFK agent to recent baselines on \qbai and \qtw, we observe the AFK agent to ask more meaningful questions and to better leverage the acquired knowledge to solve the tasks.

\section{Queryable Environments}
\label{sec:app_2}

We first  discuss a reinforcement learning (RL) context where agents can query. We then introduce two new environments, \qbai and \qtw, each expanded from prior work \cite{babyai, textworld}.

\subsection{Problem Setting}
\label{sec:problem_setting}

Reinforcement learning considers an agent interacting with an environment and collecting reward over discrete time. The environment is formalized by a partially observable Markov Decision Process (POMDP)~\cite{SuttonRL}. Formally, a POMDP is defined by a tuple $(\cS, {\cA}, \cZ, {\cT}, \cO,  R, \gamma, H)$. $\cS$ is the state space. ${\cal A}$ is the action space. $\cZ$ is the observation space. At each time step $t$, the agent receives an observation $o_t \in \cZ$ following the observation function $\cO: \cS \rightarrow \cZ$ and selects an action $a_t \in \cA$. The transition function $\cT$ maps the action $a_t$ and the current state $s_t$ to a distribution over the next state $s_{t+1}$, \ie, $\cT: \cS \times \cA \rightarrow \Delta(\cS)$. The agent receives a real-valued reward $r_t$ according to a reward function $R: \cS \times \cA \rightarrow {\mathbb R}$. The agent's goal is to maximize the return $\sum_{t=0}^{H} \gamma^t r_t$, where $\gamma$ is the discount factor and $H$ is the horizon. 

In a queryable environment, in addition to observations representing its surrounding, an agent also receives a response from the knowledge source upon issuing a query.
Formally, the observation space $\cZ = \cZ_{\text{env}} \times \cZ_{q}$ is composed of $\cZ_{\text{env}}$ and $\cZ_{q}$, representing the agent's surrounding and the reponse to a query, respectively. 

Similarly, at each step, the agent's action space   $\cA = \cA_{\text{phy}} \cup \cA_{q}$ is composed of the physical action space $\cA_{\text{phy}}$ supported by classical RL environments (\eg, navigational actions, toggle, grasp) and the query action space $\cA_{q}$. 

\noindent\textbf{Response Space $\cZ_{q}$ and Query Action Space $\cA_{q}$:}
As a controllable starting point for this research, we equip the environments with a queryable oracle knowledge source.
Specifically, whenever receiving a sequence of tokens as a query, the oracle replies with a sequence of tokens. 
To consider the compositionality of language while reducing the burden of precise natural language generation, we define a template format for queries and responses. 
This design is also compatible with our plan of extending the knowledge source to more natural forms like databases.

A query is defined as a 3-tuple of \cmd{<func, adj, noun>}. 
In this 3-tuple, \cmd{func} is a function word selected from words like \token{where's}, \token{what's} and \token{how's}, which indicates the function of a query (\eg, inquire about an object's location or affordances).
The combination of an adjective (\cmd{adj}) and a \cmd{noun} enables to refer to a unique object within the environment.

Given a query, the oracle replies with a sequence of tokens. 
For this, the oracle has access to a set of  ``knowledge facts'' associated with a particular instantiation of the environment. The knowledge facts are key-value pairs, where keys are the aforementioned 3-tuple of \cmd{<func, adj, noun>} and values are sequences of tokens.
If a given query matches a key in the set of knowledge facts, the oracle will return the corresponding value.
Otherwise, the oracle returns the message \token{I don't know}.

Crucially, the set of knowledge facts is much larger than necessary and irrelevant information is, by design, accessible to the agent. For instance, when tasked to find Mary's toy, information about Tim and Tim's toy is also available if queried. 
Gathering irrelevant information may lead to confusion and subsequent sub-optimal decisions.
Moreover, some tasks require multi-hop information gathering (\eg, \oib), in which the agent must ask follow-up questions to get all information needed to solve it.

\noindent\textbf{Information Sufficiency:}
Practically, agents that can query have two main advantages.
First, for environments containing sufficient information to be solved via exhaustive exploration, querying can provide a more natural and effective way to gather information (\eg, reducing the policy length).
Second, for environments that only provide partial information (\eg, an agent must recognize and avoid danger tiles by trial-and-error, but danger tiles are randomly assigned per episode), only querying will lead to successful completion of the tasks.

To study both advantages, we augment  BabyAI~\cite{babyai} and TextWorld~\cite{textworld} with a queryable knowledge source.
We design tasks where the environment contains sufficient information, but we add knowledge facts which can help the agent to reduce exploration if used adequately. 
In addition, we design other tasks where agents can only succeed when they are able to query.
We provide details next.


\begin{figure*}[t]
\centering
\includegraphics[width=0.95\textwidth]{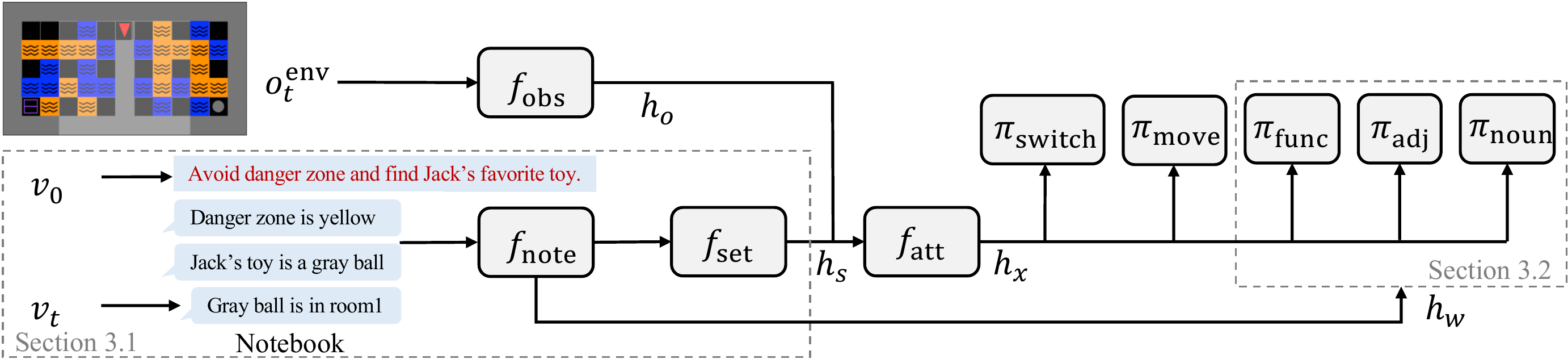}
\caption{An overview of the \afk agent. An embedding of the notebook ($h_s$; see \secref{subsec:ks}) and the environment ($h_o$) is combined ($h_x$) for use in five policy functions. The policies for query generation make use of notebook information ($h_w$) (see \secref{subsec:pointer}).}
\label{fig:pipeline}
\end{figure*}

\subsection{\qbai}
\label{subsec:q_babyai}

We first introduce \qbai, an extension of the BabyAI environment~\cite{babyai}.
We devise four level 1 tasks, namely \oib, \danger, \gtfav and \od. 
In Fig.~\ref{fig:tasks}, we provide examples of agents querying the knowledge source for each of the level 1 tasks after receiving the goal instruction for that episode.
The four tasks permit to study the two advantages mentioned above. 

Specifically, both the \oib and \danger tasks can only be solved 100\% of the time when querying is used to reveal the necessary knowledge --- opening the wrong box or stepping on the danger tile terminates the game.
In contrast, for the \gtfav and \od tasks, an agent can exhaust the environment to accomplish the goals.
However, querying the knowledge source can greatly boost the agent's efficiency in both tasks.
To prevent agents from memorizing solutions (\eg, Mary's toy is in the red box), we randomly place objects and tiles in the environment, as well as shuffle the entity names and colors in every episode.

For the \oib and \danger tasks, we use a single-room setting to separate the difficulties of navigation and querying. In the \gtfav and \od tasks, we use a multi-room setting. 
It is worth noting that in the \od task,
only querying at specific locations (\ie, next to doors) can result in meaningful answers.

Having the four level 1 tasks defined, we increase the difficulty by composing them into more challenging higher-level tasks.
For level $k$ tasks, we combine $k$ different tasks selected from the four level 1 tasks. 
As a result, we have six level 2 tasks, four level 3 tasks and one level 4 task. 
As an example, \od + \oib is a level 2 task where the instruction could be \instruction{Find the key to the door, and find Mary's toy}. 
To solve the task, an agent must figure out what is Mary's toy, where it is, and which key opens the locked door. 
We provide full details of the \qbai tasks including statistics in Appendix~\ref{appd:qbai_detail}.

\subsection{\qtw}
\label{subsec:q_textworld}
We also develop \qtw, augmenting the TextWorld environment~\cite{textworld} with a queryable knowledge source. 
Given a few configuration parameters, \qtw can generate new text-based games on the fly. 
Those interactive text environments are POMDPs with text-only observations and action space. Agents interact with the game by issuing short text phrases as actions, then get textual feedback. 
Text-based games provide a different view from their vision- or grid-based counterparts that can make things harder: 1) states are represented by highly abstracted signals requiring language understanding to interpret correctly; 2) the action space is combinatorially large due to compositionality of language; 3) levels of verbosity, \ie, amount of irrelevant text, can potentially confuse an agent.

In this work, we adopt the cooking themed setting from prior work~\cite{adhikari2020gata}. An example is shown in Fig.~\ref{fig:tasks}.
In all games, an agent must gather cooking ingredients, which are placed randomly in the room, either visible to the agent, or hidden inside some containers that need to be opened first.
In a more difficult setting, each ingredient also needs to be cut in a specific way (\ie, \token{chopped}, \token{sliced}, or \token{diced}). The agent must query the knowledge source to obtain that information and then act accordingly by issuing the right action while holding both the ingredient and a knife. We provide full details of the \qtw tasks including statistics in Appendix~\ref{appd:qtw_detail}.

Being consistent with the \qbai tasks, we study both advantages of having a querying behavior --- improve efficiency and acquire necessary information. 
In games where cutting is not involved, agents can rely on exhaustive exploration to gather all portable objects and win. However, knowing what and where the required ingredients are can  improve efficiency significantly. 
In contrast, in games where the ingredients need to be cut, an incorrect operation on the ingredient terminates the game. Hence, querying the recipe is the only way to perform above random.

\section{Asking for Knowledge (AFK) Agent}
\label{sec:app}

In this section, we first present an overview of the `Asking for Knowledge' (AFK) agent before we discuss details.

\noindent\textbf{Overview:} 
As illustrated in \figref{fig:pipeline}, 
the goal of the agent is to solve a task specified by an instruction. The language-based instruction (sequence of tokens) $v_0$ is provided at the start of each episode. At each time step $t$, the agent receives an environment observation $o_t^{\text{env}} \in \cZ_{\text{env}}$. 
Moreover, if the agent issued a query at time step $t - 1$, it also receives a language response $v_t \in \cZ_q$ from the oracle, otherwise $v_t = \emptyset$. 

To reduce the amount of noisy information (\ie, $v_t$ unrelated to the task at hand), we develop a non-parametric memory for gathered information, which we refer to as a `notebook.' The notebook is a collection of sets where related information are being combined into a single set. The AFK agent only looks at the set that contains the task instruction $v_0$, which determines relevance to the task. Upon processing the notebook we obtain a representation $h_s$ (details in \secref{subsec:ks}).

We combine the environment observation $o_t^{\text{env}}$ and the notebook representation $h_s$ relevant for the task via an aggregator module~\cite{film,transformer}. 
Given the output of the aggregator, $h_x\in{\mathbb R}^l$, where $l$ is the encoding size, we use five heads to generate the physical actions and the language query actions. Specifically, we use a switch head $\pi_{\text{switch}}(\cdot|h_x): {\mathbb R}^l \rightarrow \Delta(\{0, 1\})$, a physical action head $\pi_{\text{phy}}(\cdot|h_x): {\mathbb R}^l \rightarrow \Delta(\cal A_{\text{phy}})$, a function word head: $\pi_{\text{func}}(\cdot|h_x): {\mathbb R}^l \rightarrow \Delta(V_{\text{func}})$, an adjective head $\pi_{\text{adj}}(\cdot|h_x): {\mathbb R}^l \rightarrow \Delta(V_{\text{adj}})$, and a noun head $\pi_{\text{noun}}(\cdot|h_x): {\mathbb R}^l \rightarrow \Delta(V_{\text{noun}})$.
Here, $\Delta(X)$ represents a distribution with support $X$, $\cal A_{\text{phy}}$ is the physical action space, and $V_{\text{func}}, V_{\text{adj}}, V_{\text{noun}}$ are the function word, adjective, and noun vocabulary spaces.

The switch head decides whether the agent executes a physical action or issues a query. Conceptually, the agent first samples a value $z$ according to $\pi_{\text{switch}}(\cdot|h_x)$. If $z = 0$, the agent will sample a physical action $a_{\text{phy}}$ from $\pi_{\text{phy}}(\cdot|h_x)$ which is subsequently executed. 
In contrast, if $z = 1$, the agent will issue the query $[w_{\text{func}}, w_{\text{adj}}, w_{\text{noun}}]$ by sampling a function word $w_{\text{func}}$, an adjective $w_{\text{adj}}$, and a noun $w_{\text{noun}}$ independently from $\pi_{\text{func}}(\cdot|h_x)$, $\pi_{\text{adj}}(\cdot|h_x)$, and  $\pi_{\text{noun}}(\cdot|h_x)$. 
Note that the query action space $\cA_{q}$ is large, which makes RL training  challenging.
We provide detailed statistics in Appendix~\ref{appd:env_task_detail}.

To alleviate issues due to the large action space, we adopt a pointer network~\cite{pointer} to generate queries. Specifically, the pointer network is restricted to `point' to words occurring in the notebook.  
This ensures that the generated query uses  words that are related to the already gathered information (details in \secref{subsec:pointer}).

In addition, to deal with delayed and sparse rewards, we propose an episodic exploration  method which further incentivizes an agent to ask questions that are related to the task at hand (details in \secref{subsec:exploration}).

\subsection{Notebook}
\label{subsec:ks}
In the following, we discuss the notebook's construction and describe the computation of the encoding $h_s$.

\noindent\textbf{Notebook construction:} 
Let $F$ (for facts) denote the notebook, which is a non-parametric memory. 
Formally, $F$ is a set of disjoint sets, \ie, $F = \{A_i\}_{i=0}^{|F|-1}$ and $A_i \cap A_j = \emptyset, \forall i \neq j$. 
For each set $A_i$, each element $v \in A_i$ represents either a response from the oracle or the task instruction. 

At the beginning of each episode, the notebook $F$ is initialized with a singleton $A_0 = \{v_0\}$ that contains the task instruction $v_0$, \ie, $F=\{\{v_{0}\}\}$. 
When an agent receives a new response $v_i \neq \emptyset$, we first find all sets that contain information related to $v_i$ in the notebook.  Formally, we construct an index set $S$ that consists of the indices of related sets, \ie, $S=\{j|\exists v \in A_j~\text{s.t.}~\text{Sim}(v_i, v) \ge \alpha\}$, where  $\text{Sim}(u, v) \in [0, 1]$ is a similarity function and $\alpha \in [0, 1]$ is a threshold. We study both the uni-gram and bi-gram similarity~\cite{ngram}. 
If $S$ is not empty, we combine all the related sets and the new response $v_i$ to obtain a new set $A_k$. Formally, $A_{k} =  \bigcup_{j \in S} A_j \cup \{v_i\} $, where $k = \min_{j \in S} j$. Then, all sets $\{A_j\}_{j \in S}$ are replaced with the new set $A_k$. 
If the index set $S$ is empty, we add $A_{k} = \{v_i\}$ to $F$, where $k$ is the next available index. Importantly, note that the task instruction $v_0$ is always part of the set $A_0$.

\noindent\textbf{Notebook encoding:} To discard noisy information coming from responses unrelated to the task at hand, the AFK agent only considers the set $A_0$ which contains the task instruction $v_0$. 
We use a recurrent neural network $f_\text{note}$~\cite{cho2014gru} to encode each `note' in $A_0$, \ie, for each $v_i \in A_0$ ($i\in\{1, \dots, |A_0|\}$), we have $h_i = f_{\text{note}}(v_i) \in \mathbb{R}^{|v_i| \times l}$, where $|v_i|$ is the number of words in $v_i$ and $l$ is the hidden dimension. 
To further encode 
the instruction related notes (\ie, $A_0$) as a whole,
we use a Deep Set model $f_{\text{set}}$~\cite{zaheer2017deep}, 
\ie, $h_s = f_{\text{set}}([h_1, \dots, h_{|A_0|}]) \in \mathbb{R}^{l}$, where $h_s$ is the resulting  encoding. 
In addition, the input observation $o^{\text{env}}$ is encoded via a neural network $f_{\text{obs}}$, \ie, $h_o = f_{\text{obs}}(o^{\text{env}}) \in \mathbb{R}^{l}$, where $h_o$ is the resulting observation encoding. 
{An aggregator module is used to combine $h_o$ and $h_s$, 
\ie, $h_{x} = f_{\text{att}}(h_o, h_s) \in \mathbb{R}^{l} $.}

\subsection{Pointer Mechanism for Language Generation}
\label{subsec:pointer}

To address the challenge of a combinatorially growing action space, we develop a pointer mechanism for the policies $\pi_{\text{adj}}$ and $\pi_{\text{noun}}$. 
Concretely, the pointer mechanism restricts the AFK agent queries to use only the words appearing in the set $A_0$.

We achieve this by first applying a mask 
before computing the policy distributions $\pi_{\text{adj}}$ and $\pi_{\text{noun}}$, \ie, $\pi_{\text{adj}}$ and $\pi_{\text{noun}}$ only have non-zero probability for adjectives and nouns in the instruction related set of notes $A_0$. 
We use the generation process of the noun as an example.
Let $m_{\text{noun}}$ denote the number of nouns in $A_0$, and let $h_w \in \mathbb{R}^{m_{\text{noun}} \times l}$ denote the word encodings of all nouns in $A_0$. Using attention queries $q \in \mathbb{R}^{l}$ and keys $k \in \mathbb{R}^{m_{\text{noun}} \times l}$ such that 
    \begin{equation}
        q = h_x \cdot W_q,
\quad\text{and}\quad
        k = h_w \cdot W_k,
    \end{equation}
    with learnable parameters $W_q, W_k \in \mathbb{R}^{l \times l}$, we compute the attention $e_{\text{noun}}$ over all nouns in $A_0$ as
    \begin{equation}
        e_{\text{noun}} = \text{softmax}(q \cdot k^T )\in \mathbb{R}^{m_{\text{noun}}}.
    \end{equation}
A distribution over the noun vocabulary, \ie,  $V_{\text{noun}}$, is then constructed from $e_{\text{noun}}$. Specifically, for each word $w \in V_{\text{noun}}$, we have $\pi_{\text{noun}}(w) = \sum_{i=1}^{m_{\text{noun}}}e^i_{\text{noun}}\text{I}[d(i) = w]$, where $d(i)$ maps the index $i$ to the corresponding word in $A_0$ and $e^i_{\text{noun}}$ represents the $i$-th element of $e_{\text{noun}}$. $\text{I}$ is the indicator function. The pointer mechanism for $\pi_{\text{adj}}$ is constructed similarly. We defer  details to Appendix~\ref{appd:implementation_details}.

\subsection{Episodic Exploration}
\label{subsec:exploration}

To deal with delayed and sparse rewards, inspired by~\citet{savinov2018episodic}, we develop an episodic exploration mechanism to encourage the agent to ask questions related to the task at hand.
    
At each time step, the agent receives reward $r = r^{\text{env}} + b$, where $r^{\text{env}}$ is the external reward and $b$ is the bonus reward. A positive bonus reward $b$ is obtained whenever a query's response $v_i\neq\emptyset$ expands the agent's knowledge about the task, \ie, $A_0$. The reward is only given for new information.
Formally, 
\begin{equation}
b = \beta(\text{I}[(v_i \in A_0) \land (v_i \notin A'_0)]),
\end{equation}
where $v_i$ denotes a new response returned by the oracle, and $A'_0$ denotes the set from the previous game step containing the task instruction $v_0$. $\beta > 0$ is a scaling factor and $\text{I}$ is the indicator function.

\section{Experimental Results}
\label{sec:experimental}


\begin{table}[t] 

\begin{center}
{\small
\begin{tabular}{c|c|cccccc}
\specialrule{.15em}{.05em}{.05em} 
                      & Tasks & \multicolumn{1}{c}{\makecell{No Query}} & \multicolumn{1}{c}{\makecell{Query \\ Baseline} }  & \multicolumn{1}{c}{\makecell{AFK (Ours)} } \\ \toprule\toprule

\multirow{4}{*}{Lv. 1}     
& \sclub &   50.5$\pm$2.0      	&49.8$\pm$1.2 	 &	\bf{100.0$\pm$0.0}            \\
& \sspade &  68.3$\pm$2.4     	&73.8$\pm$1.2  &	\bf{100.0$\pm$0.0}      \\
& \sdiamond &	98.9$\pm$0.8  	& 99.3$\pm$0.3 	 &	\bf{100.0$\pm$0.0}      \\
& \sheart &	99.7$\pm$0.3 	& 85.3$\pm$22.3 	 &	\bf{100.0$\pm$0.0}        \\ 
 \midrule
 \multirow{6}{*}{Lv. 2}     
& \sclub \sspade&         0.0$\pm$0.0  		& 0.0$\pm$0.0  &	\bf{90.3$\pm$1.8}          \\
& \sclub \sdiamond&         0.1$\pm$0.1  	& 0.6$\pm$0.5 	 &	\bf{94.3$\pm$2.3}       \\
& \sclub \sheart&	0.0$\pm$0.0  	& 0.0$\pm$0.0 	 &	\bf{99.0$\pm$0.4}     \\
 
& \sspade \sdiamond &0.4$\pm$0.1&0.2$\pm$0.2&\bf{100.0$\pm$0.0} \\ 
& \sspade \sheart&	 0.0$\pm$0.0    	&  0.0$\pm$0.0   &	{ 0.0$\pm$0.0 }      \\
& \sdiamond \sheart&	84.1$\pm$0.3   	& 94.0$\pm$3.3 	 &	\bf{98.7$\pm$0.2}           \\ 
 
 \midrule
 \multirow{4}{*}{Lv. 3}     
& \sclub \sspade \sdiamond&         0.0$\pm$0.0 	& 0.0$\pm$0.0 	 &	{0.15$\pm$0.2}           \\
& \sclub \sspade \sheart &         0.0$\pm$0.0  	& 0.0$\pm$0.0 	 &	{0.0$\pm$0.0}       \\
& \sclub \sdiamond \sheart &	0.0$\pm$0.0  	& 0.0$\pm$0.0 	 &	{2.1$\pm$0.8}      \\
& \sspade \sdiamond \sheart &	4.3$\pm$1.0  	& 4.4$\pm$0.8 	 &	{4.8$\pm$0.9}          \\ 
 \midrule
 \multirow{1}{*}{Lv. 4}     
& \sclub \sspade \sdiamond \sheart&          0.0$\pm$0.0   	 &  0.0$\pm$0.0    &	{ 0.0$\pm$0.1 }            \\

\end{tabular}
}
\caption{Success rate ($\%$) on \qbai. \sclub: \oib, \sspade: \danger, \sdiamond: \gtfav, \sheart: \od. 
}
\label{tb:main_new}
\end{center}
\end{table}

\begin{table}[t] 

\begin{center}
{\small
\begin{tabular}{c|cccccc}
\specialrule{.15em}{.05em}{.05em} 
                      Tasks & \multicolumn{1}{c}{\makecell{No Query}} & \multicolumn{1}{c}{\makecell{Query Baseline} }  & \multicolumn{1}{c}{\makecell{AFK (Ours)} } \\ \toprule\toprule

\sdiamond &   30.2$\pm$1.5      	&26.7$\pm$1.5 	 &	\bf{16.8$\pm$6.7}            \\
\sheart &  26.2$\pm$0.9     	&36.8$\pm$1.0  &	\bf{20.6$\pm$0.2}      \\

\end{tabular}
}
\caption{Number of steps required to solve a task. \sdiamond: \gtfav, \sheart: \od. 
}
\label{tb:len}
\end{center}
\end{table}




\begin{table}[t] 

\begin{center}
{\small
\begin{tabular}{c|ccc}
\specialrule{.15em}{.05em}{.05em} 
Task  & No Query & Query Baseline & AFK (Ours)\\
\toprule\toprule
Take 1      & 75.1$\pm$4.1  & 73.5$\pm$5.8  & \textbf{85.1$\pm$2.9}  \\
Take 2      & 24.0$\pm$6.6  & 13.7$\pm$8.5  & \textbf{61.9$\pm$6.5}  \\
Take 1 Cut  & 24.6$\pm$1.0  & 22.9$\pm$3.6  & \textbf{43.5$\pm$15.9}  \\
Take 2 Cut  & 0.0$\pm$0.0  & 0.0$\pm$0.0  & 0.0$\pm$0.0  \\

\end{tabular}
}
\caption{ Success rate ($\%$) on \qtw.}
\label{tb:tw}
\end{center}
\end{table}

\begin{table}[t] 
\centering
\begin{center}
{\small
\begin{tabular}{c|cccccccc}
\specialrule{.15em}{.05em}{.05em} 
                      Task  & \multicolumn{1}{c}{\makecell{AFK w/o \\ Notebook}} & \multicolumn{1}{c}{\makecell{AFK w/o \\ Pointer\\ Mechanism} } &  
                      \multicolumn{1}{c}{\makecell{AFK w/o \\ Episodic\\ Exploration} }  &  
                      \multicolumn{1}{c}{\makecell{AFK \\(Ours)} }  
                       \\ \toprule\toprule

\makecell{\sclub}    &   50.0$\pm$0.8  	& 49.4$\pm$0.7    &	49.8$\pm$0.7  & \bf{100.0$\pm$0.0}    \\
\makecell{\sspade}    &   99.1$\pm$0.2  	& 100.0$\pm$0.0   &	93.8$\pm$0.7  & \bf{100.0$\pm$0.0}          \\
\makecell{\sdiamond}    &   99.2$\pm$0.4  	& 99.7$\pm$0.2   &	99.3$\pm$0.2 & \bf{100.0$\pm$0.0}           \\
\makecell{\sheart}    &   85.1$\pm$1.0  	& 100.0$\pm$0.0    &	77.8$\pm$0.7  & \bf{100.0$\pm$0.0}           \\
\makecell{\sclub\sheart}    & 48.5$\pm$1.9  	& 90.5$\pm$1.4    &	50.0$\pm$1.8  & \bf{99.0$\pm$0.4}        \\
\bottomrule
\makecell{Mean}    &  76.4   & 87.9 &  74.1 &   99.8           \\

\end{tabular}
}
\caption{Ablation Study. Success rate ($\%$) on \qbai. \sclub: \oib, \sspade: \danger, \sdiamond: \gtfav, \sheart: \od.}
\label{tb:abl}
\end{center}
\end{table}

\begin{table}[t] 

\begin{center}
{\small
\begin{tabular}{c|c|cc}
\specialrule{.15em}{.05em}{.05em} 
                      		   \multirow{1}{*}{\makecell{Target  Task}}   		         & \multirow{1}{*}{\makecell{Source Tasks}}          &   Succ. ($\%$)   &    Eps. Len.                      \\ \hline

\sclub\sspade  &   \makecell{\sspade\sdiamond$+$\sdiamond\sclub}      	& 22.1$\pm$0.7 & 32.0$\pm$0.3              \\ \midrule
  \sspade\sdiamond    & \makecell{\sspade\sclub$+$\sdiamond\sclub} &     35.6$\pm$2.1 & 71.6$\pm$1.1                      \\

\end{tabular}
}
\caption{Zero-shot generalization study of \afk. 
}
\label{tb:gen}
\end{center}
\end{table}

\begin{table}[t] 

\begin{center}
{\small
\vspace{-0.1cm}
\begin{tabular}{c|c|ccc}
\specialrule{.15em}{.05em}{.05em} 
                           	
                      		   Task        & \makecell{$|Q_t|$}          &   Precision  &    Recall       &     F1         \\ \hline

\sclub\sheart  &   5      	& 0.804 & 0.823 & 0.812              \\ 
   \sdiamond\sheart    & 4 &     0.771 & 0.560 & 0.601                \\
  \sspade\sdiamond    & 3 &     0.989 &0.989    & 0.989                  \\

\end{tabular}
}
\caption{Query quality of \afk. \sclub: \oib, \sspade: \danger, \sdiamond: \gtfav, \sheart: \od.}
\label{tb:q_quality}
\end{center}
\end{table}

\begin{figure*}[t]
\centering
\includegraphics[width=\textwidth]{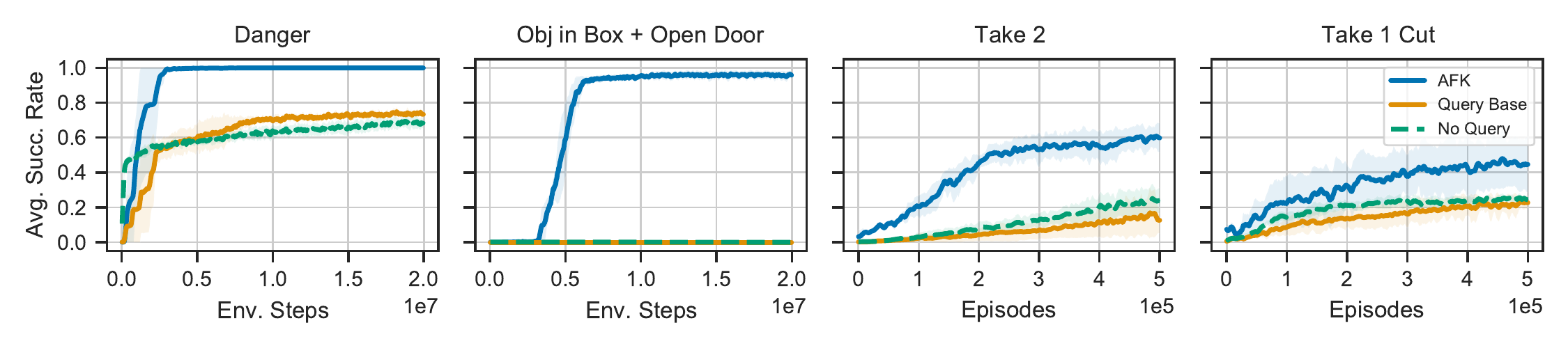}
\vspace{-0.7cm}
\caption{Training curves of \afk, non-query baseline, query baseline on \qbai (left) and \qtw (right).}
\label{fig:plt_main}
\end{figure*}

In this section, we
present  the experimental setup, evaluation protocol, and our results on \qbai and \qtw.

\textbf{Experimental Setup:}
We adopt the publicly available BabyAI and TextWorld code released by the authors\footnote{BabyAI: \href{https://github.com/mila-iqia/babyai}{github:mila-iqia/babyai}}\footnote{TextWorld: \href{https://github.com/xingdi-eric-yuan/qait_public}{github:xingdi-eric-yuan/qait\_public}} as our non-query baseline system, denoted as \textbf{\noquery}. 
We consider a vanilla query agent~\cite{socialai} (\textbf{\query}), in which query heads are added to the baseline agent to generate language queries.
We refer to the proposed agent via \textbf{\ours}, which is the  agent with 1) notebook, 2) pointer mechanism, and 3) episodic exploration.

We follow the original training protocols used in BabyAI and TextWorld.
Specifically, we train all agents in \qbai environments with proximal policy optimization  (PPO)~\cite{ppo} for $20$M - $50$M environment steps, depending on the tasks' difficulty. 
For \qtw agents, we use the Deep Q-Network~\cite{mnih2013playing,hessel18rainbow} and the agents are trained for $500$K episodes. 
We provide implementation details in Appendix~\ref{appd:implementation_details}.

\textbf{Evaluation Protocol:} In \qbai, the policy is evaluated in an independent evaluation environment every $50$ model updates and each evaluation consists of $500$ evaluation episodes. To ensure a fair and rigorous evaluation, we follow the evaluation protocols suggested by \citet{Henderson17, Colas18} and report the `final metric'. The final metric is the average evaluation success rate of the last ten models in the training process, \ie, average success rate of the last $5000$ evaluation episodes. 
In \qtw, we report the final running average training scores with a window size of $1000$. 
Note, in each episode, entities are randomly spawned preventing agents from memorizing training games. 
All experiments are repeated  five times with different random seeds.

\textbf{\qbai Results:}
We first compare our \afk agent with baselines on all level 1 and level 2 tasks of \qbai. The final metrics and standard deviation of average evaluation success rate are reported in~\tabref{tb:main_new}. As shown in~\tabref{tb:main_new}, for level 1 and level 2 tasks, the \afk agent achieves significantly higher success rates than the baselines, particularly in  \oib (\sclub) and \danger (\sspade) where information has to be queried. 
This demonstrates that the \afk agent asks more meaningful questions and successfully leverages oracle replies to solve tasks. 
We provide extra analysis to support this in a later subsection.
In addition, we observe that in tasks where the instruction provides sufficient information, \eg, \gtfav (\sdiamond) and \od (\sheart), all agents are able to solve the tasks. However, as shown in~\tabref{tb:len}, an \afk agent needs fewer steps to solve the tasks. 
This suggests that meaningful queries can result in better efficiency. Training curves are shown in~\figref{fig:plt_main}. See Appendix~\ref{appd:curves} for training curves and results of all experiments.

To show the limitation of the proposed approach, we run experiments on the very challenging level 3 and level 4 tasks of \qbai. As shown in~\tabref{tb:main_new}, \afk as well as all baselines fail to solve the level 3 and level 4 tasks due to the tasks' high complexity and very sparse rewards. 
This shows that training RL agents to query in language is still a very challenging and open problem which needs more attention from our community.

\textbf{\qtw Results:}
We compare \afk with the baseline agents on \qtw in~\tabref{tb:tw}.
Specifically, we conduct experiments on four settings with  gradually increasing difficulty.
Here, `Take $k$'  denotes that an agent  needs to collect $k$ food ingredients, which may  spawn in containers and are hence invisible to the agent before the container is opened.
`Cut' indicates that the collected food ingredients need to be cut in  specific ways, for which the recipe needs to be queried.
As shown in~\tabref{tb:tw}, \afk significantly outperform the baselines on three of the tasks.
Analogously to \qbai experiments, the \noquery agent sometimes outperforms the \query agent.
We believe this is caused by 1) the larger action space of the \query compared to \noquery, and 2) a missing mechanism helping the agent to benefit from queries --- together they reduce the \query's chance to experience meaningful trajectories.
We observe that none of the agents can get non-zero scores on the Take 2 Cut task. We investigate the agents' training reward curves (\figref{fig:tw_reward},  Appendix~\ref{appd:curves}): while the baselines get 0 reward, \afk actually learns to obtain higher reward.
We suspect that due to the richer entity presence in \qtw, and the resulting larger  number of valid questions (connected to $A_0$), \afk may exploit the exploration bonus and ask more questions than necessary.
This suggests that better reward assignment methods are needed for agents to perform in more complex environments.

\textbf{Ablation Study:} 
We perform an ablation study to examine the effectiveness of the proposed 1) notebook, 2) pointer mechanism, and 3) episodic exploration. 
For this, we use various level 1 and level 2 \qbai tasks. The results are reported in~\tabref{tb:abl}. As shown in~\tabref{tb:abl}, removing the notebook, the pointer mechanism, or the episodic exploration results in the success rate dropping by {$22.9\%$, $11.4\%$, and $25.2\%$ on average}. 
This demonstrates that all three proposed components are essential for an \afk agent to successfully generate meaningful queries and solve tasks.  

\textbf{Generalization:} To assess an \afk agent's capability of making meaningful queries and solving different, novel, unseen tasks, we perform a generalization study. Specifically, we train \afk agents on a set of level 2 source tasks. Then the trained \afk agent is tested on an unseen level 2 target task (new combination of sub-tasks used in training). The results are summarized in~\tabref{tb:gen}. As shown in~\tabref{tb:gen}, upon training on source tasks, an \afk agent achieves $22.1\%$ and $35.6\%$ success rate on the level 2 target tasks `\oib+ \danger'(\sclub\sspade) and `\danger+ \gtfav'(\sspade\sdiamond), which the agent has never seen during training. In contrast, the \query only achieves a  $0.0\%$ and $0.2\%$ success rate on the target tasks after training $20$M steps directly on the target tasks (\tabref{tb:main_new}).

\textbf{Query Quality:} 
To gain more insights, we study the quality of the queries issued by an agent. Each episode of our tasks is associated with a set of queries $Q_t$ which are useful for solving the task. If an agent issues a query $q \in Q_t$, the query is considered `good.' We refer to the number of good queries (not counting duplicates) and total number of queries (counting duplicates) generated by the agent in one episode as $n_{g}$ and $n_{\text{tot}}$. We report the average precision, recall, and F1 score~\cite{f_score} of the generated queries over $200$ evaluation episodes. Specifically, $\text{precision} = \frac{n_g}{n_{\text{tot}}}$, $\text{recall} = \frac{n_g}{|Q_t|}$, and F1 score is the harmonic mean of precision and recall.  
As shown in~\tabref{tb:q_quality}, the \afk agent achieves high F1 scores across various tasks. In contrast, the \query converges to a policy that does not issue any query and thus has zero precision and recall in all tasks. This demonstrates \afk's ability to learn to ask relevant questions.

\section{Related Work}
\label{sec:related}
\noindent{\bf Information Seeking Agents:}
In recent years a host of works discussed building of information seeking agents.
\citet{anna} propose to leverage an oracle in 3D navigation environments. 
The oracle is activated in response to a special signal from the agent and provides a language instruction describing a subtask the agent could follow.
\citet{socialai} design grid-world tasks similar to ours, but focus on the social interaction perspective.
For instance, some agents are required to emulate their social peers' behavior to successfully communicate with them.
\citet{potash19} propose a game setting which requires agents to ask sequences of questions efficiently to guess the target sentence from a set of candidates.  
\citet{yuan2021interactive, webgpt} propose agents that can generate sequences of executable commands (\eg, Ctrl+F a token) to navigate through partially observable text environments for information gathering.
The line of research on curiosity-driven exploration and intrinsic motivation shares the same overall goal to seek information \cite{oudeyer2007intrinsic, oudeyer2009intrinsic}.
A subset of them, count-based exploration methods, count the visit of observations or states and encourage agents to gather more information from rarely experienced states \cite{bellemare2016count, Ostrovski17, savinov2018episodic, LiuICML2021}.
Our work also loosely relates to the active learning paradigm, where a system selects training examples wisely so that it achieves better model performance, while also consuming fewer training examples \cite{cohn1994improving, pmlr-v70-bachman17a, fang-etal-2017-learning}. 
Different from existing work, we aim to study explicit querying behavior using language.
We design tasks where querying behavior can either greatly improve efficiency or is needed to succeed. 

In a concurrent work, \citet{nguyen22info} propose a framework tailored to 3D navigation environments: agents can query an oracle to obtain useful information (e.g., current state, current goal and subgoal).
They show that navigation agents can take advantage of an assistance-requesting policy and improve navigation in unseen environments.

\noindent{\bf Reinforcement Learning with External Knowledge:}

Training reinforcement learning agents which use external knowledge sources also received attention recently~\cite{he17, bougie17, kimura21,tutor4rl, rtfm}. Various forms of external knowledge sources are considered. 
\citet{he17} consider a set of documents as external knowledge source. An agent needs to learn to read the documents to solve a task.   \citet{bougie17} consider  environment information obtained by an object detector as external knowledge. They show that  the additional information form the detector enables agents to learn faster. 
\citet{kimura21} consider a set of detailed instructions as knowledge source. They propose an architecture to aggregate  the given external knowledge with the RL model. 
The aforementioned works assume the external knowledge is given and the agent doesn't need to learn to query. In contrast, we consider a task-agnostic interactive knowledge source. In our \qbai and \qtw environments, an agent must learn to actively execute meaningful queries in language to solve a task.

\noindent{\bf Question Generation and Information Retrieval:}
Question generation is a thriving direction  at the intersection of multiple areas like natural language processing and information retrieval.
In the machine reading comprehension literature, \citet{du-etal-2017-learning, yuan2017machine, JainCVPR2018} propose to reverse  question answering: 
given a document and a phrase, a model is required to generate a question.
The question can be answered by the phrase using the document as context.
In  later work, \citet{scialom-staiano-2020-ask} define curiosity-driven question generation.
Query reformulation is a technique which aims to obtain better answers from the knowledge source (\eg, a search engine) by training agents to modify  questions \cite{nogueira-cho-2017-task, Buck18}.
Another loosely related area is multi-hop retrieval \cite{minerva, xiong2021answering, feldman-el-yaniv-2019-multi}, where a large scale supporting knowledge source is involved and  systems must gather  information in a sequential manner.
Inspired by these works, we leverage properties of language such as compositionality, to help form a powerful query representation that is manageable by RL training.

\section{Limitations and Future Work }
\label{sec:conclusion}
In this section, we conclude by discussing limitations of this work and future directions. 

\noindent\textbf{Environments:} 
As an initial attempt to study agents that learn to query knowledge sources with language, we settled on  oracle-based knowledge sources. This ensures better experimental controllability and reproducibility. 
However, it can be  improved in multiple directions.
\begin{enumerate}
    \item Beyond the use of hand-crafted key-value pairs as the knowledge source, a set of more realistic knowledge sources can be considered.
    For instance, databases can be queried using similar template language~\cite{zhong2017seq2sql}; an information retrieval system or a pre-trained question answering system can be used to extract knowledge from large scale language data~\cite{lewis2021question, borgeaud2021improving}; a search engine is naturally queryable~\cite{webgpt}; pre-trained language models can be queried via prompt engineering~\cite{huang2022language}; humans can also be a great knowledge source~\cite{socialai}.
    \item The query grammar can be extended to be more natural and informative (\eg, \cmd{Where's Mary's toy and where can I find it?}). 
    \item We plan to include tasks that require non-linear reasoning. This will further decrease agents' incentive to memorize an optimal trajectory, and presumably increase  generalizability.
\end{enumerate}

\noindent\textbf{Agent design:} For agents, future directions include:
\begin{enumerate}
    \item When the state space is large (\eg, {in \qtw}), agents sometimes keep on querying different question to exploit the exploration bonus.
    This demands a better reward assignment strategy, since agents performing in more complex environments may encounter this issue too.
    \item It is worth exploring other structured knowledge representations~\cite{ammanabrolu20} and parametric memories~\cite{weston2014memory, metalearn_memory} beyond the notebook we used.
    \item Asking questions essentially serves to reduce entropy. 
    One could further use  exploration strategies that maximize information gain~\cite{infogain_max_exploration}.
\end{enumerate}

Overall, we are excited by the challenges and opportunities posed by agents that are able to learn to query external knowledge while acting in their environments.
We strive to call attention from researchers for the development of agents capable of querying external knowledge sources --- we believe this is a strong and natural skill.
We make an initial effort towards this goal, which hopefully can be proven to be valuable and helpful to the community.
\section{Acknowledgement}
\label{sec:ack}
This work is supported in part by Microsoft Research, the National Science Foundation under Grants No.\ $1718221$, $2008387$, $2045586$, $2106825$, MRI $\#1725729$, NIFA award $2020$-$67021$-$32799$, and AWS Research Awards.

\bibliography{biblio}
\bibliographystyle{icml2022}

\clearpage
\appendix
\twocolumn[
{\centering \Large \textbf{Appendix: Asking for Knowledge \includegraphics[width=0.05\textwidth]{figures/afk-8.png}: Training RL Agents to Query External Knowledge Using Language}}
\vspace{0.2cm}
]

\noindent The appendix is structured as follows:
\begin{enumerate}

\item In~\hyperlink{appd:env_task_detail}{\secref{appd:env_task_detail}}, we provide the details of each task in \qbai and \qtw.
\item In~\hyperlink{appd:arch}{\secref{appd:arch}}, we provide the model details of our \afk agent. 
\item In~\hyperlink{appd:implementation_details}{\secref{appd:implementation_details}}, we provide the implementation and training details for the \afk agent. 
\item In~\hyperlink{appd:add_res}{\secref{appd:add_res}},  we provide additional experimental results on~\qbai. 
\item In~\hyperlink{appd:curves}{\secref{appd:curves}}, we provide training curves for all experiments on~\qbai and~\qtw. 

\end{enumerate} 

\noindent The Python code of \qbai, \qtw, the \afk agent and all baselines are available at \url{https://ioujenliu.github.io/AFK}.

\section{Environment and Task Details}
\label{appd:env_task_detail}

\subsection{\qbai}
\label{appd:qbai_detail}

\begin{table}[h] 

\begin{center}
{\small
\begin{tabular}{l|l}
\specialrule{.15em}{.05em}{.05em} 
\toprule
\multicolumn{2}{c}{General information} \\
\midrule
Word vocabulary size  & 62 \\
Function word vocabulary size ($|V_{\text{func}}|$)  & 2 \\
Adjective vocabulary size ($|V_{\text{adj}}|$)  &  22 \\
Noun vocabulary size ($|V_{\text{noun}}|$)  & 24 \\
\# of Physical action  & 7\\
Visual range  & $7\times7$ \\
\# of object colors  & 6 \\
\# of actionable object types & 4 \\
\# of names  & 2 \\
\# of danger zone colors  & 2 \\

\end{tabular}
}
\caption{Statistics of the \qbai environment.}
\label{tb:appd_qbai_stat_general}
\end{center}
\end{table}


\begin{table*}[t] 

\begin{center}
{
\begin{tabular}{c|c|ccccccc}
\specialrule{.15em}{.05em}{.05em} 
                      & Tasks & $|Q_t|$ & \# of rooms  & room size & \makecell{Early \\Terminate}  \\ \toprule\toprule

\multirow{4}{*}{Lv. 1}     
& \sclub &        3	& 1	 & 9$\times$9	& True \\
& \sspade & 1 & 1 & 7$\times$7 &   True  \\
& \sdiamond & 2& 9&  5$\times$5& False     \\
& \sheart  & 1& 2 & 7$\times$7& False       \\ 
 \midrule
 \multirow{6}{*}{Lv. 2}     
& \sclub \sspade&      5    & 2 & 7$\times$7&     True    \\
& \sclub \sdiamond&    4      &9 & 5$\times$5&   True    \\
& \sclub \sheart& 5	 & 2& 7$\times$7&   True  \\
 
& \sspade \sdiamond  &3 &2 & 7$\times$7&  True\\ 
& \sspade \sheart&	2  & 2& 7$\times$7&   True    \\
& \sdiamond \sheart& 4	 & 9& 5$\times$5& False          \\ 
 
 \midrule
 \multirow{4}{*}{Lv. 3}     
& \sclub \sspade \sdiamond&      5   & 2& 7$\times$7& True           \\
& \sclub \sspade \sheart &    6      &3 & 7$\times$7&  True    \\
& \sclub \sdiamond \sheart &	5 & 9& 5$\times$5&     True\\
& \sspade \sdiamond \sheart &	4 & 3& 7$\times$7&     True      \\ 
 \midrule
 \multirow{1}{*}{Lv. 4}     
& \sclub \sspade \sdiamond \sheart&      7     & 9 & 7$\times$7&     True    \\

\end{tabular}
}
\caption{Statistics of each task in \qbai. \sclub: \oib, \sspade: \danger, \sdiamond: \gtfav, \sheart: \od. 
}
\label{tb:appd_qbai_stat}
\end{center}
\end{table*}
The general statistics of \qbai tasks are summarized in~\tabref{tb:appd_qbai_stat_general}. The statistics for each individual \qbai task are summarized in~\tabref{tb:appd_qbai_stat}, where $|Q_t|$ represents the number of `good queries' an agent should make to solve a task efficiently. `Early Terminate' indicates that an episode will be terminated if the agent makes a mistake, \eg, stepping on a danger zone or opening the wrong box. 
In addition,  we present the details of the four basic \qbai tasks in the following. 
 
\textbf{\oib \sclub}: There are two suitcases in the environment. Each suitcase contains one toy. The instruction is \cmd{find <name>'s toy}, where \cmd{<name>} is sampled from a set of names at the start of each episode. However, the agent doesn't know what is the referred toy. Neither does it know the content of each suitcase. The episode terminates when a suitcase is opened by the agent. Therefore, the agent needs to ask multiple question to figure out what the desired toy is and which suitcase to open.  If the opened suitcase contains the desired toy, the agent receives a positive reward. Otherwise, it doesn't receive any reward.

\textbf{\danger \sspade}: There are different colors of tiles in the environment. One of the colors represents the danger zone. The episode terminates if the agent steps on a danger zone. The instruction is \cmd{avoid danger zone, and go to the green target square}. However, the agent doesn't know what color represents the danger zone. Therefore, to safely reach the target square and receive rewards, the agent must ask the oracle for information on the danger zone. Importantly, the color of the danger zone differs from episode to episode. 

\textbf{\gtfav \sdiamond}: There are nine rooms in the environment. The instruction is \cmd{Go to <name>'s favorite toy}, where \cmd{<name>} and \cmd{<name>'s favorite toy} are sampled from a set of names and a set of toys at the start of each episode. There are irrelevant objects scattered around the environment. To solve the task efficiently, the agent should issue queries to figure out what and where is the referred toy. Note, if the agent doesn't ask any question, it can still solve the task, but in a much less efficient manner, \ie, by exhaustively searching all rooms for the referred toy. The agent receives positive reward when it goes to the referred toy. 

\textbf{\od \sheart}: There are three keys and one door in the environment. One of the three keys could open the door. The agent needs to find the right key and open the door to complete the task and receive a positive reward. The instruction is \cmd{Find the key to the door}. Note, the agent could still complete the task without asking any question, \ie, by exhaustively trying all keys.

\begin{table}[!ht] 

\begin{center}
{\small
\begin{tabular}{l|l}
\specialrule{.15em}{.05em}{.05em} 
\toprule
\multicolumn{2}{c}{General information} \\
\midrule
Word vocabulary size  & 835 \\
\# Function word ($|V_{\text{func}}|$) & 7 \\
\# Adjective  ($|V_{\text{adj}}|$)  & 38 \\
\# Noun ($|V_{\text{noun}}|$)  & 45 \\
\# holders & 6 \\
\# ingredients & 42 \\
\# cuttable ingredients & 26 \\
\midrule
\multicolumn{2}{c}{Take 1} \\
\midrule
  \# recipes & 42 \\
  \# configurations & 1242 \\
  Avg. instruction length & 33.75 ± 0.44 \\ 
  Avg. walkthrough length & 1.49 ± 0.50 \\ 
  Avg. nb. entities & 8.63  ± 0.96 \\ 
  Avg. observation length & 150.94  ± 65.43  \\ 
  Avg. valid actions per step & 4.90  ± 1.29  \\ 
\midrule
\multicolumn{2}{c}{Take 2} \\
\midrule
  \# recipes & 1722 \\
  \# configurations & 1332156 \\
  Avg. instruction length & 36.48 ± 0.60  \\ 
  Avg. walkthrough length & 3.01  ± 0.70  \\ 
  Avg. nb. entities & 10.69  ± 1.22  \\ 
  Avg. observation length & 141.76  ± 57.56  \\ 
  Avg. valid actions per step & 5.83  ± 1.65  \\ 
\midrule
\multicolumn{2}{c}{Take 1 Cut} \\
\midrule
  \# recipes & 17576 \\
  \# configurations & 1026 \\
  Avg. instruction length & 33.80 ± 0.40  \\ 
  Avg. walkthrough length & 2.50  ± 0.50  \\ 
  Avg. nb. entities & 9.37  ± 0.84  \\ 
  Avg. observation length & 143.00  ± 59.30  \\ 
  Avg. valid actions per step & 5.45  ± 1.70  \\ 
\midrule
\multicolumn{2}{c}{Take 2 Cut} \\
\midrule
  \# recipes & 274625000 \\
  \# configurations & 859620 \\
  Avg. instruction length & 36.61 ± 0.55  \\ 
  Avg. walkthrough length & 5.00  ± 0.71  \\ 
  Avg. nb. entities & 11.67  ± 1.13  \\ 
  Avg. observation length & 138.59  ± 50.04  \\ 
  Avg. valid actions per step & 7.73  ± 2.54  \\ 
  
\end{tabular}
}
\caption{ Statistics of the \qtw environment.}
\label{tb:appd_qtw_stat}
\end{center}
\end{table}

\begin{figure}
\scriptsize
\begin{mdframed}
\begin{elaboration}
  \parbox{.99\textwidth}{
\emph{\textbf{Instruction:}} You find yourself at friend's house and you are both hungry! Collect some parsley to prepare a delicious meal. Ask Charlie to know where to find each ingredient and for the recipe directions.
}
\end{elaboration}
\begin{elaboration}
\parbox{.99\textwidth}{
\emph{\textbf{Description:}} You are in a kitchen. A messy kind of place.

\vspace{0.1cm}
You see a closed fridge. You can make out an oven. Look over there! a counter. The counter is wooden. However, the counter, like an empty counter, has nothing on it. Oh! Why couldn't there just be stuff on it? You can see a stove. The stove appears to be empty. You see a griller. Is this what you came to TextWorld for? This... griller?

\vspace{0.1cm}
There is a cookbook on the floor.

\vspace{0.2cm}
\emph{\textbf{Inventory:}} You are carrying: a knife.

\vspace{0.2cm}
\emph{\textbf{Notebook:}} \{"You find yourself \dots for the recipe directions."\}
}
\end{elaboration}
\begin{flushleft}
\emph{Action 1:} \textbf{ask Charlie where's the parsley}
\end{flushleft}
\begin{elaboration}
  \noindent\parbox{.99\textwidth}{
\emph{\textbf{Feedback:}} The parsley is in the oven.

\vspace{0.2cm}
\emph{\textbf{Notebook:}} \{"You find yourself \dots for the recipe directions.", "The parsley is in the oven."\}
}
\end{elaboration}
\begin{flushleft}
\emph{Action 2:} \textbf{open oven}
\end{flushleft}
\begin{elaboration}
  \noindent\parbox{.99\textwidth}{
\emph{\textbf{Feedback:}} Opened.
}
\end{elaboration}
\begin{flushleft}
\emph{Action 3:} \textbf{take parsley}
\end{flushleft}
\begin{elaboration}
  \parbox{.99\textwidth}{
\emph{\textbf{Feedback:}} Taken.
}
\end{elaboration}
\begin{flushleft}
\emph{Action 4:} \textbf{ask Charlie how to cut the parsley}
\end{flushleft}
\begin{elaboration}
  \parbox{.99\textwidth}{
\emph{\textbf{Feedback:}} The parsley needs to be sliced. Use a knife to cut it.

\vspace{0.2cm}
\emph{\textbf{Notebook:}} \{"You find yourself \dots for the recipe directions.", "The parsley is in the oven.", "The parsley needs to be sliced. Use a knife to cut it."\}
}
\end{elaboration}
\begin{flushleft}
\emph{Action 5:} \textbf{slice parsley}
\end{flushleft}
\begin{elaboration}
  \parbox{.99\textwidth}{
\emph{\textbf{Feedback:}} Sliced.
}
\end{elaboration}
\begin{flushleft}
\emph{\textbf{Done after 5 steps. Score 1/1.}}
\end{flushleft}

\end{mdframed}
\caption{An excerpt from a \qtw game.}
\label{fig:qtw-example}
\end{figure}

\subsection{\qtw}
\label{appd:qtw_detail}

For all games, the objective is to find cooking ingredients which are randomly hidden throughout the kitchen. Once found, those ingredients may require some processing depending on the task difficulty. Once all required ingredients are in the player's inventory and processed the right way, the game terminates with a reward of 1.
We provide statistics of the \qtw environment in~\tabref{tb:appd_qtw_stat} and a transcript of a game can be seen in \figref{fig:qtw-example}.

\textbf{Take [1/2]}: In this task, the player has to find 1 or 2 ingredients mentioned in the {\color{darkred}instruction}. Ingredients are either visible to the agent right from the start (\eg, on the table), or hidden inside some container that needs to be opened first (\eg, in the fridge). The player can ask the oracle where it can find a particular object (\eg, \cmd{Ask Charlie\footnote{In \qtw, the oracle is named Charlie.} where's hot pepper?}). In return, the oracle will indicate where the object can be found (\eg, \token{Hot pepper is in the fridge.}).

\textbf{Take [1/2] + Cut}: This task extends \textbf{Take [1/2]} as the ingredients also need to be cut in the right way (\ie, \token{chopped}, \token{sliced}, or \token{diced}). Each cutting type is achieved by a different action command (\ie, \cmd{chop X}, \cmd{slice X}, or \cmd{dice X}) while the player is holding a knife in their inventory. The player can also ask the oracle how to process a particular ingredient (\eg, \cmd{Ask Charlie how to cut the hot pepper?}\footnote{Nonessential words can be omitted, \eg, \cmd{Ask Charlie how hot pepper?}}). In return, the oracle will indicate which type of cutting is needed (\eg, \token{Hot pepper needs to be sliced, use a knife to cut it}).
Note, in the reported games, the player always start with a kitchen knife in their inventory.

\section{Modeling Details}
\label{appd:arch}
In this section, we provide detailed information regarding our agents.
In Appendix~\ref{appd:arch:qbai}, we describe our agent used for the \qbai environments.
In Appendix~\ref{appd:arch:qtw}, we describe our agent used for the \qtw environments.

\subsection{\afk --- \qbai}
\label{appd:arch:qbai}
\noindent\textbf{Observation Encoder ($f_{\text{obs}}$):}
Following BabyAI~\cite{babyai}, the environment observation $o^{\text{env}}$ of \qbai is a $7\times7\times4$ symbolic observation that contains a partial and local egocentric view of the environment and the direction of the agent. To encode $o^{\text{env}}$, we use a 
convolutional neural network (CNN). Following~\citet{babyai}, the observation encoder consists of three convolutional layers. The first convolutional layer has  $128$ filters of size $8 \times 8$ and stride $8$.
The second and third convolutional layers have  $128$ filters of size $3 \times 3$ and stride $1$. 
Batch normalization and ReLU unit are applied to the output of each layer. At the end, a 2D pooling layer with filter size $2 \times 2$ is applied to obtain the representation $h_o$ of $256$ dimensions. 

\noindent\textbf{Word Encoder ($f_{\text{note}}$):}
Following~\citet{babyai}, we use a gated recurrent unit (GRU)~\cite{gru} to perform word encoding.
Specifically, for each $v_i \in A_0$, we have $h_i = f_{\text{gru}}(v_i) \in \cR^{|v_i| \times l}$, where $|v_i|$ is the number of words in $v_i$ and $l=128$ is the encoding dimension.

\noindent\textbf{Aggregator ($f_{\text{att}}$):}
Following the \noquery baseline~\cite{babyai}, the aggregator consists of FiLM~\cite{film} modules, $f_{\text{FiLM}}$, followed by a long short term memory (LSTM)  $f_{\text{LSTM}}$~\cite{lstm}. That is, $f_{\text{att}} = f_{\text{LSTM}} \circ f_{\text{FiLM}}$. Specifically, we stack two FiLM modules. Each FiLM module has $128$ filters with size $3 \times 3$ and the output dimension is $128$. The LSTM has $128$ units.  

\noindent\textbf{Physical Action and Query Heads ($\pi_{\text{switch}}, \pi_{\text{phy}}, \pi_{\text{fun}}, \pi_{\text{adj}}, \pi_{\text{noun}}$):}
The switch head $\pi_{\text{switch}}$, physical action head $\pi_{\text{phy}}$, and function word head $\pi_{\text{fun}}$ are two-layer MLPs with $64$ units in each layer. The output dimension of $\pi_{\text{switch}}, \pi_{\text{phy}}, \pi_{\text{fun}}$ are $2, 7, 2$. $\pi_{\text{adj}}$ and $\pi_{\text{noun}}$ are single-head pointer networks (\secref{subsec:pointer}) with hidden dimension $l=128$.

\subsection{\afk --- \qtw}
\label{appd:arch:qtw}

\noindent\textbf{Text Encoder ($f_{\text{obs}}$, $f_{\text{note}}$):} 
Due to the nature of the \qtw environment, where all inputs are in pure text, we share the two encoders (\ie, $f_{\text{obs}}$ and $f_{\text{note}}$) in our text agent.

We use a transformer-based text encoder, which consists of an embedding layer and a transformer block \citep{transformer}.
Specifically, we tokenize an input sentence (either a text observation or an entry in the notebook) with the spaCy tokenizer.\footnote{\url{https://spacy.io/}}
We convert the sequence of tokens into 128-dimensional embeddings, the embedding matrix is initialized randomly.

The transformer block consists of a stack of 4 convolutional layers, a self-attention layer, and a 2-layer MLP with a ReLU non-linear activation function in between. 
Within the block, each convolutional layer has 128 filters, with a kernel size of 7.
The self-attention layers use a block hidden size of 128, with 4 attention heads.
Layer normalization \citep{ba16layernorm} is applied after each layer inside the block. 
We merge positional embeddings into each block's input.

Given an input $o \in \mathbb{R}^{|o|}$, where $|o|$ denotes the number of tokens in $o$, the encoder  produces a representation $h_{o} \in \mathbb{R}^{|o| \times H}$, with $H = 128$ the hidden size.

In practice, we use mini-batches to parallelize the training. 
Following standard NLP methods, we use special padding tokens when the number of tokens within a batch are different, we use masks to prevent the model from taking the padding tokens into computation.
A text input $o$ will be associated with a mask $m_o \in \mathbb{R}^{|o|}$.

Note for all the three agent variants (\ie, \noquery, \query and \afk), we use the concatenation of [\textit{feedback}, \textit{description}, \textit{inventory}] as the input to $f_{\text{obs}}$.
See examples of \textit{feedback}, \textit{description} and \textit{inventory} text in \figref{fig:qtw-example}.

\noindent\textbf{Aggregator ($f_{\text{att}}$):} 
To aggregate two input encodings $P \in \mathbb{R}^{|P| \times H}$ and $Q \in \mathbb{R}^{|Q| \times H}$, we use the standard multi-head attention mechanism \citep{transformer}.
Specifically, we use $P$ as the \textit{query}, $Q$ as the \textit{key} and \textit{value}. 
This results in an output $P_Q \in \mathbb{R}^{|P| \times H}$, where at every time step $i \in [0, |P|)$, $P_Q^i$ is the weighted sum of $Q$, the weight is the attention of $P^i$ on $Q$.
We refer readers to \citet{transformer} for detailed information.

We apply a residual connection on top of the multi-head attention mechanism in order to maintain the original information contained in $P$. 
Specifically, 
\begin{equation}
    h_{PQ} = \text{Tanh}(\text{Linear}([P_Q; P])),
\end{equation}
where $h_{PQ} \in \mathbb{R}^{|P| \times H}$, brackets $[\cdot;\cdot]$ denote vector concatenation.

We denote the above attention layer as 
\begin{equation}
    h_{PQ} = \text{Attention}(P, Q).
\end{equation}
Using two of such layers (without sharing parameters), we aggregate three inputs: $h_\text{obs} \in \mathbb{R}^{|\text{obs}| \times H}$, $h_\text{task} \in \mathbb{R}^{|\text{task}| \times H}$ and $h_{s} \in \mathbb{R}^{|\text{note}| \times H}$, where $|\text{obs}|$, $|\text{task}|$ and $|\text{note}|$ denote the number of tokens in a text observation, the number of tokens in the instruction, and the number of nodes in the notebook:
\begin{equation}
\begin{aligned}
    h_{\text{obs}, \text{task}} &= \text{Attention}(h_\text{obs}, h_\text{task}),\\
    h_{x} &= \text{Attention}(h_{\text{obs}, \text{task}}, h_{s}).\\
    \end{aligned}
\end{equation}
Here, $h_{\text{obs}, \text{task}} \in \mathbb{R}^{|\text{obs}| \times H}$, $h_x \in \mathbb{R}^{|\text{obs}| \times H}$.

\noindent\textbf{Action Generator ($\pi_{\text{func}}$, $\pi_{\text{adj}}$, $\pi_{\text{noun}}$):} 
In \qtw, all actions follow the same format of \cmd{<func, adj, noun>}. Therefore, the query action space $\cA_{q}$ and the physical action space $\cA_{\text{phy}}$ are shared (\ie, the vocabularies are shared).
We use a three-head module to generate three vectors. Their sizes correspond to the function word, adjective, and noun vocabularies.
The generated vectors are used to compute an overall Q-value.

Taking the aggregated representation $h_x \in \mathbb{R}^{|\text{obs}| \times H}$ as input, we first compute its masked average, using the mask of the text observation. This results in $\overline{h_s} \in \mathbb{R}^{H}$.

Specifically, the action generator consists of four multi-layer perceptrons (MLPs): 
\begin{equation}
\begin{aligned}
    h_{\text{shared}} &= \text{ReLU}(\text{Linear}_{\text{shared}}(\overline{h_s})), \\
    Q_{\text{func}} &= \text{Linear}_{\text{func}}(h_{\text{shared}}), \\
    Q_{\text{adj}} &= \text{Linear}_{\text{adj}}(h_{\text{shared}}), \\
    Q_{\text{noun}} &= \text{Linear}_{\text{noun}}(h_{\text{shared}}). \\
\end{aligned}
\end{equation}
Here, $Q_{\text{func}} \in \mathbb{R}^{|\text{func}|}$, $Q_{\text{adj}} \in \mathbb{R}^{|\text{adj}|}$, $Q_{\text{noun}} \in \mathbb{R}^{|\text{noun}|}$. $|\text{func}|$, $|\text{adj}|$, and $|\text{noun}|$ denote the vocabulary size of function words, adjectives, and nouns.

In order to alleviate the difficulties caused by a large action space, similar to the pointer mechanism in the \qbai agent, we apply masks over vocabularies when sampling actions.
In the masks, only tokens appearing in the current notebook are labeled as 1, \ie, the text agent only performs physical interaction with objects noted in its notebook. It also only asks questions about objects it has heard of.

Finally, we compute the Q-value of an action \cmd{<u, v, w>}:
\begin{equation}
Q_{<u, v, w>} = (Q_u + Q_v + Q_w) / 3, \\
\end{equation}
where $u$, $v$ and $w$ are tokens in the function word, adjective, and noun vocabulary.

\section{Implementation Details}
\label{appd:implementation_details}

In this section, we provide implementation and training details of our agents.
In Appendix~\ref{appd:implementation:qbai}, we provide implementation details for our agent used for the \qbai environments.
In Appendix~\ref{appd:implementation:qtw}, we provide implementation details for our agent used for the \qtw environments.

\subsection{\afk --- \qbai}
\label{appd:implementation:qbai}
We closely follow the training procedure of the publicly available code of the BabyAI \noquery agent~\cite{babyai}. We train our \afk and all baselines with PPO~\cite{ppo}. Specifically, we use the Adam~\cite{adam} optimizer with learning rate $0.0001$. We update the model every $2560$ environment steps. The batch size is $1280$. The PPO epoch is $4$ and the discount factor is $0.99$. We use $64$ parallel processes for collecting data from the environment. 
The scaling factor $\beta$ of the episodic exploration bonus is set to $0.1$ for all experiments. For all experiments, we study uni-gram and bi-gram similarity models and report the better results. We tuned the episodic-exploration scaling factor $\beta \in \{0.001, 0.01, 0.1, 0.5\}$, hidden size of the pointer network $l \in \{32, 64, 128, 256\}$, learning rate $\in \{10^{-5}, 10^{-4}, 10^{-3}\}$, and similarity function $\in \{\text{uni-gram}, \text{bi-gram}\}$. We train all agents with 5 different random seeds: $[24, 42, 123, 321, 3407]$.

\subsection{\afk --- \qtw}
\label{appd:implementation:qtw}

We adopt the training procedure from the official code base released by TextWorld authors~\cite{textworld}.
Our text agent is trained with Deep Q-Learning \cite{mnih2013playing}.
We use a prioritized replay buffer with memory size of $500,000$, and a priority fraction of $0.5$.
During model update, we use a replay batch size of $64$.
We use a discount factor $\gamma = 0.9$.
We use noisy nets, with a $\sigma_0$ of $0.5$.
We update the target network after every 1000 episodes.
We sample the multi-step return $n \sim \text{Uniform}[1, 3]$.
We refer readers to \citet{hessel18rainbow} for more information about different components of DQN training.

For all experiments, we use \emph{Adam} \citep{adam} as the optimizer.
The learning rate is set to $0.00025$ with a clip gradient norm of $5$.
We train all agents with 5 different random seeds: $[24, 42, 123, 321, 3407]$. 
For replay buffer data collection, we use a batch size of $20$.
We train our agents with 500K episodes, each episode has a maximum number of steps $20$.
After every $2$ data collection steps, we randomly sample a batch from the replay buffer, and perform a network update.

\paragraph{Resources:}
We use a mixture of Nvidia V100, P100 and P40 GPUs to conduct all the experiments.
On average, experiments on \qbai take 1 day,
experiments on \qtw take 2 days.

\section{Additional Results}
\label{appd:add_res}
\paragraph{Success rate and episode length:}

\begin{table*}[t] 

\begin{center}
{
\begin{tabular}{c|c|cccccc}
\specialrule{.15em}{.05em}{.05em} 
                      &  & \multicolumn{2}{c}{\makecell{No Query}} & \multicolumn{2}{c}{\makecell{Query Baseline} }  & \multicolumn{2}{c}{\makecell{AFK (Ours)} } \\ \toprule\toprule
                      				& Tasks & Succ.($\%$)             & Eps. Len.             & Succ.($\%$)              & Eps. Len.           & Succ.($\%$)              & Eps. Len.         \\ \hline

\multirow{4}{*}{Lv. 1}     
& \sclub &   50.5$\pm$0.6      & 5.9$\pm$0.1	&49.8$\pm$1.1 &6.0$\pm$0.1	 &	\bf{100.0$\pm$0.0} &  \bf{10.8$\pm$0.1}              \\
& \sspade &  68.3$\pm$0.8     &	8.2$\pm$0.0	&73.8$\pm$0.9 &8.3$\pm$0.1	 &	\bf{100.0$\pm$0.0} &   \bf{10.4$\pm$0.0}           \\
& \sdiamond &	98.9$\pm$0.4  &	30.2$\pm$1.5	& 99.3$\pm$0.2 &26.7$\pm$1.5	 &	\bf{100.0$\pm$0.0} &    \bf{16.8$\pm$6.7}     \\
& \sheart &	99.7$\pm$0.2 &	26.2$\pm$0.9	& 85.3$\pm$1.1 & 36.8$\pm$1.0	 &	\bf{100.0$\pm$0.0} &    \bf{20.6$\pm$0.2}          \\ 
 \midrule
 \multirow{6}{*}{Lv. 2}     
& \sclub \sspade&         0.0$\pm$0.0  & 43.3$\pm$0.7		& 0.0$\pm$0.0 &	44.1$\pm$0.6 &	\bf{90.3$\pm$1.8} &      \bf{15.5$\pm$0.3}          \\
& \sclub \sdiamond&         0.1$\pm$0.1  &	224.8$\pm$0.4	& 0.6$\pm$0.5 & 224.3$\pm$0.5	 &	\bf{94.3$\pm$2.3} &    \bf{ 18.8$\pm$0.3}       \\
& \sclub \sheart&	0.0$\pm$0.0  & 98.0$\pm$0.0	& 0.0$\pm$0.0 & 98.0$\pm$0.0	 &	\bf{99.0$\pm$0.4} &    \bf{ 32.3$\pm$0.6 }     \\
 
& \sspade \sdiamond &0.4$\pm$0.1&90.8$\pm$1.3&0.2$\pm$0.2&91.4$\pm$1.1&\bf{100.0$\pm$0.0} & \bf{ 14.5$\pm$0.0 }\\ 
& \sspade \sheart&	 0.0$\pm$0.0   &	 76.8$\pm$2.1 	&  0.0$\pm$0.0  & 79.5$\pm$1.7 	 &	\bf{ 0.0$\pm$0.0 } &    \bf{ 79.0$\pm$0.9 }     \\
& \sdiamond \sheart&	10.8$\pm$1.6  & 202.9$\pm$4.0 	& 10.2$\pm$2.1 & 203.2$\pm$5.1	 &	\bf{98.7$\pm$0.2} &    \bf{66.9$\pm$3.2 }          \\ 
 
 \midrule
 \multirow{4}{*}{Lv. 3}     
& \sclub \sspade \sdiamond&         0.0$\pm$0.0  & 91.8$\pm$1.1	& 0.0$\pm$0.0 & 92.2$\pm$0.6	 &	{0.15$\pm$0.2} &    85.8$\pm$1.4          \\
& \sclub \sspade \sheart &         0.0$\pm$0.0  & 93.9$\pm$3.4	& 0.0$\pm$0.0 & 102.5$\pm$3.5	 &	{0.0$\pm$0.0} &   109.4$\pm$2.3      \\
& \sclub \sdiamond \sheart &	0.0$\pm$0.0  & 225.0$\pm$0.0	& 0.0$\pm$0.0 &  225.0$\pm$0.0 	 &	{2.1$\pm$0.8} &    {220.9$\pm$1.1 }     \\
& \sspade \sdiamond \sheart &	4.3$\pm$1.0  &99.3$\pm$2.9	& 4.4$\pm$0.8 &  97.7$\pm$3.0	 &	{4.8$\pm$0.9} &   105.0$\pm$2.0         \\ 
 \midrule
 \multirow{1}{*}{Lv. 4}     
& \sclub \sspade \sdiamond \sheart&          0.0$\pm$0.0   &	 96.6$\pm$8.1	 &  0.0$\pm$0.0  & 	150.5$\pm$7.5  &	{ 0.0$\pm$0.1 } &    177.2$\pm$9.2            \\

\end{tabular}
}
\caption{Evaluation success rate and episode length on \qbai. \sclub: \oib, \sspade: \danger, \sdiamond: \gtfav, \sheart: \od. 
}
\label{tb:appd_main_new}
\end{center}
\end{table*}

 In~\tabref{tb:appd_main_new} we report success rate and episode length of \noquery, \query, and \afk on all levels of \qbai  tasks. Note, due to the early termination mechanism, the comparison of episode length is only meaningful when the agent is able to solve the task. For instance, in \oib (\sclub) and \danger (\sspade), \noquery and \query have shorter episode length than \afk because they either step on the danger tile or open the wrong box, resulting in the termination of an episode. 
 
\begin{figure}[]
\centering
\renewcommand{\arraystretch}{0}
\begin{tabular}{ccc}
\includegraphics[width=0.25\textwidth]{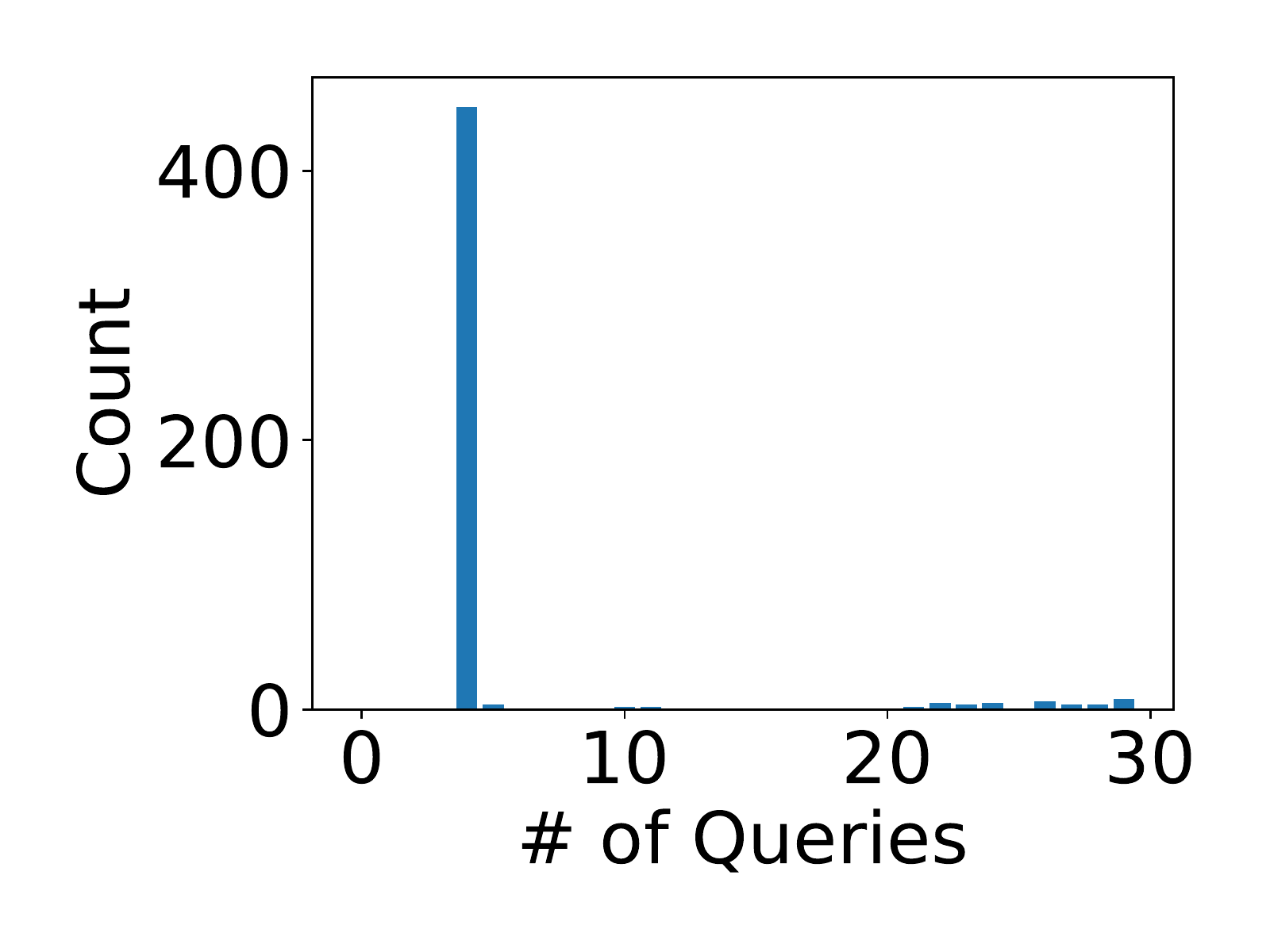}
&
\hspace{-0.6cm}
\includegraphics[width=0.25\textwidth]{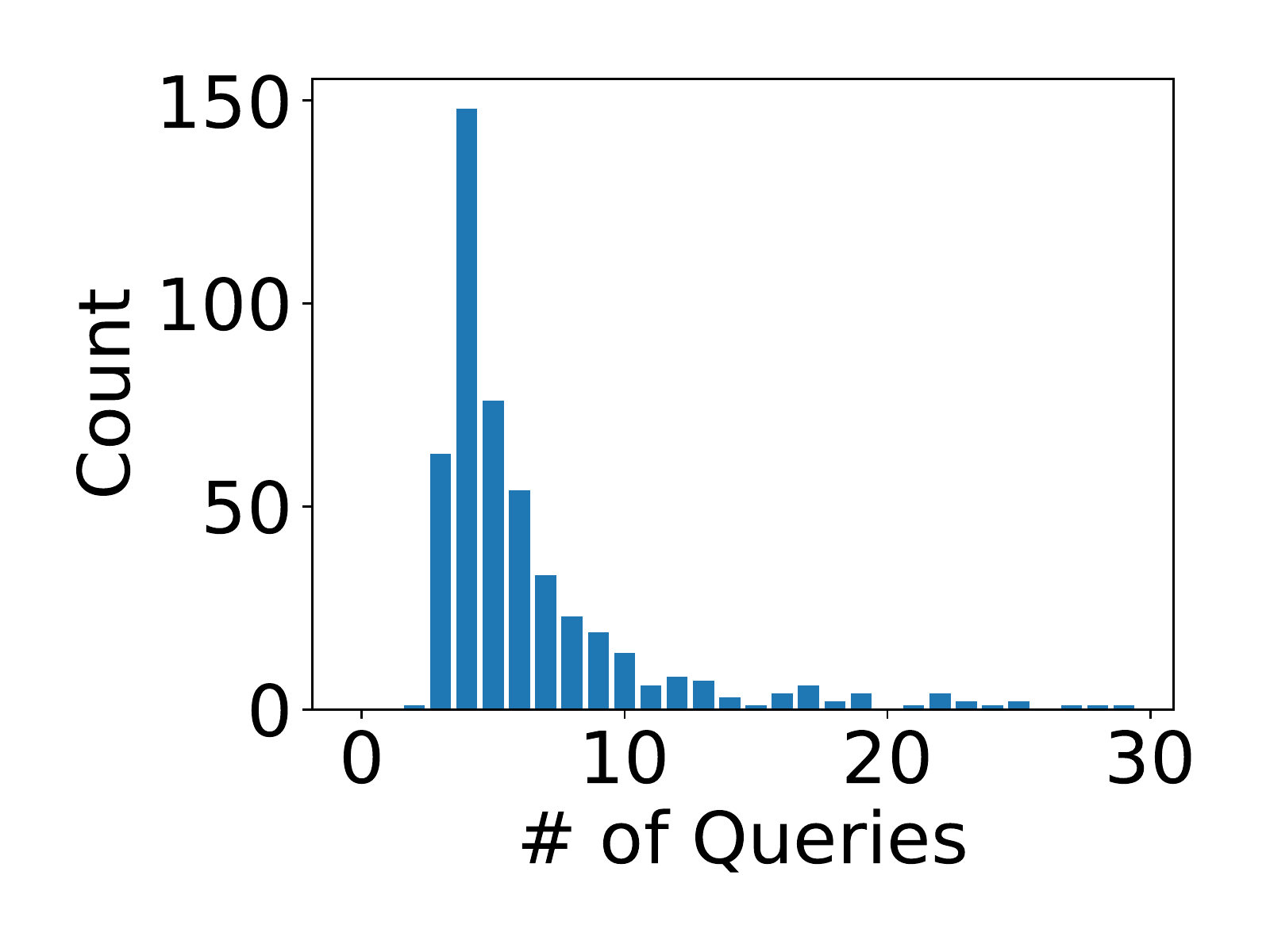}

\end{tabular}%
\caption{Evaluation episode count over number of queries. \textbf{Left}: train on \sclub\sspade, evaluate on \sclub\sspade. 
         \textbf{Right}: train on \sspade\sdiamond$+$\sdiamond\sclub, evaluate on \sclub\sspade.
         \sclub: \textbf{Obj. in Box}, \sspade: \danger, \sdiamond: \gtfav. 
         } 
\label{fig:gen}
\end{figure}

\paragraph{Number of queries made by an AFK agent in seen and unseen tasks:} 
To better understand the agent's behavior in seen and unseen tasks, we report the number of queries an agent made  
across $500$ evaluation episodes. As shown in~\figref{fig:gen} (left), when an agent is trained and evaluated on the same tasks (\sclub\sspade), in most episodes, the agent makes four queries, which is the optimal number of queries of the task.
In contrast, when the agent is trained and evaluate on different tasks, \figref{fig:gen} (right), it made more queries.

\section{Training Curves}
\label{appd:curves}
\begin{figure*}[t]
\centering

\begin{tabular}{ccc}

\includegraphics[width=0.33\textwidth]{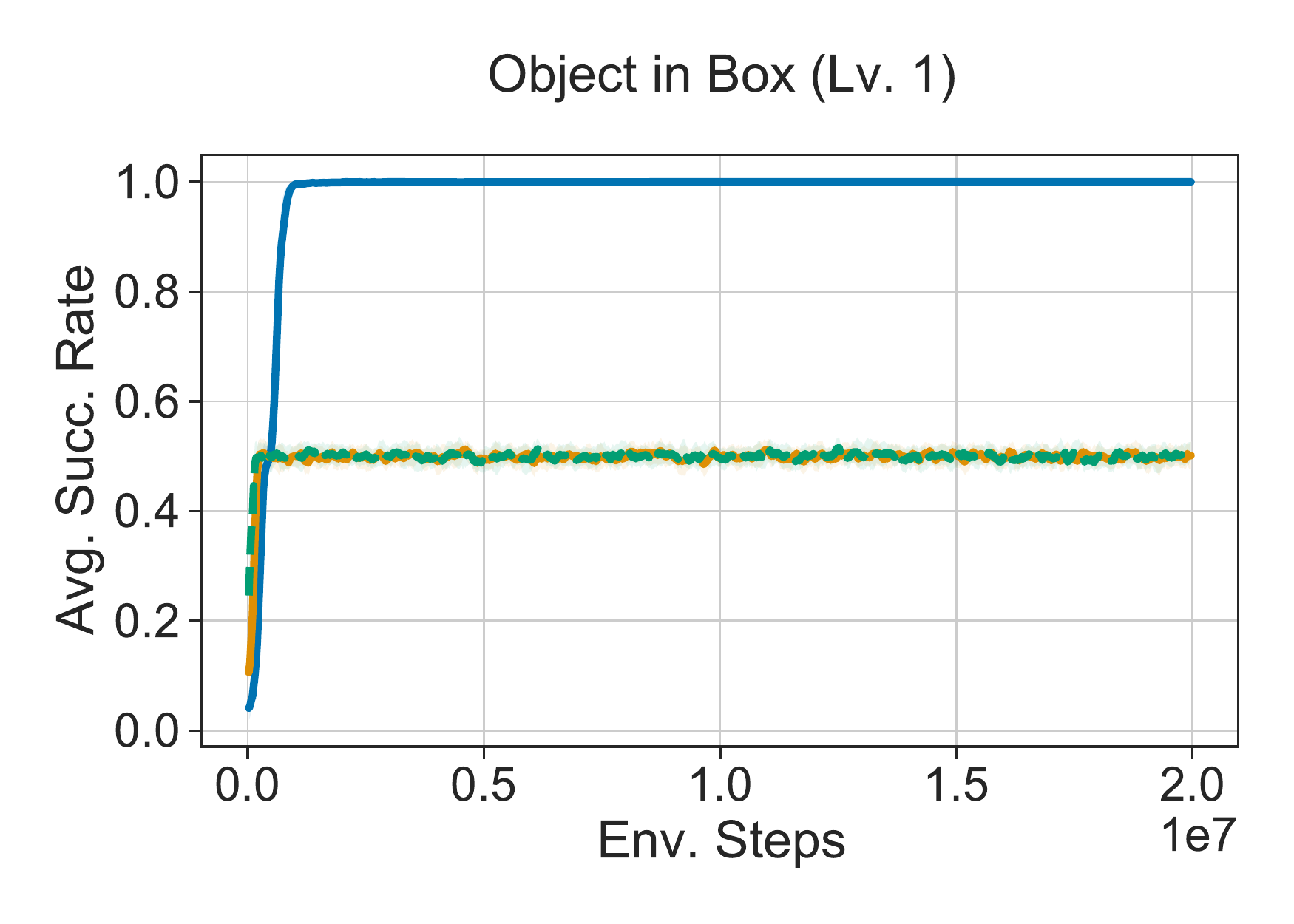}
&
\hspace{-0.6cm}\includegraphics[width=0.33\textwidth]{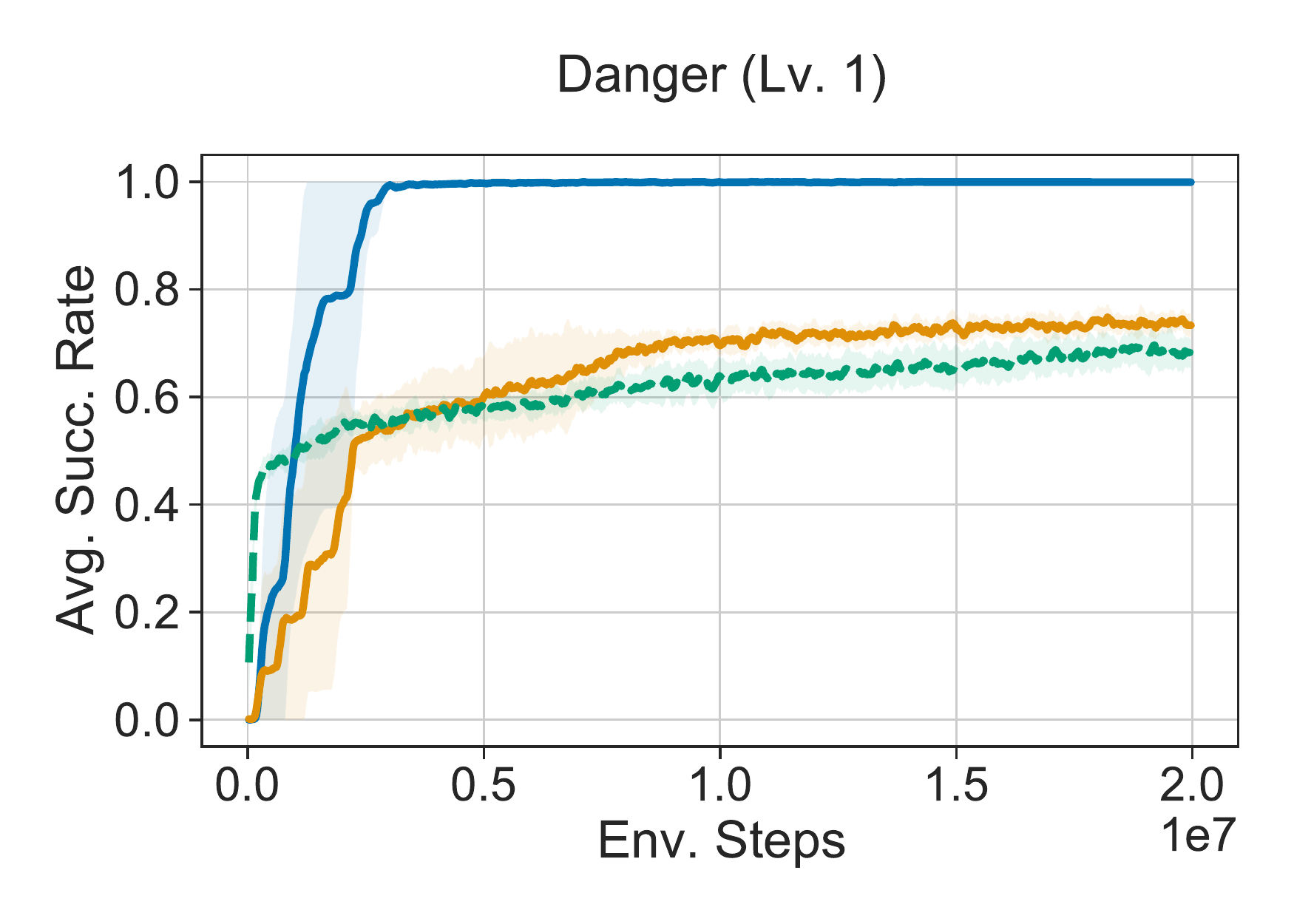}
&
\hspace{-0.6cm}\includegraphics[width=0.33\textwidth]{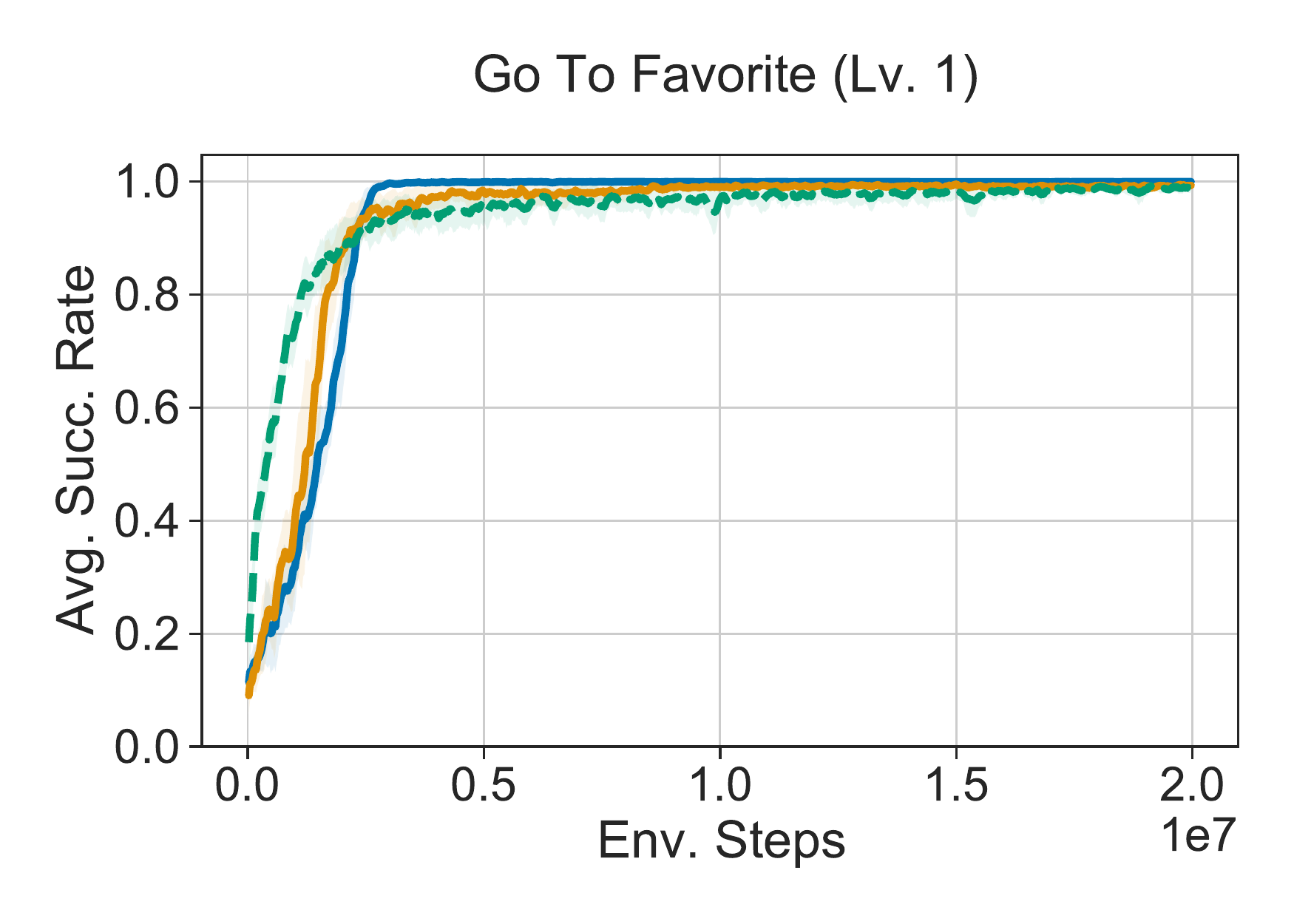}\\

\includegraphics[width=0.33\textwidth]{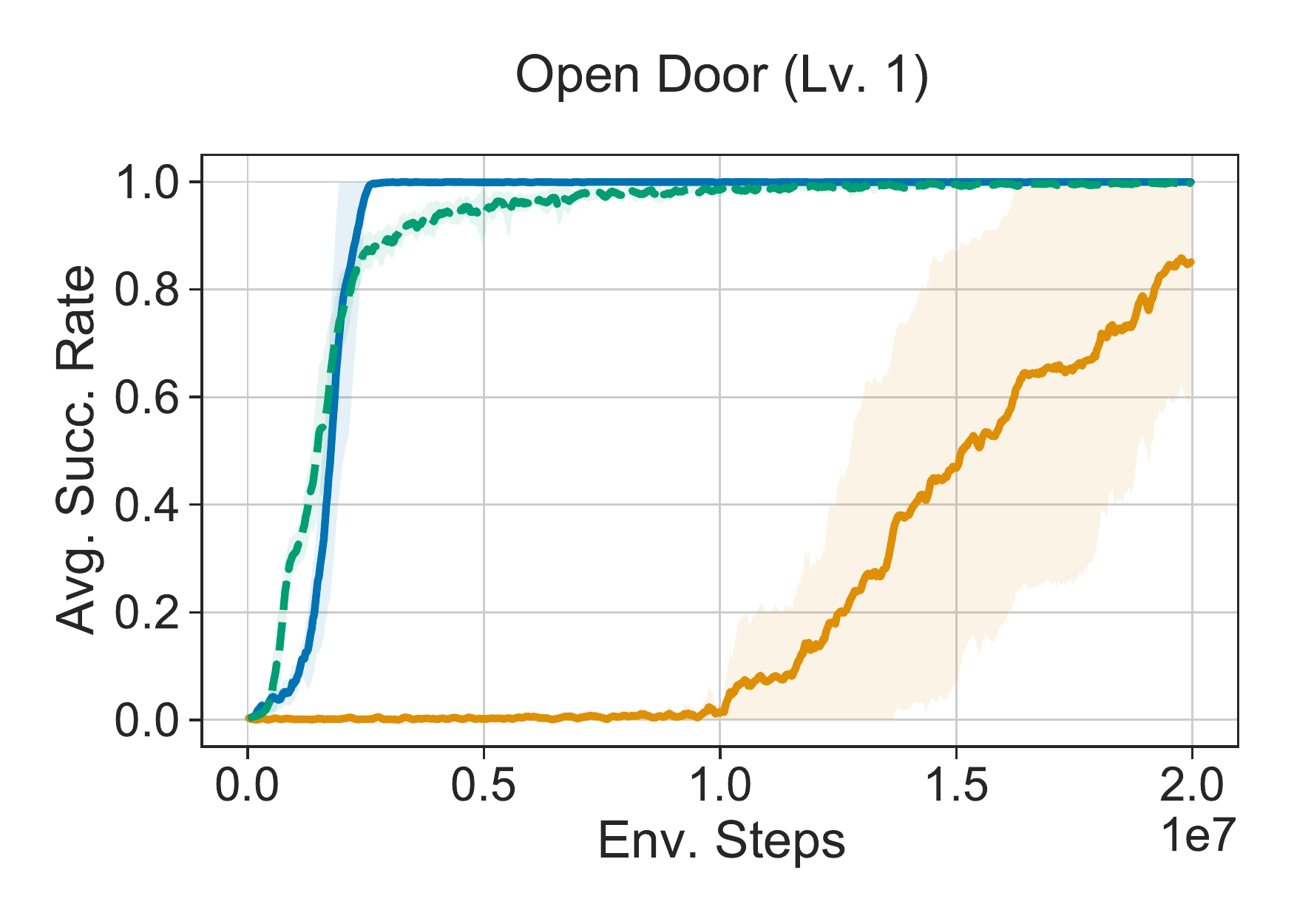}
&
\hspace{-0.6cm}\includegraphics[width=0.33\textwidth]{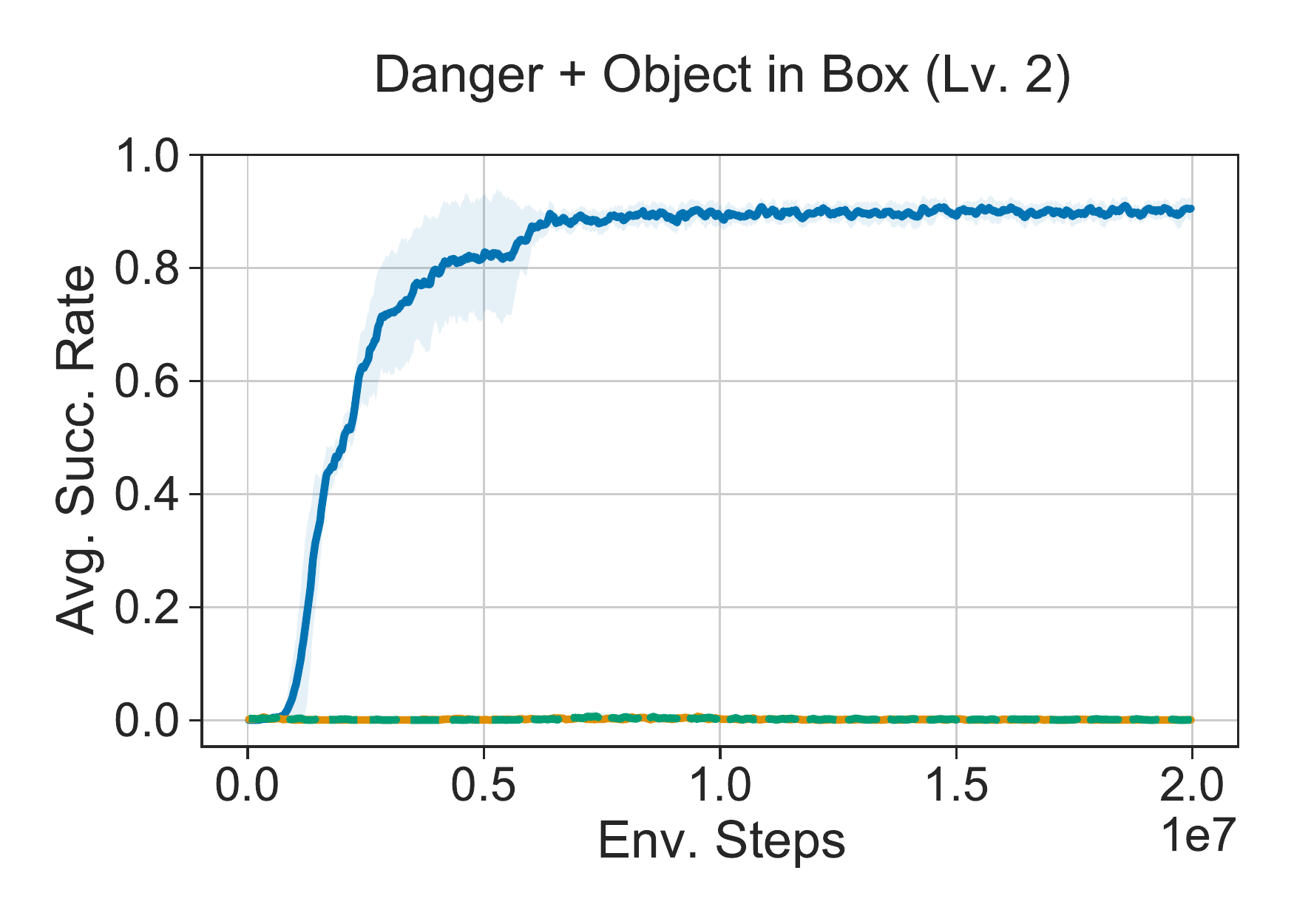}
&
\hspace{-0.6cm}\includegraphics[width=0.33\textwidth]{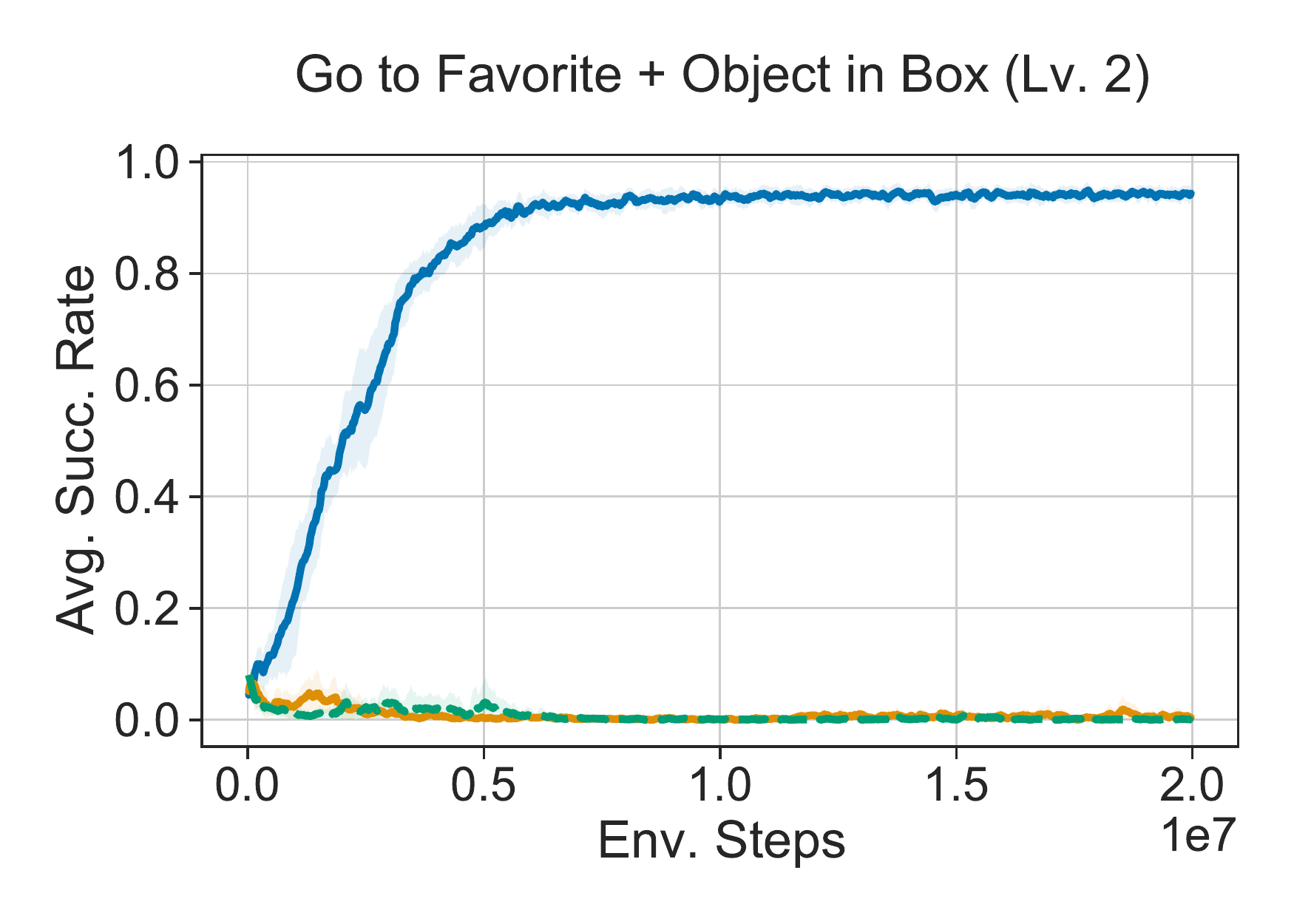}\\

\includegraphics[width=0.33\textwidth]{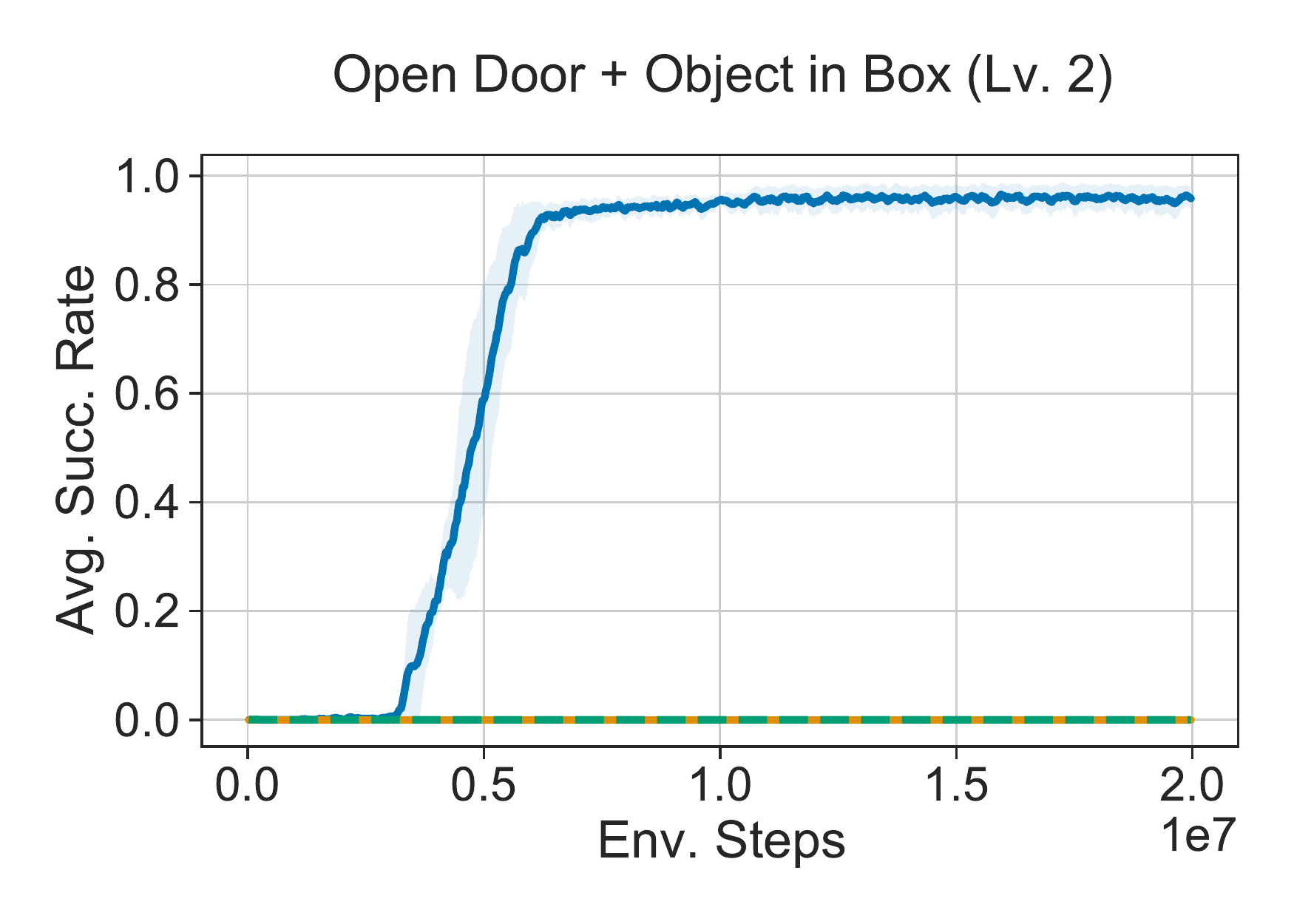}
&
\hspace{-0.6cm}\includegraphics[width=0.33\textwidth]{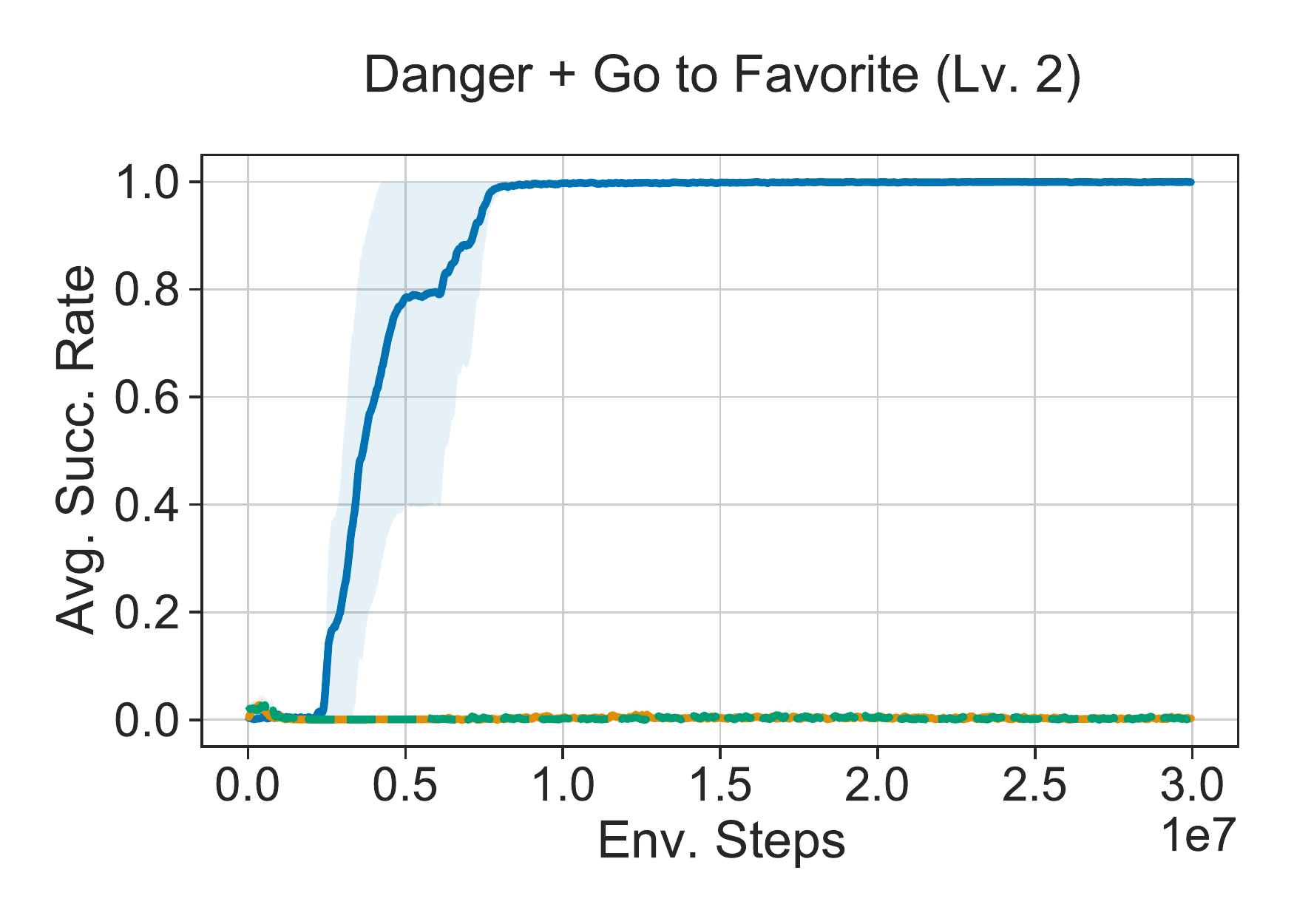}
&
\hspace{-0.6cm}\includegraphics[width=0.33\textwidth]{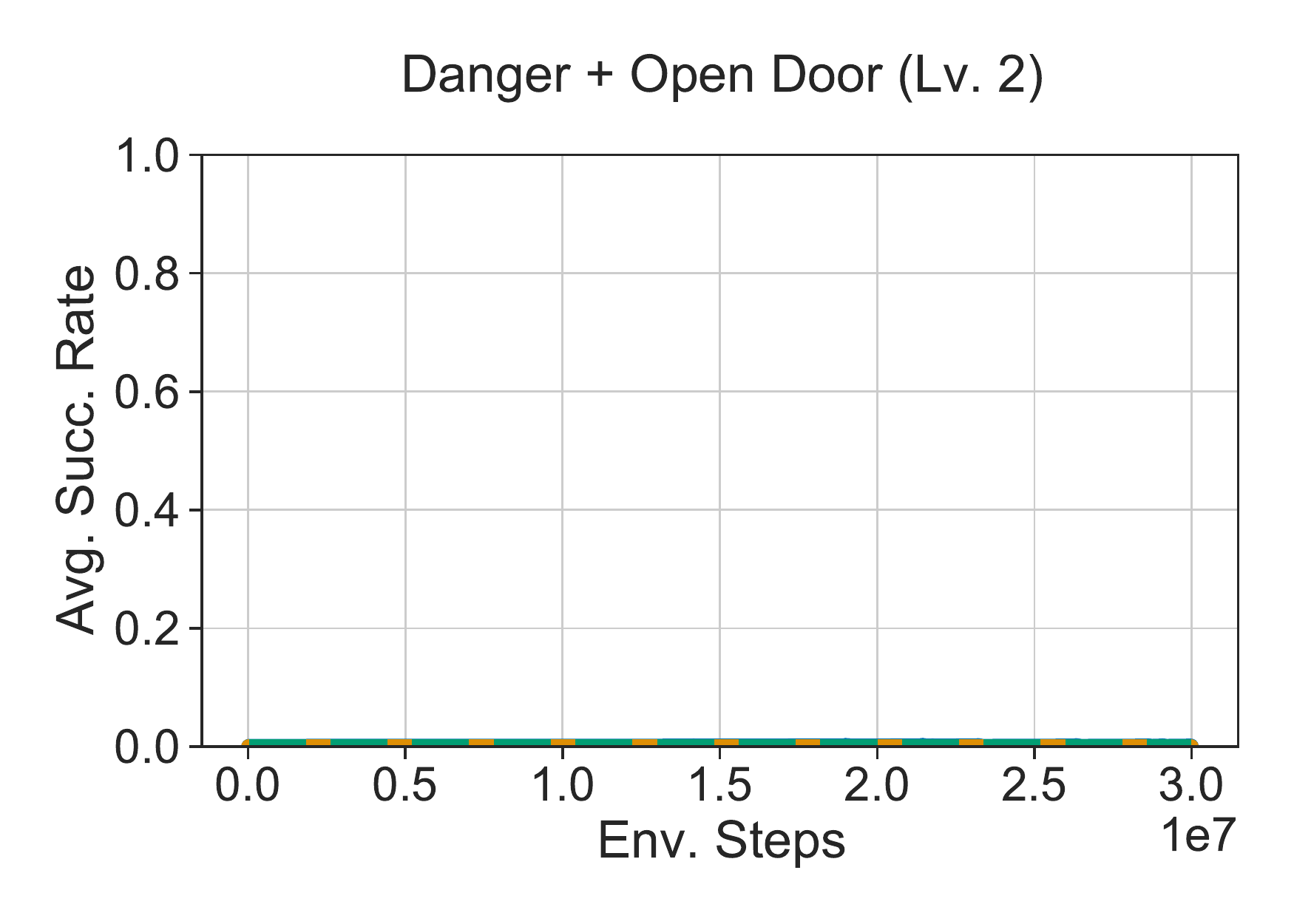}\\

\includegraphics[width=0.33\textwidth]{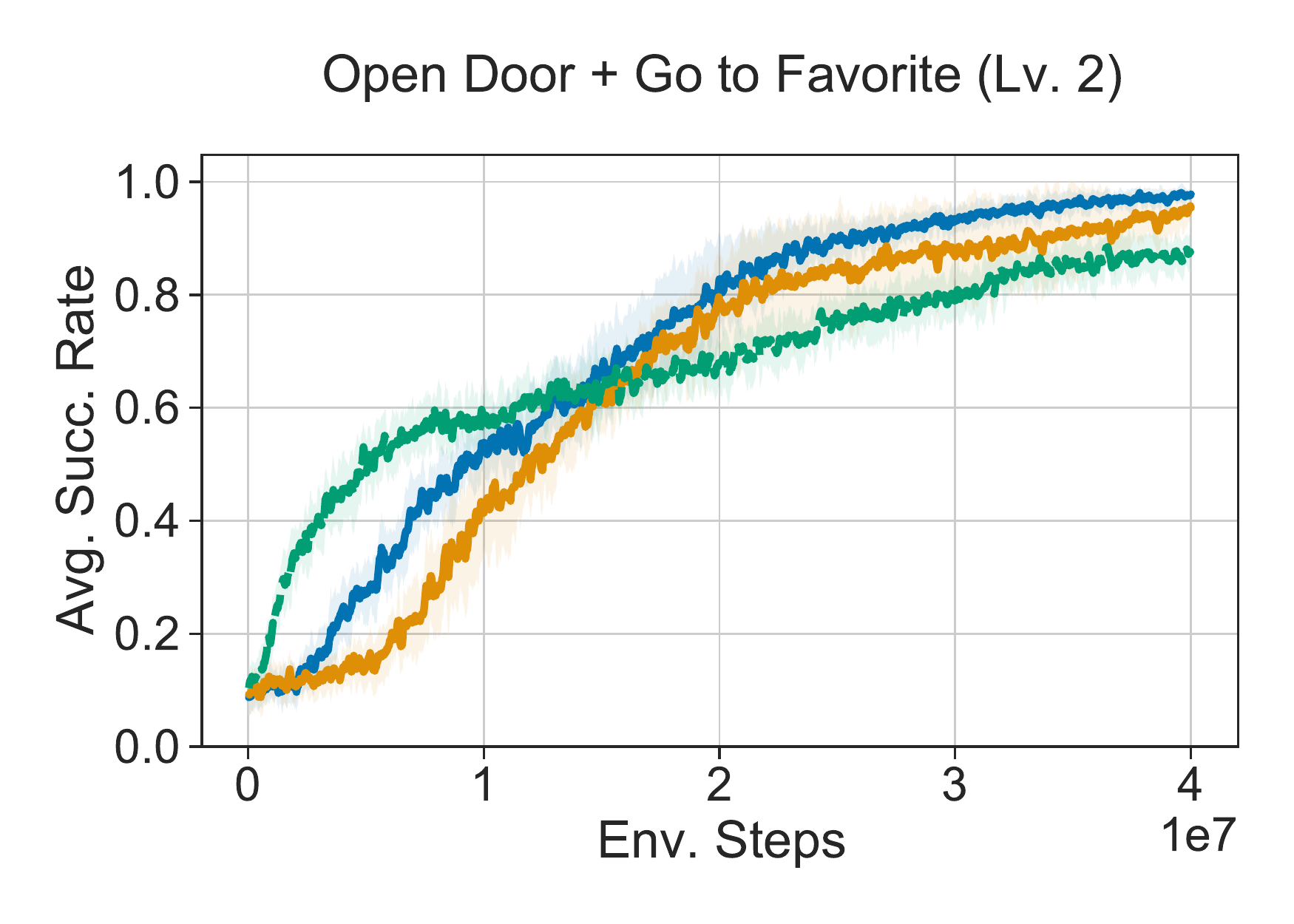}
&
\includegraphics[width=0.33\textwidth]{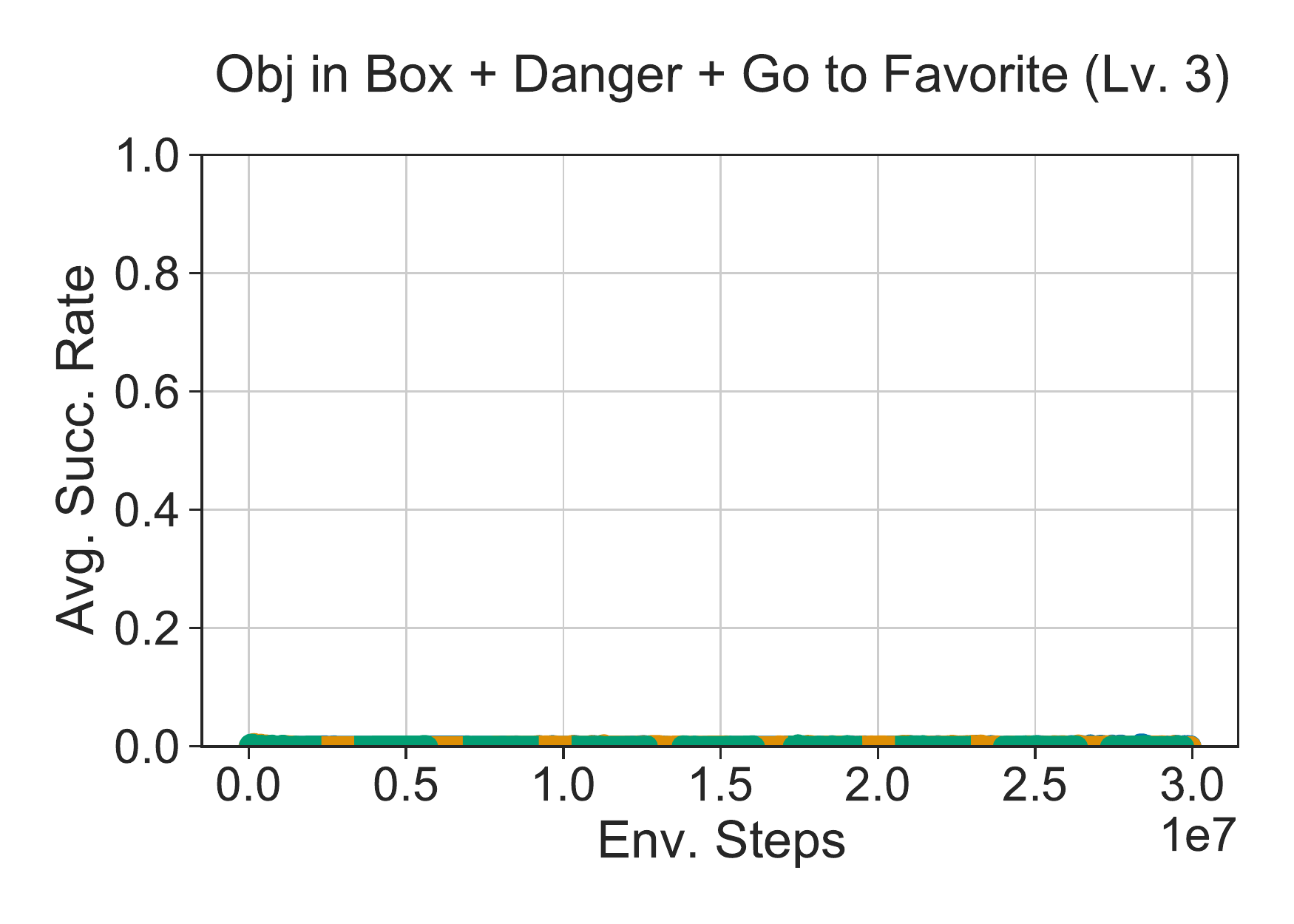}
&
\includegraphics[width=0.33\textwidth]{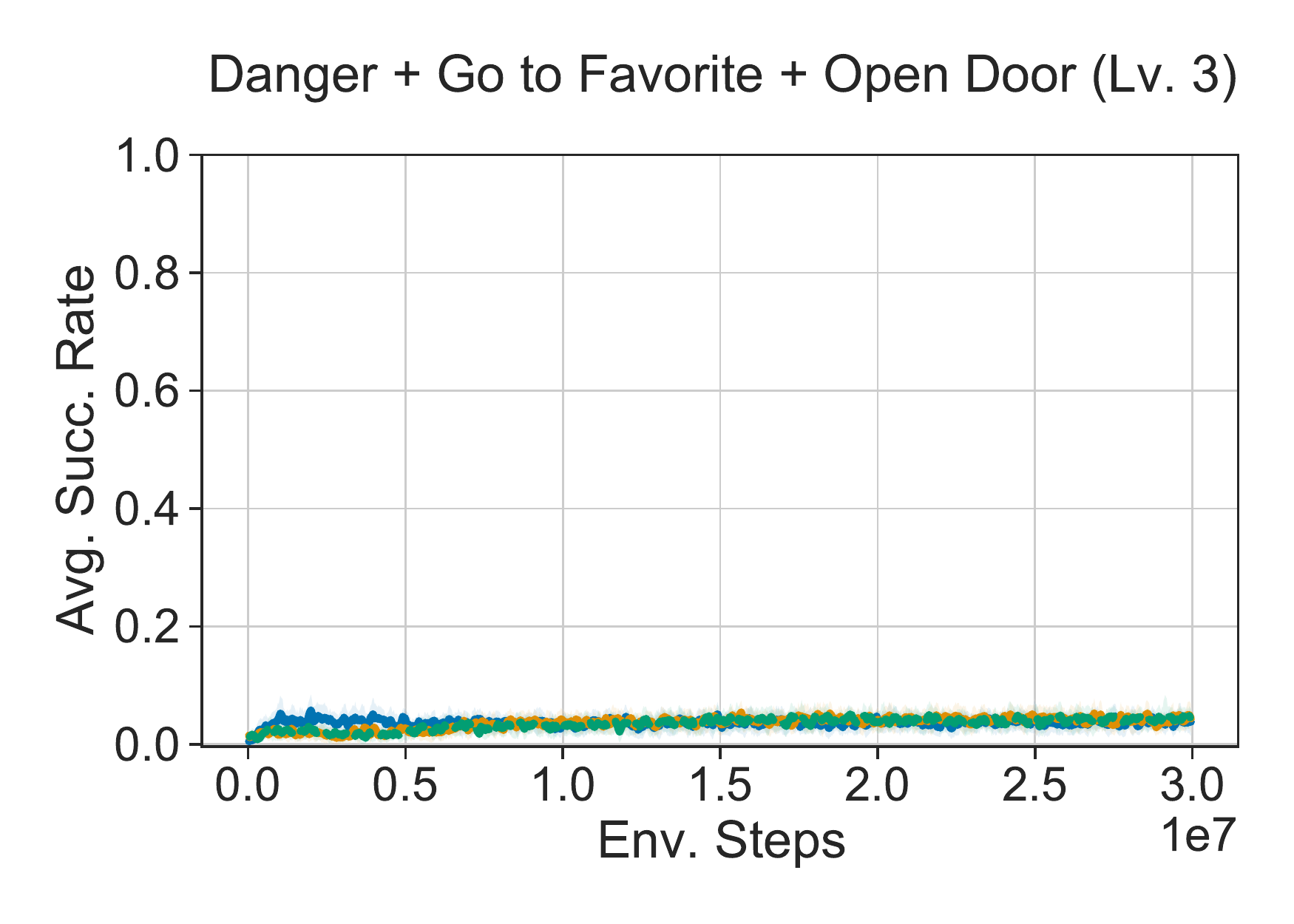}
\\
\includegraphics[width=0.33\textwidth]{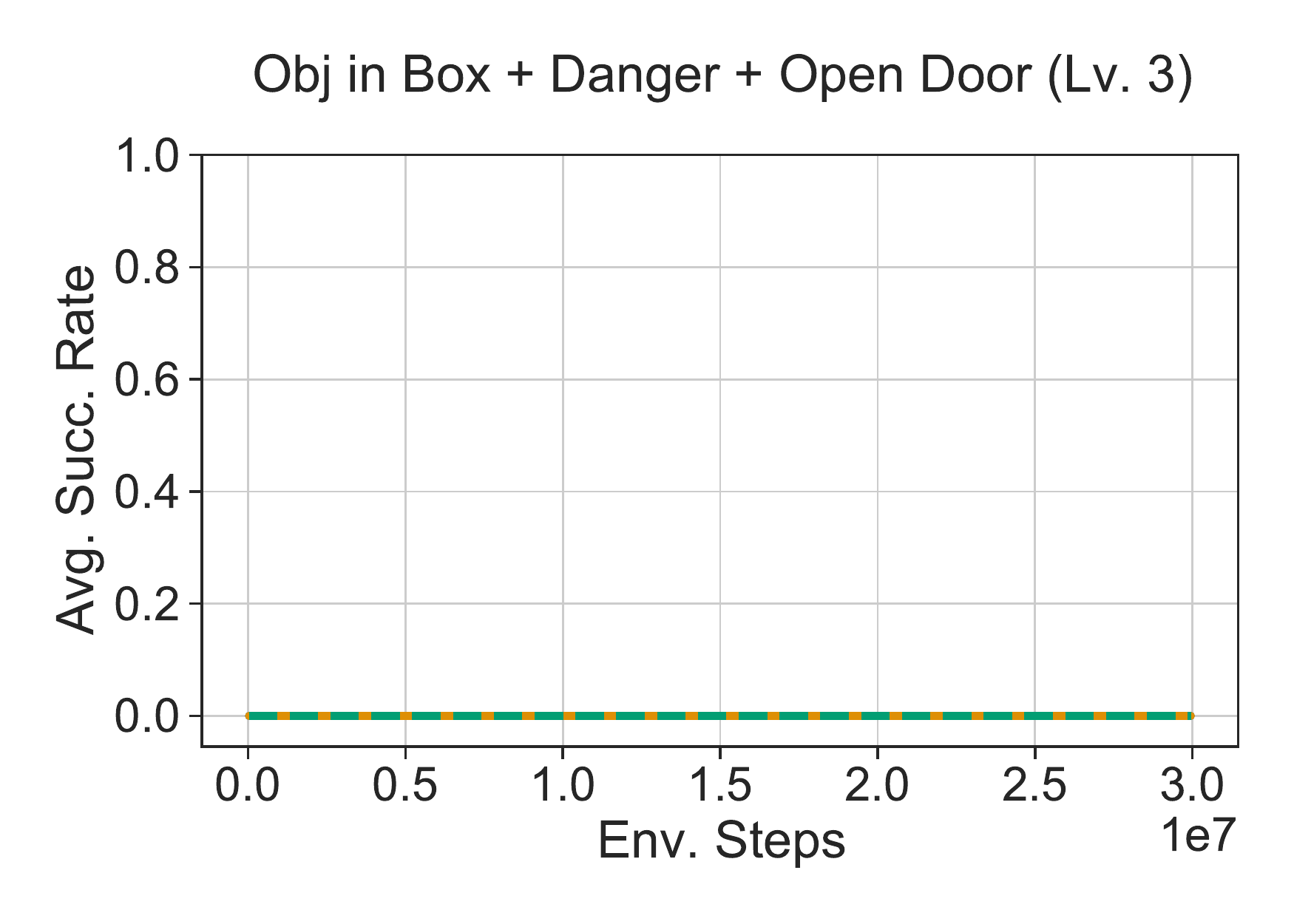}
&

\includegraphics[width=0.33\textwidth]{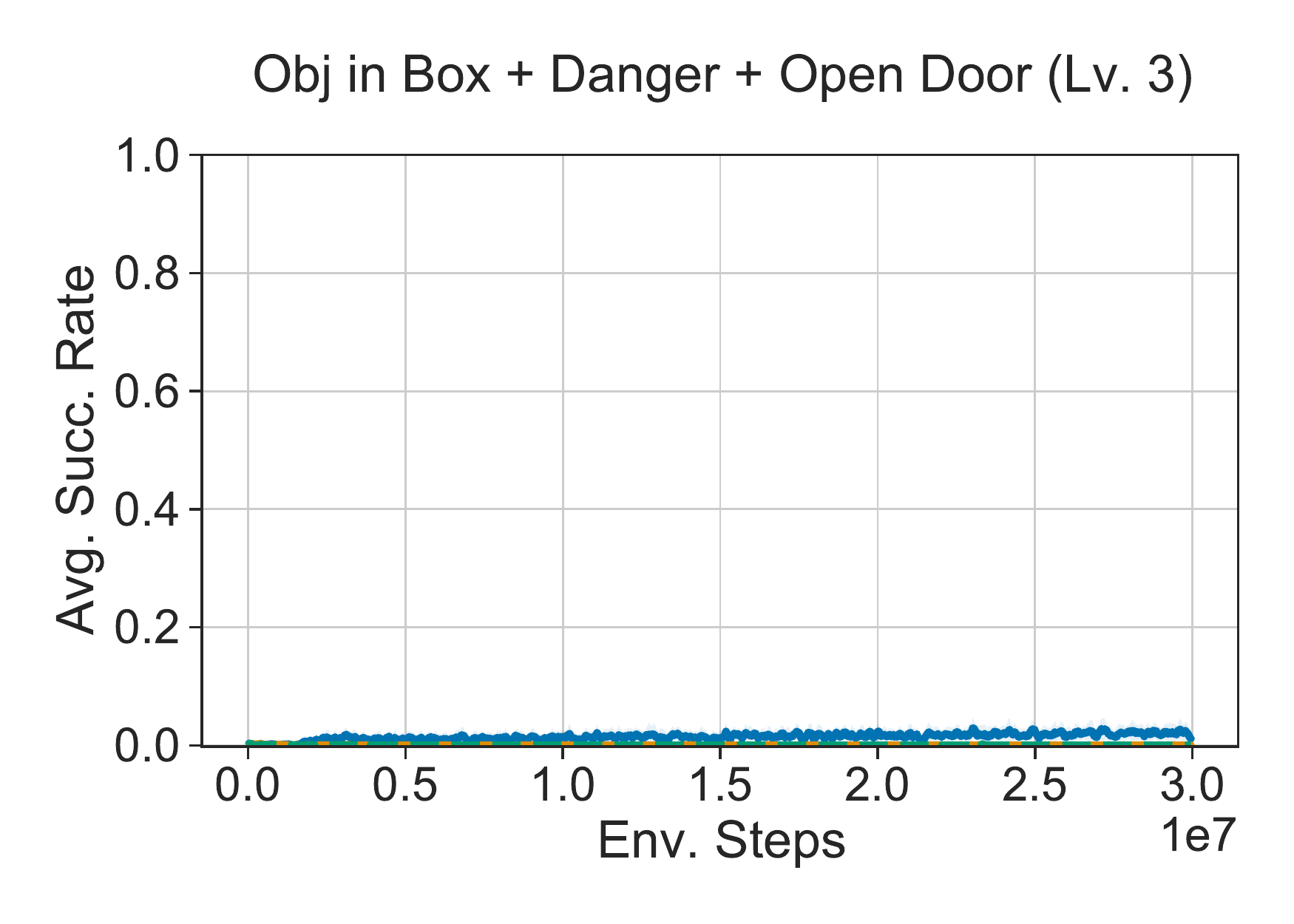}
&
\includegraphics[width=0.33\textwidth]{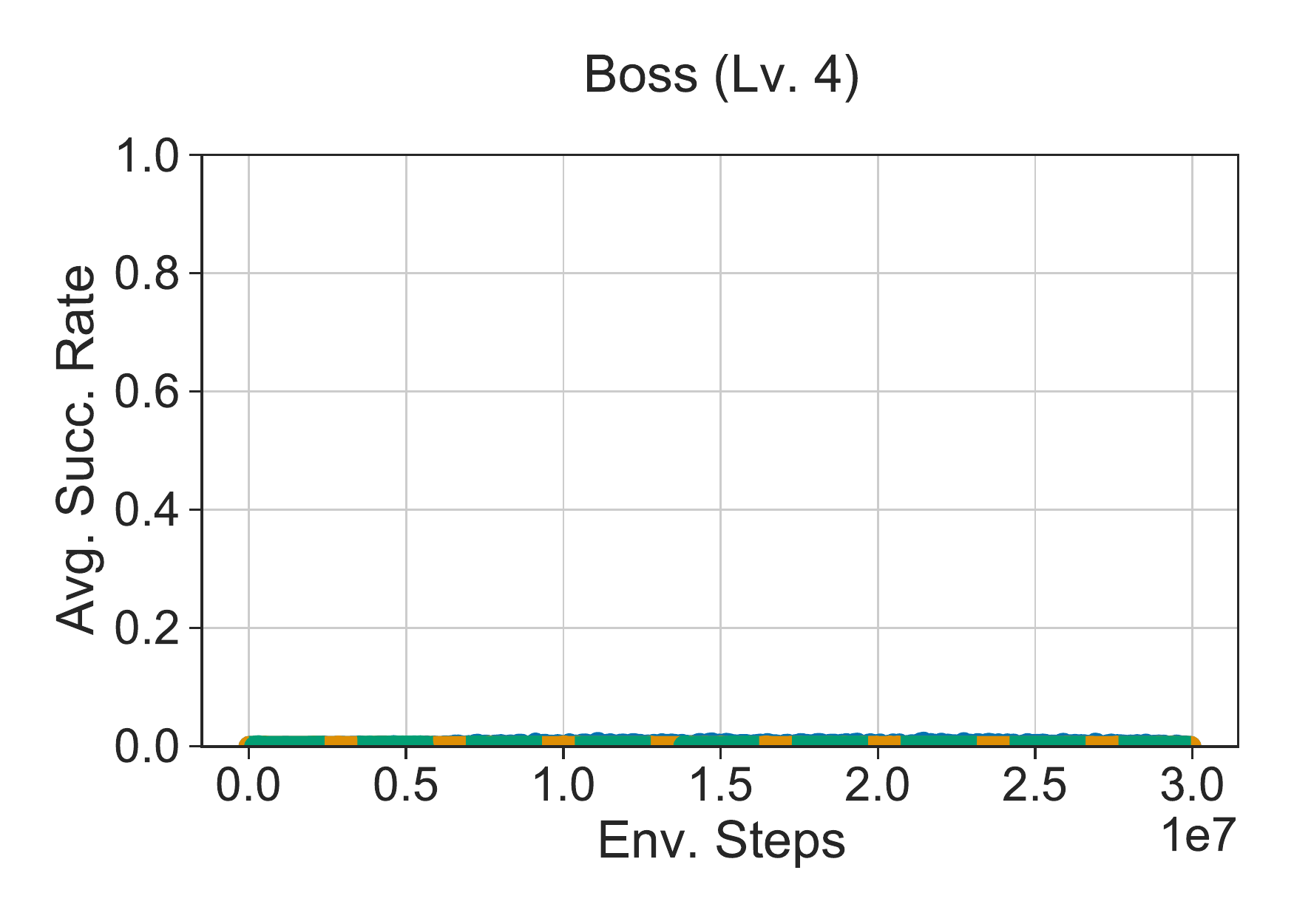}

\\
\multicolumn{3}{c}{\includegraphics[width=0.5\textwidth]{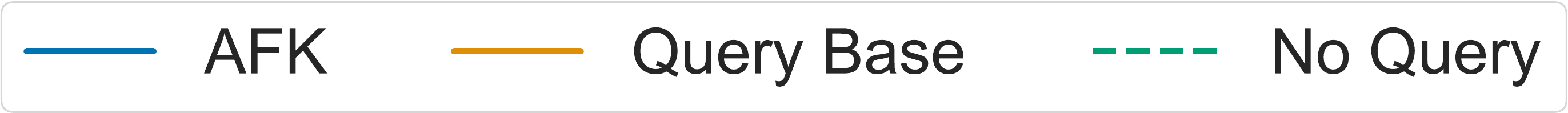}}

\end{tabular}

\caption{Success rate of \afk and baselines on \qbai.}
\label{fig:appd_plt_qbai_succ}

\end{figure*}

\begin{figure*}[t]
\centering
\begin{tabular}{ccc}

\includegraphics[width=0.33\textwidth]{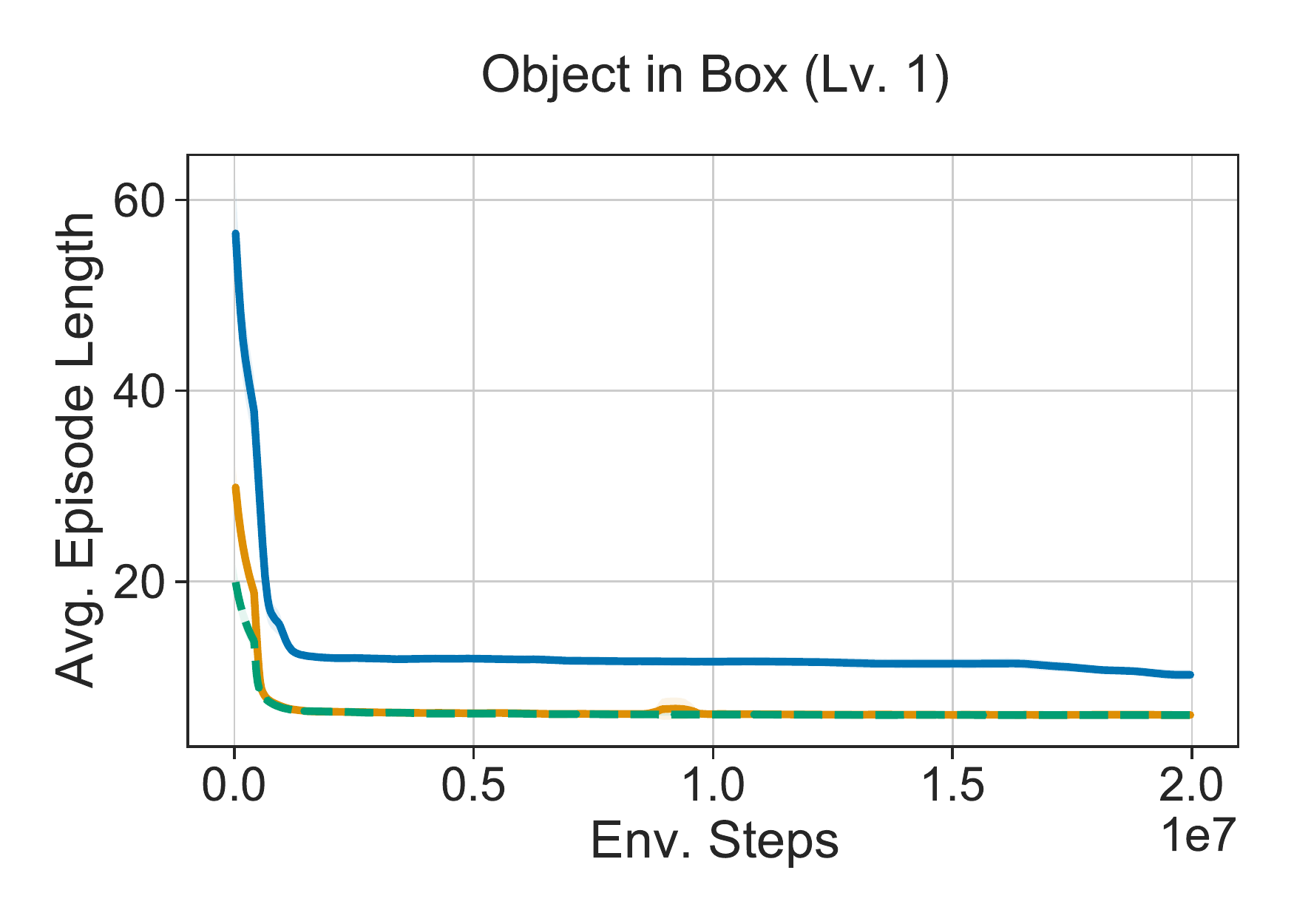}
&
\hspace{-0.6cm}\includegraphics[width=0.33\textwidth]{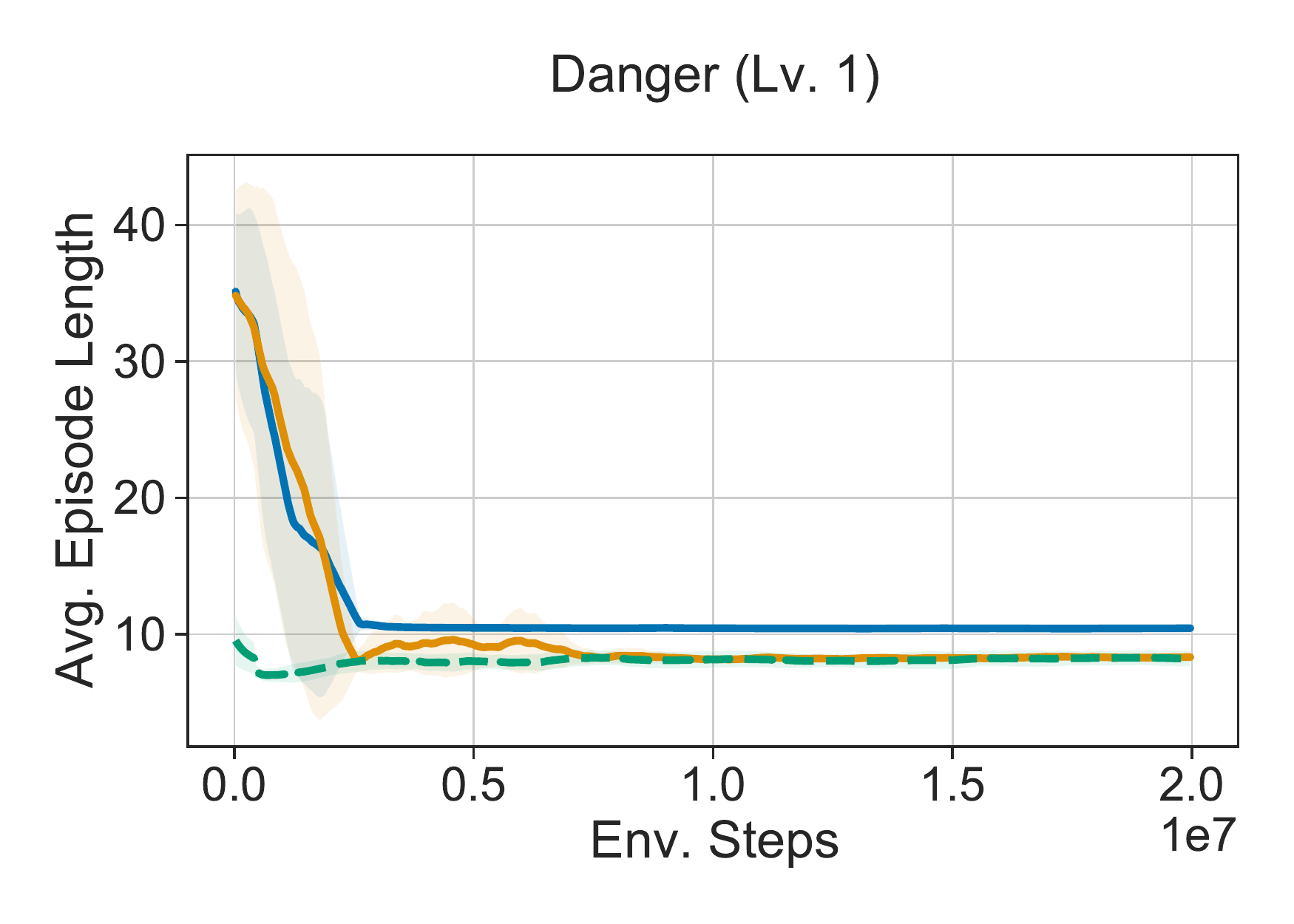}
&
\hspace{-0.6cm}\includegraphics[width=0.33\textwidth]{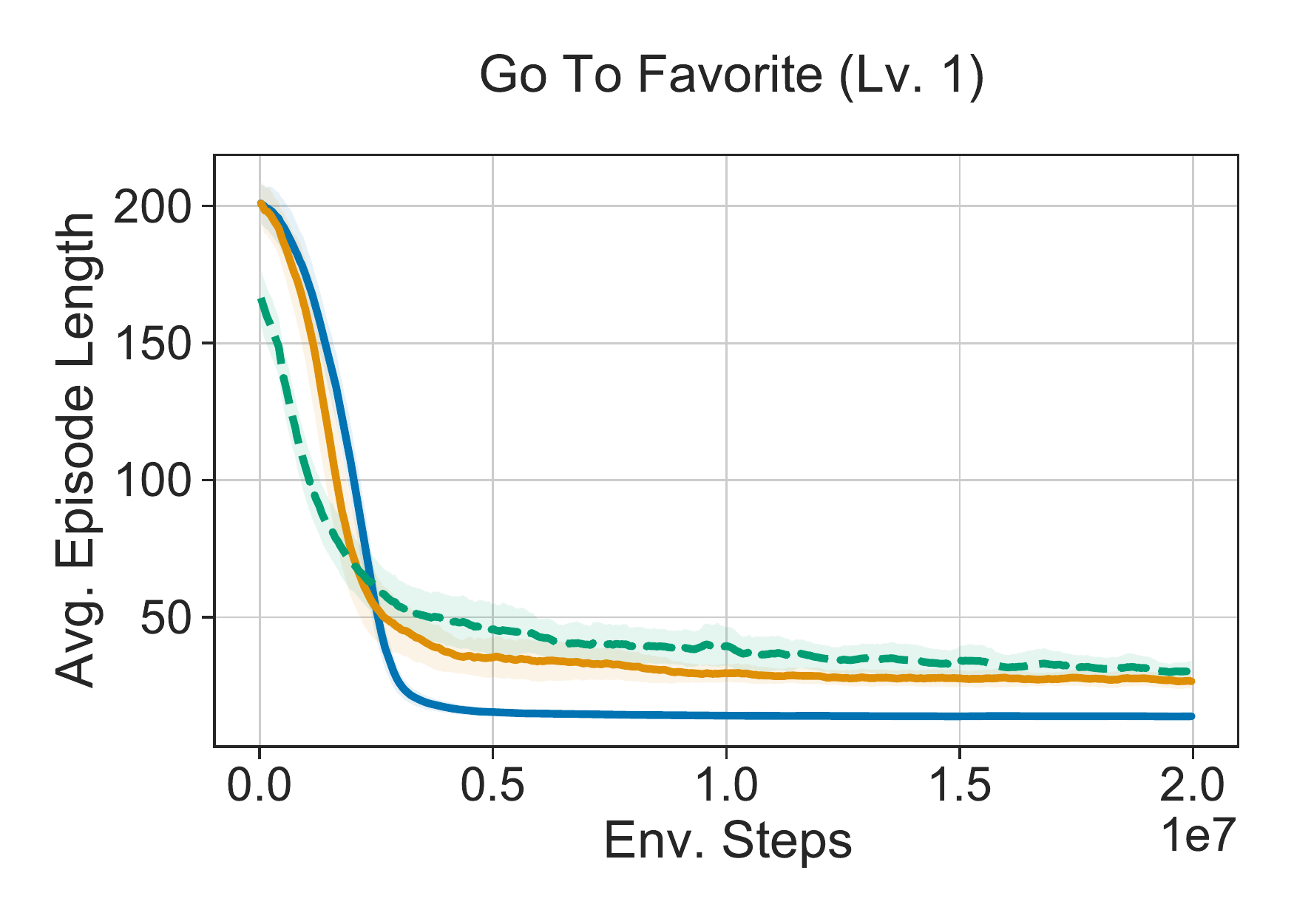}\\

\includegraphics[width=0.33\textwidth]{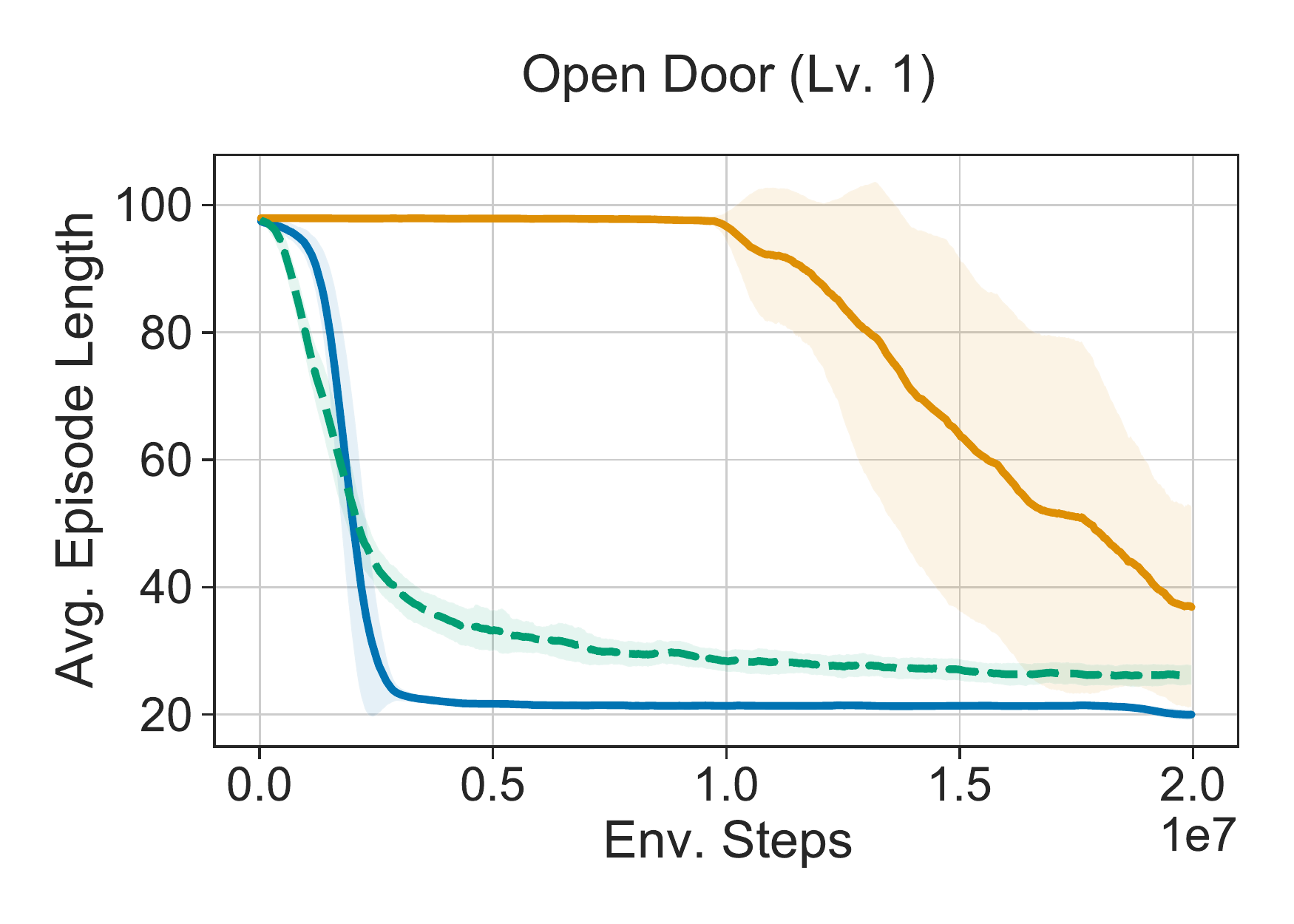}
&
\hspace{-0.6cm}\includegraphics[width=0.33\textwidth]{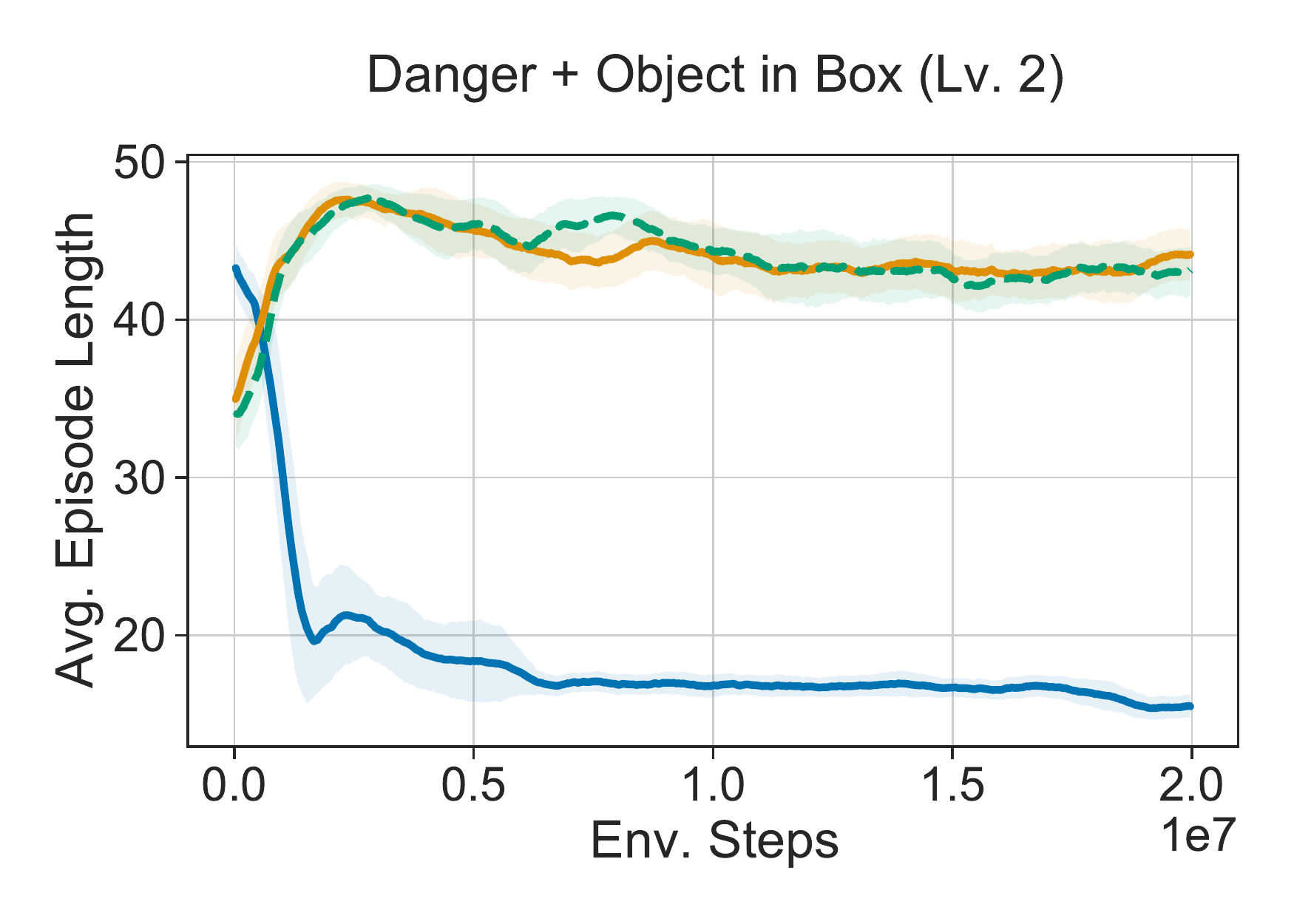}
&
\hspace{-0.6cm}\includegraphics[width=0.33\textwidth]{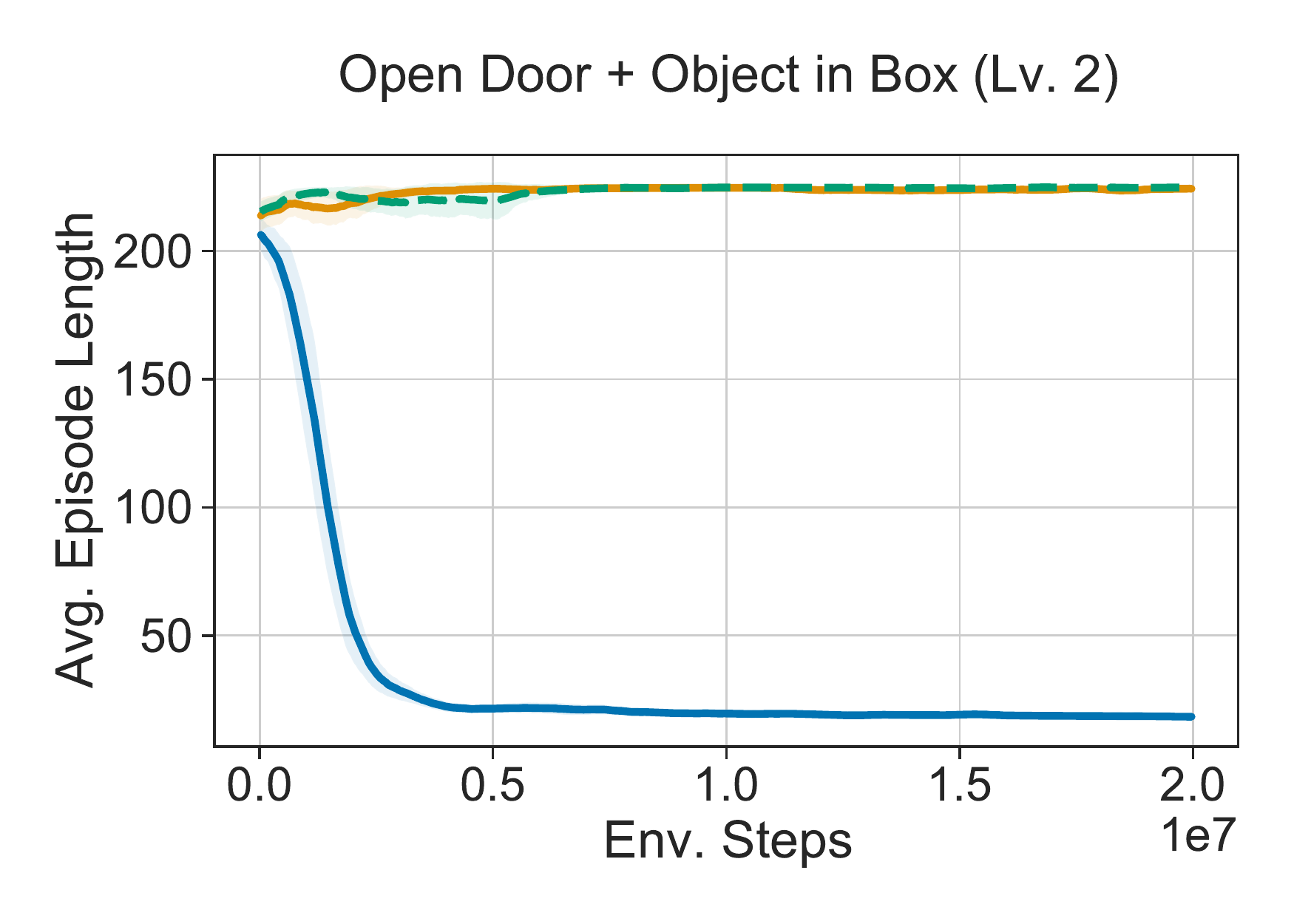}\\

\includegraphics[width=0.33\textwidth]{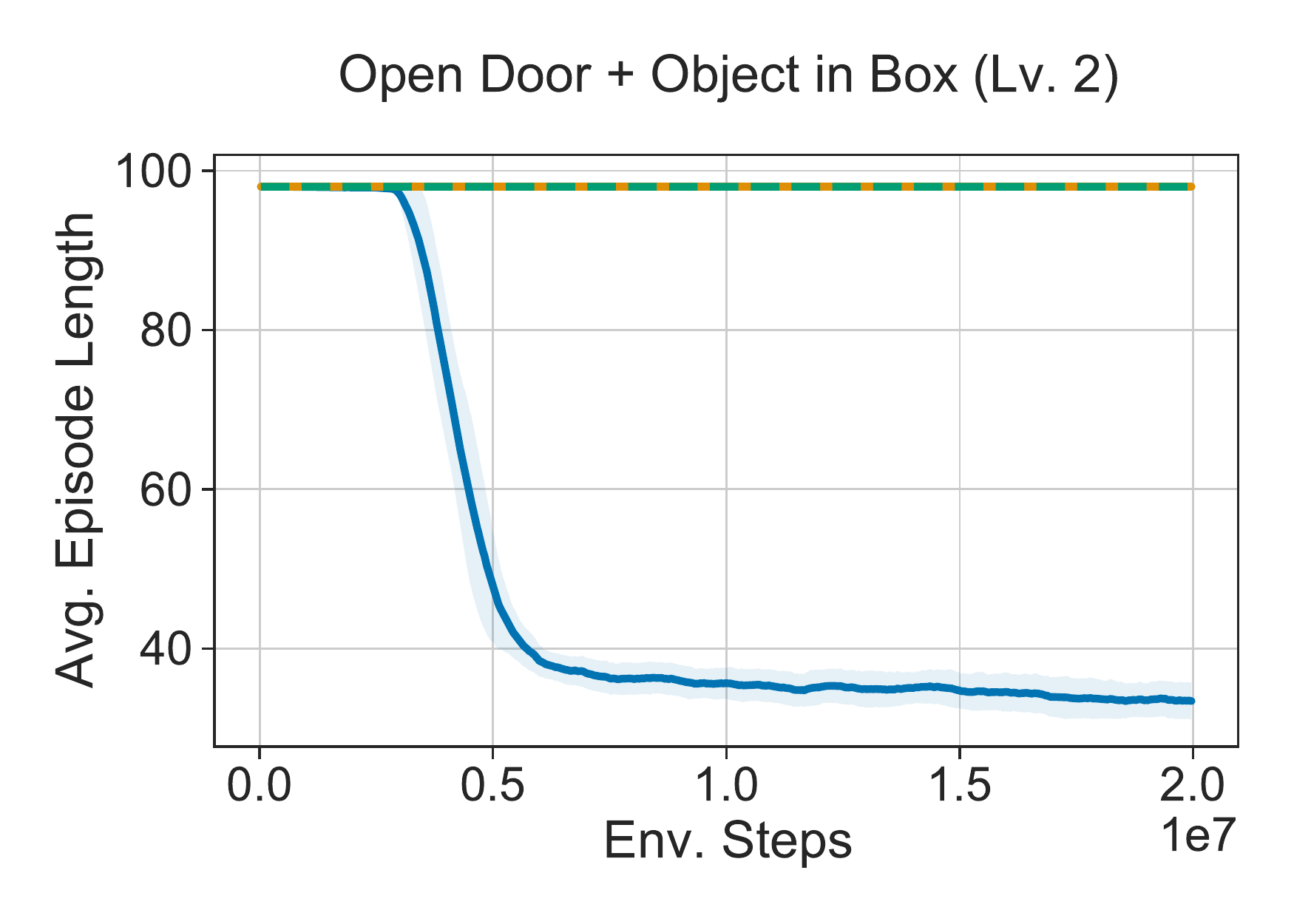}
&
\hspace{-0.6cm}\includegraphics[width=0.33\textwidth]{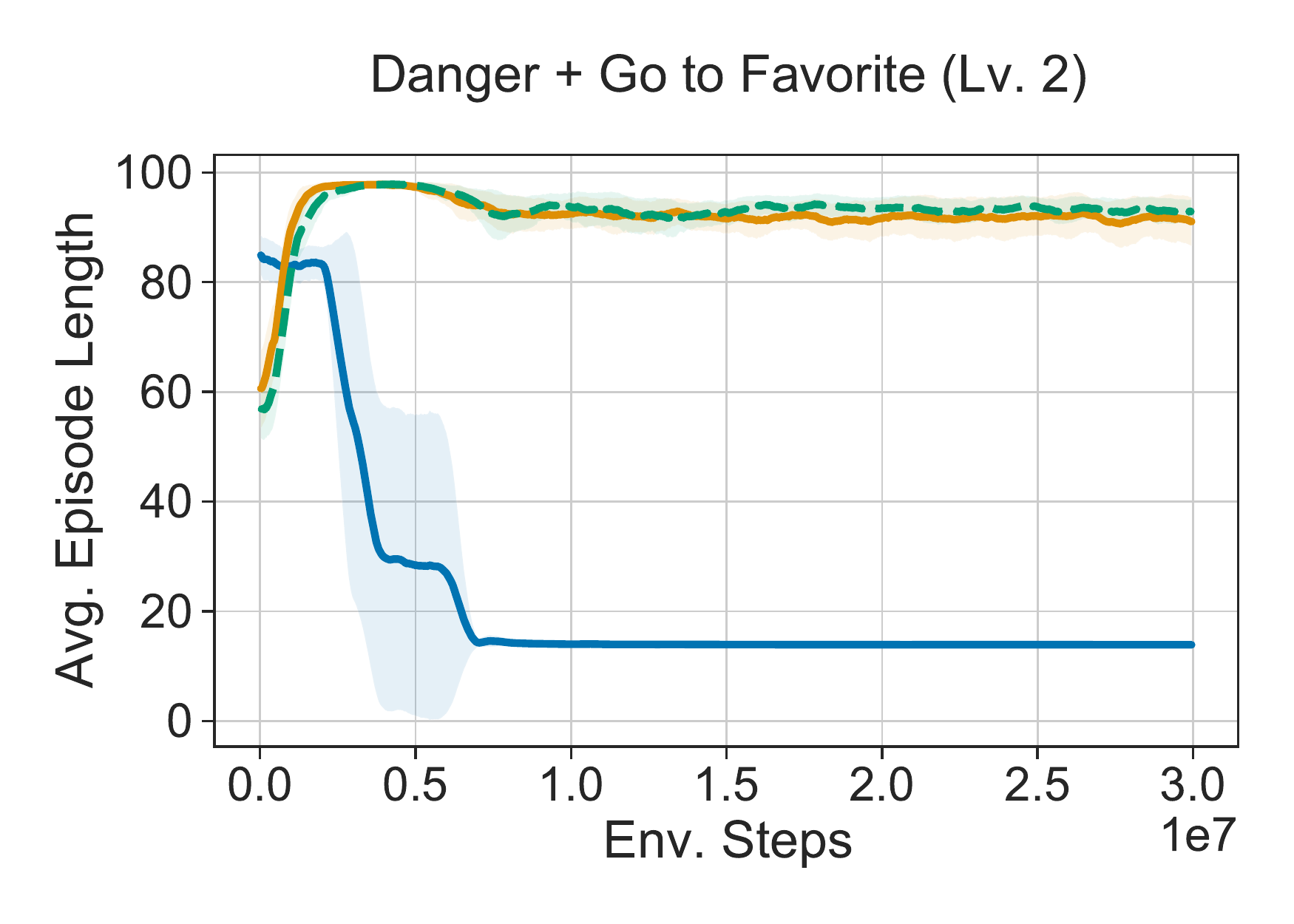}
&
\hspace{-0.6cm}\includegraphics[width=0.33\textwidth]{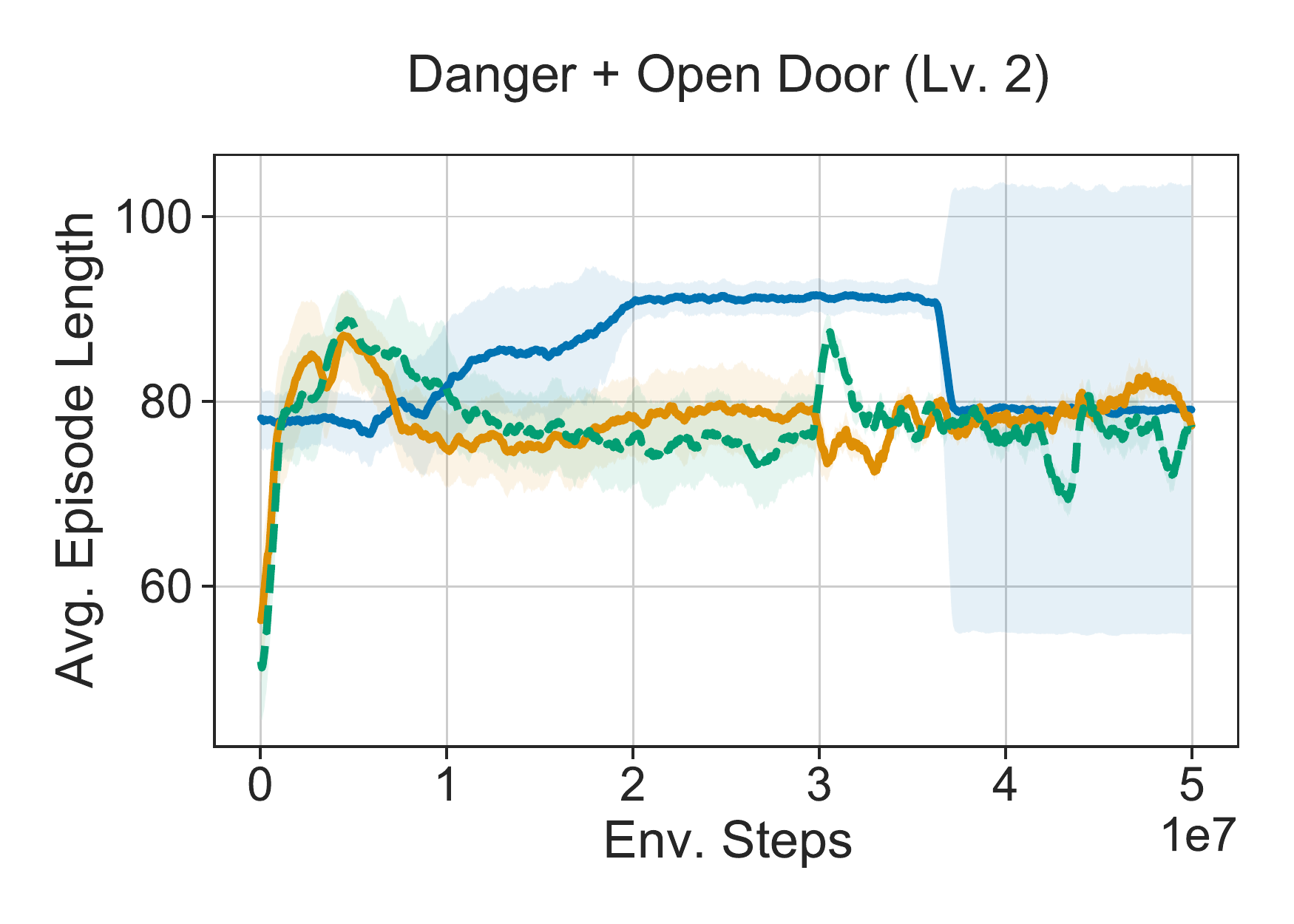}\\

\includegraphics[width=0.33\textwidth]{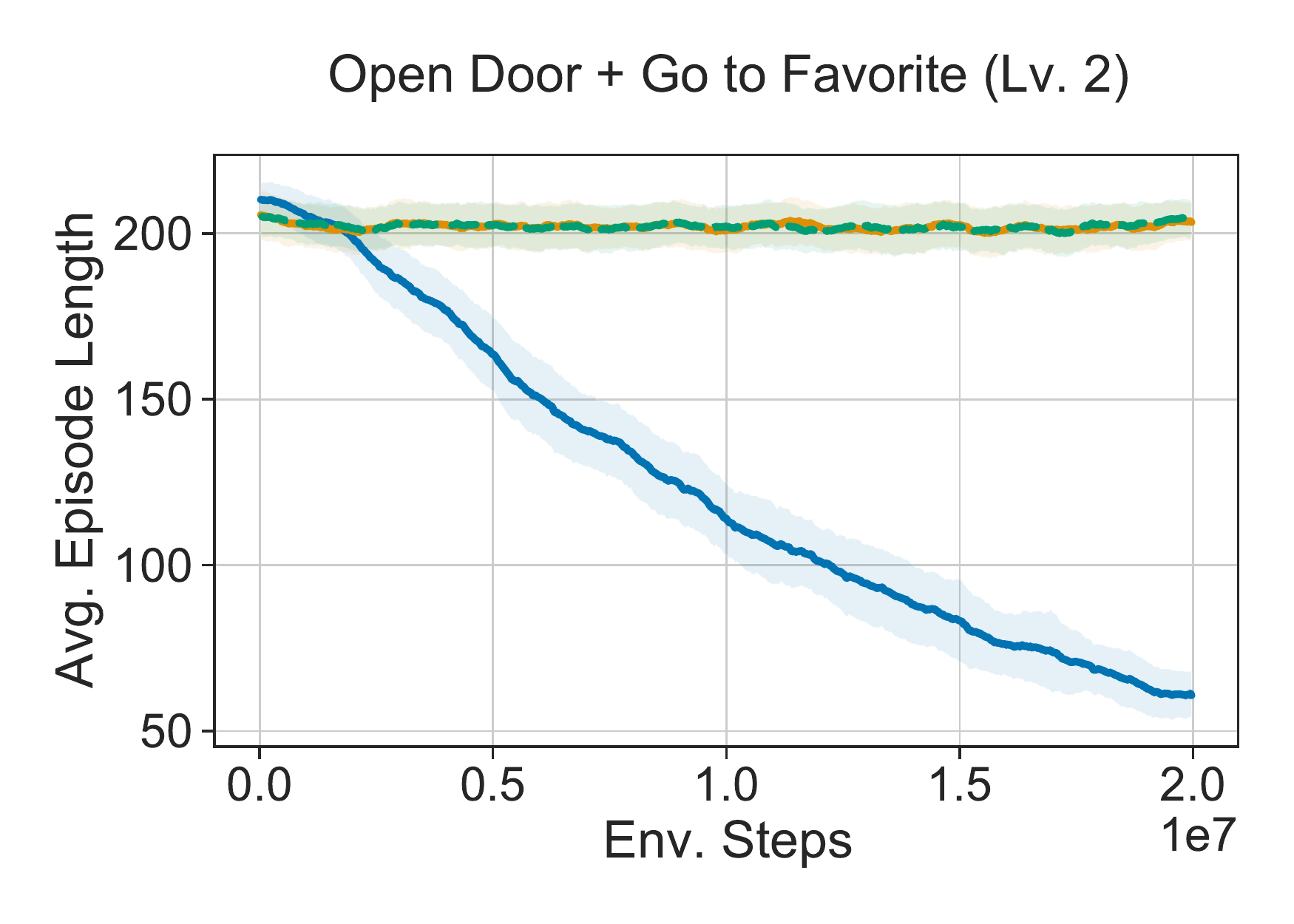}
&
\includegraphics[width=0.33\textwidth]{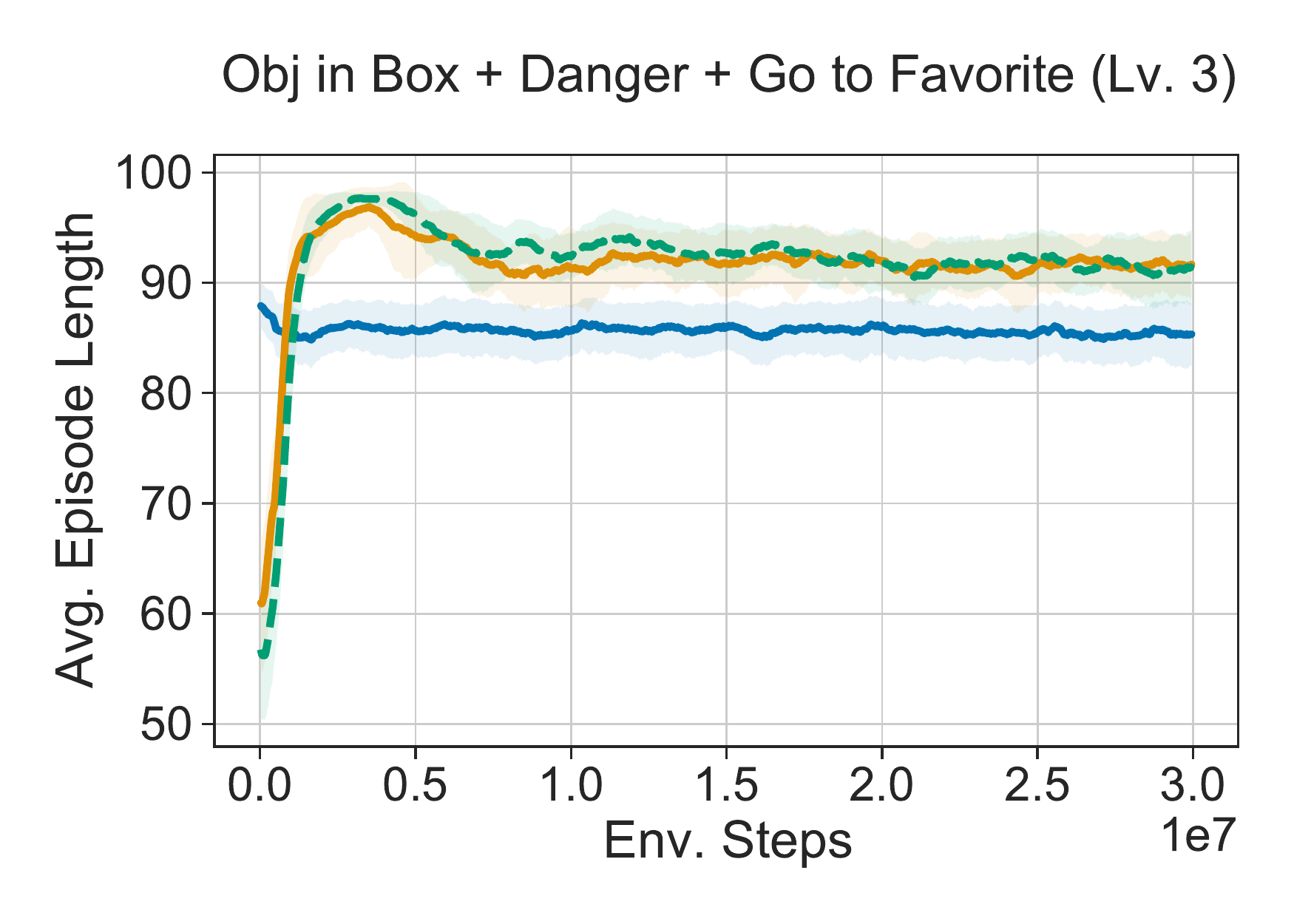}
&
\includegraphics[width=0.33\textwidth]{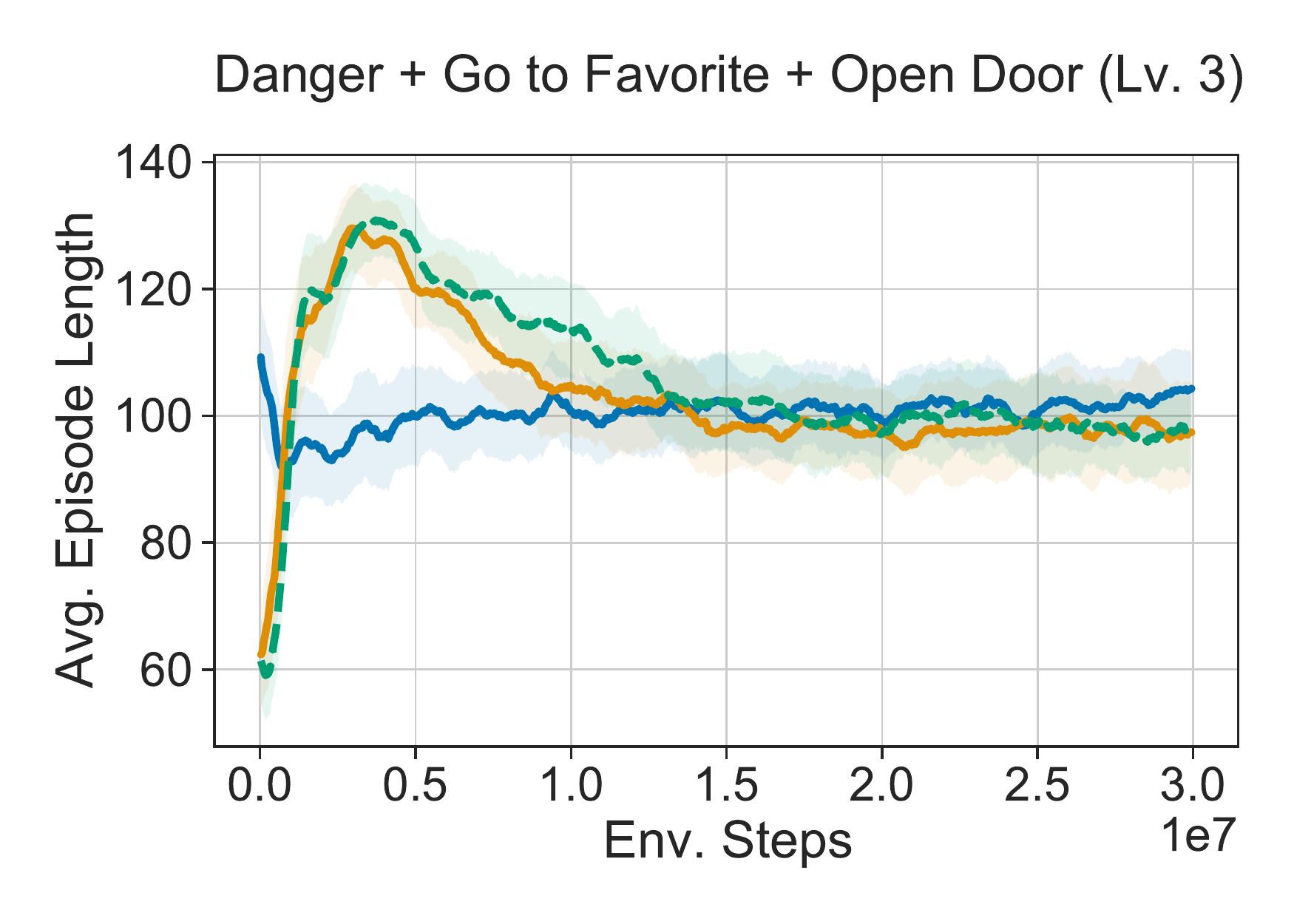}
\\
\includegraphics[width=0.33\textwidth]{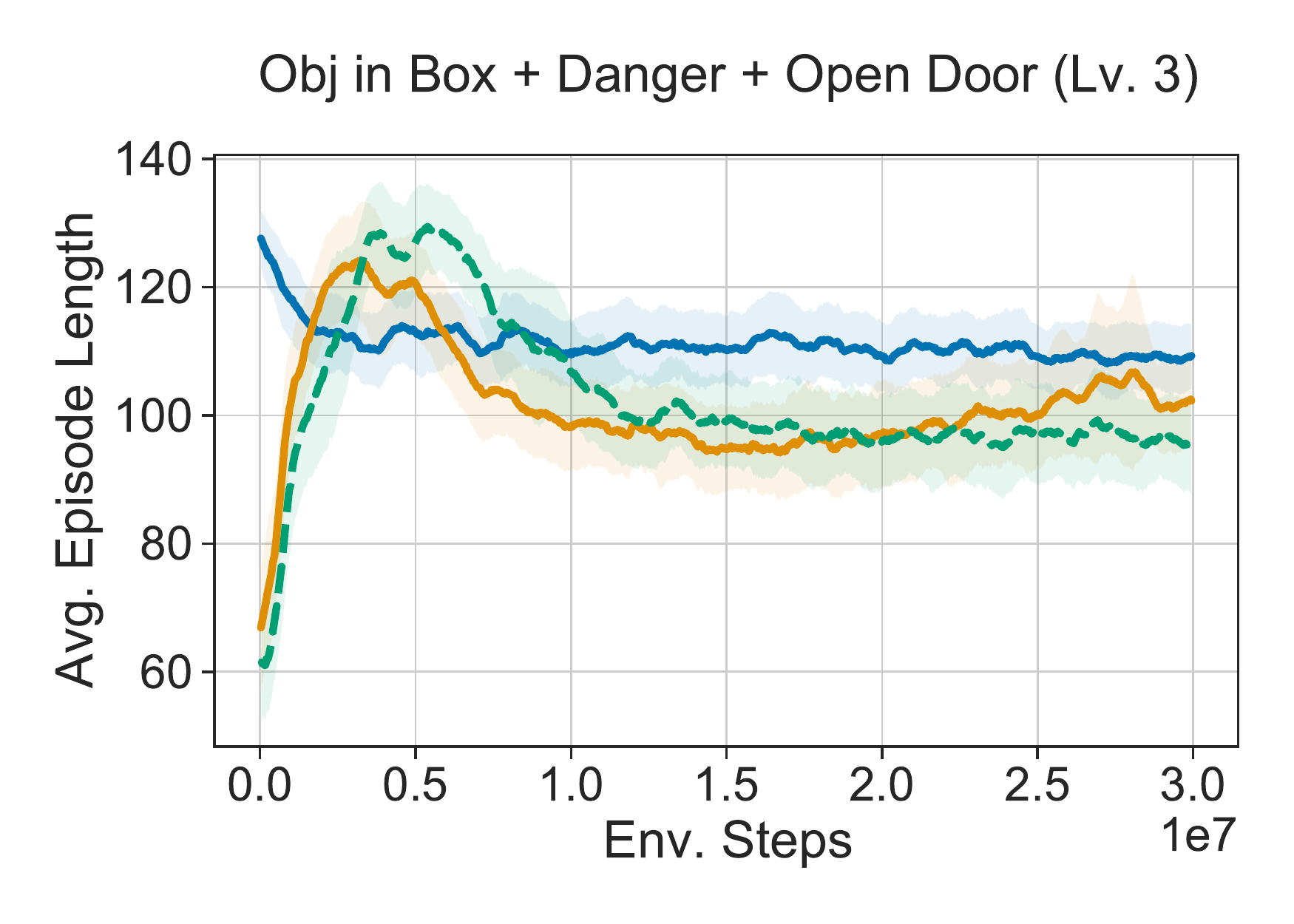}
&

\includegraphics[width=0.33\textwidth]{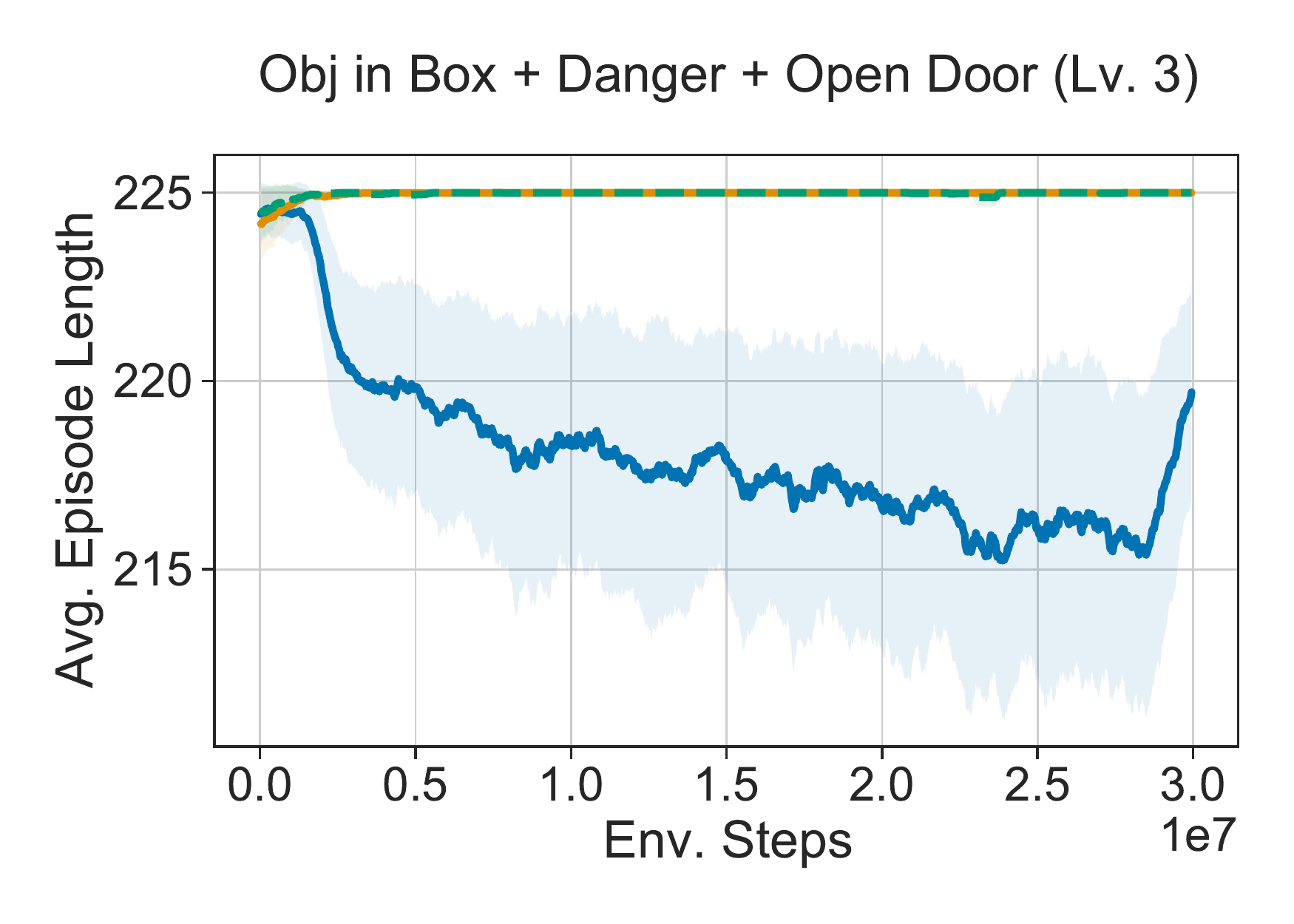}
&
\includegraphics[width=0.33\textwidth]{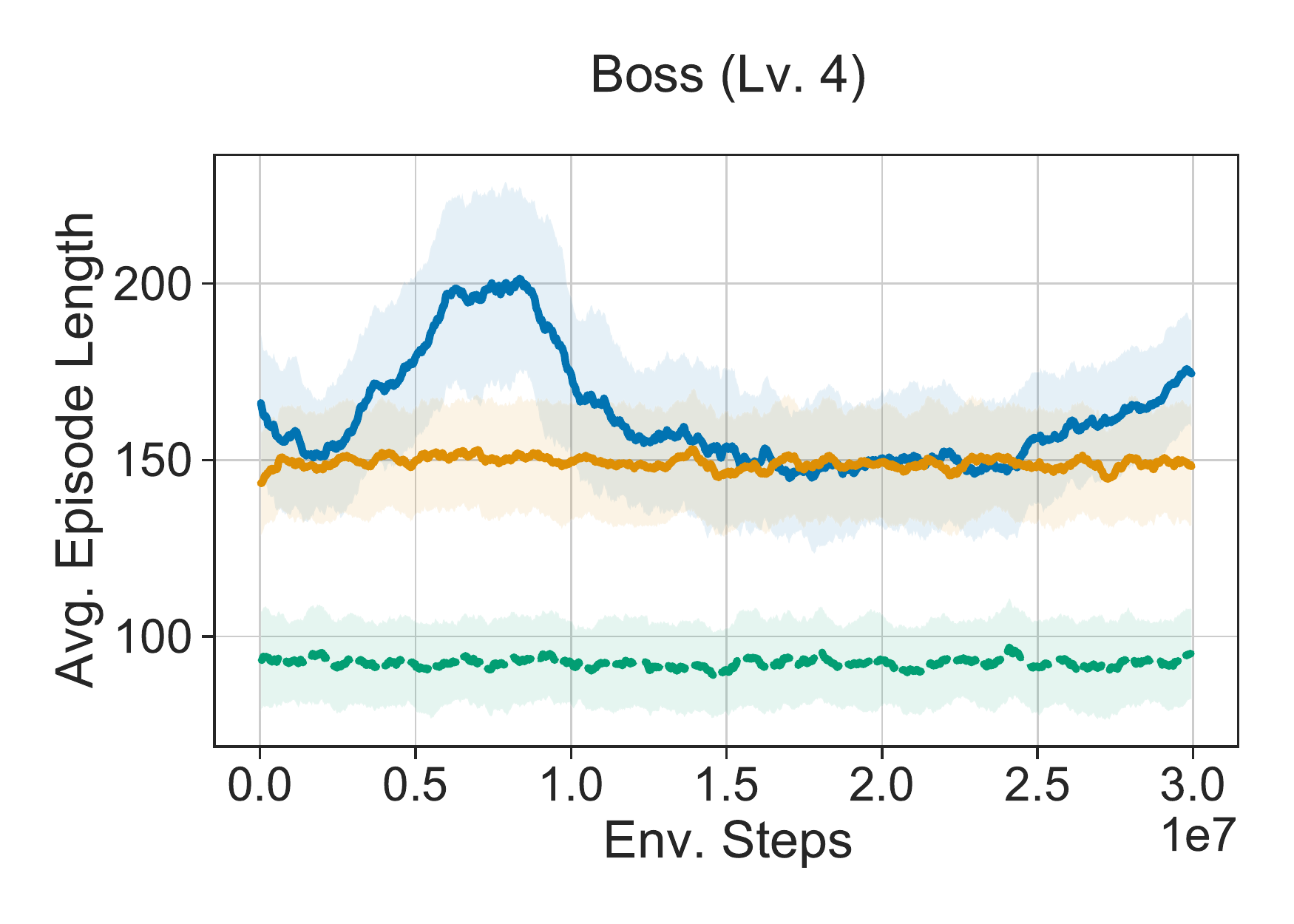}

\\
\multicolumn{3}{c}{\includegraphics[width=0.5\textwidth]{figures/legend.pdf}}

\end{tabular}

\caption{Episode length of \afk and baselines on \qbai.}
\label{fig:appd_plt_qbai_len}

\end{figure*}

\begin{figure*}[t]
\centering

\begin{tabular}{cc}

\includegraphics[width=0.33\textwidth]{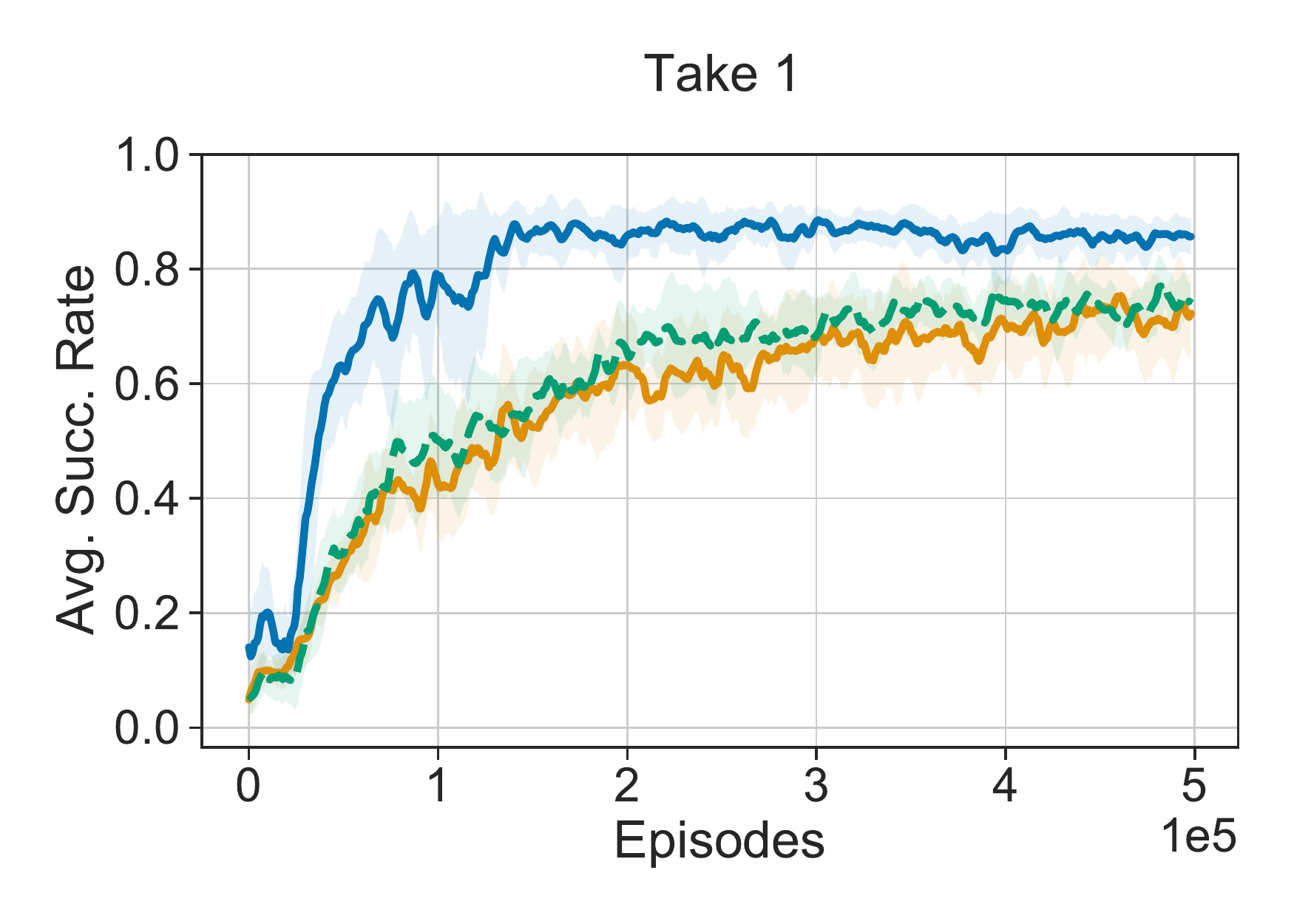}
&
\hspace{-0.6cm}\includegraphics[width=0.33\textwidth]{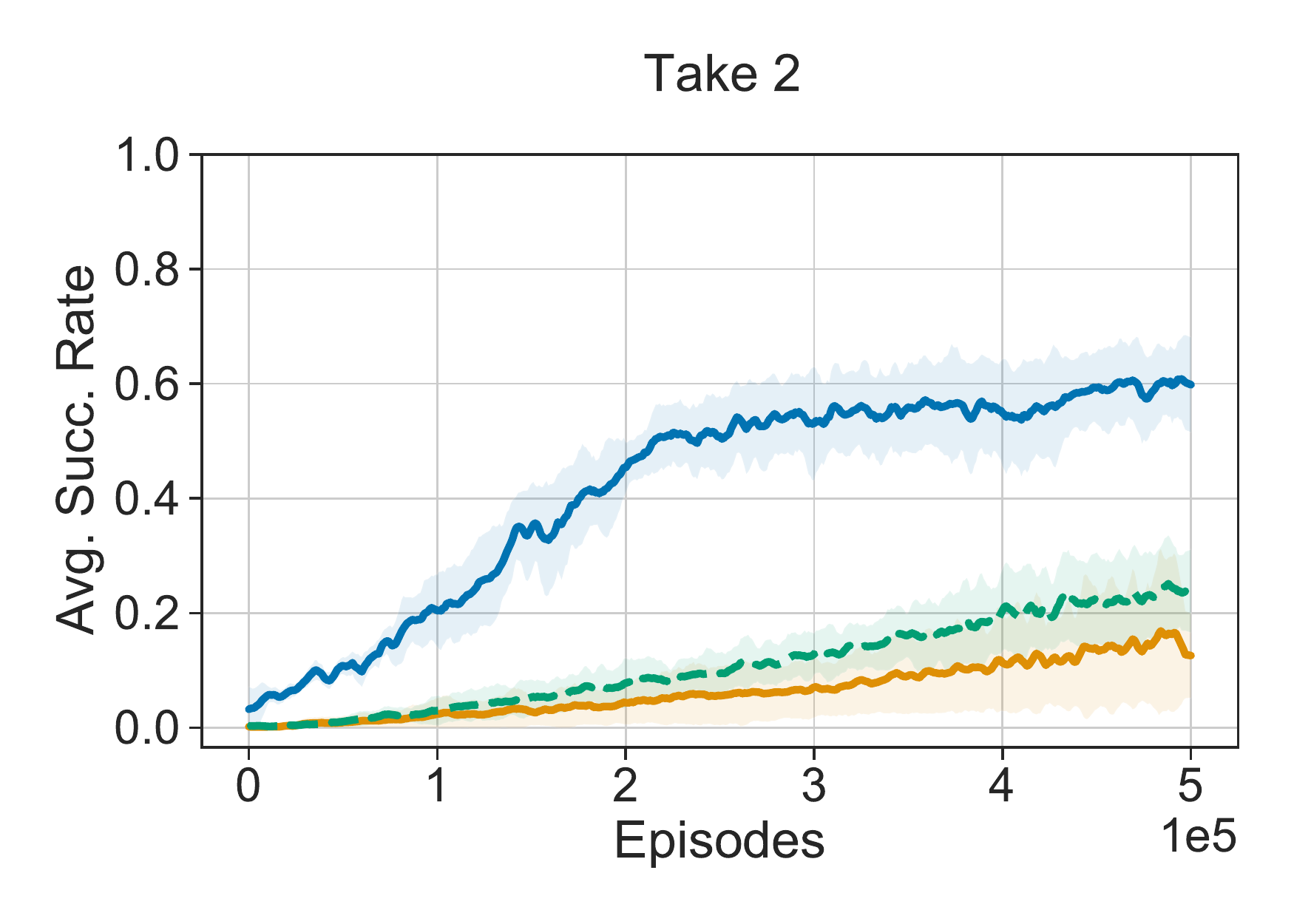}\\

\includegraphics[width=0.33\textwidth]{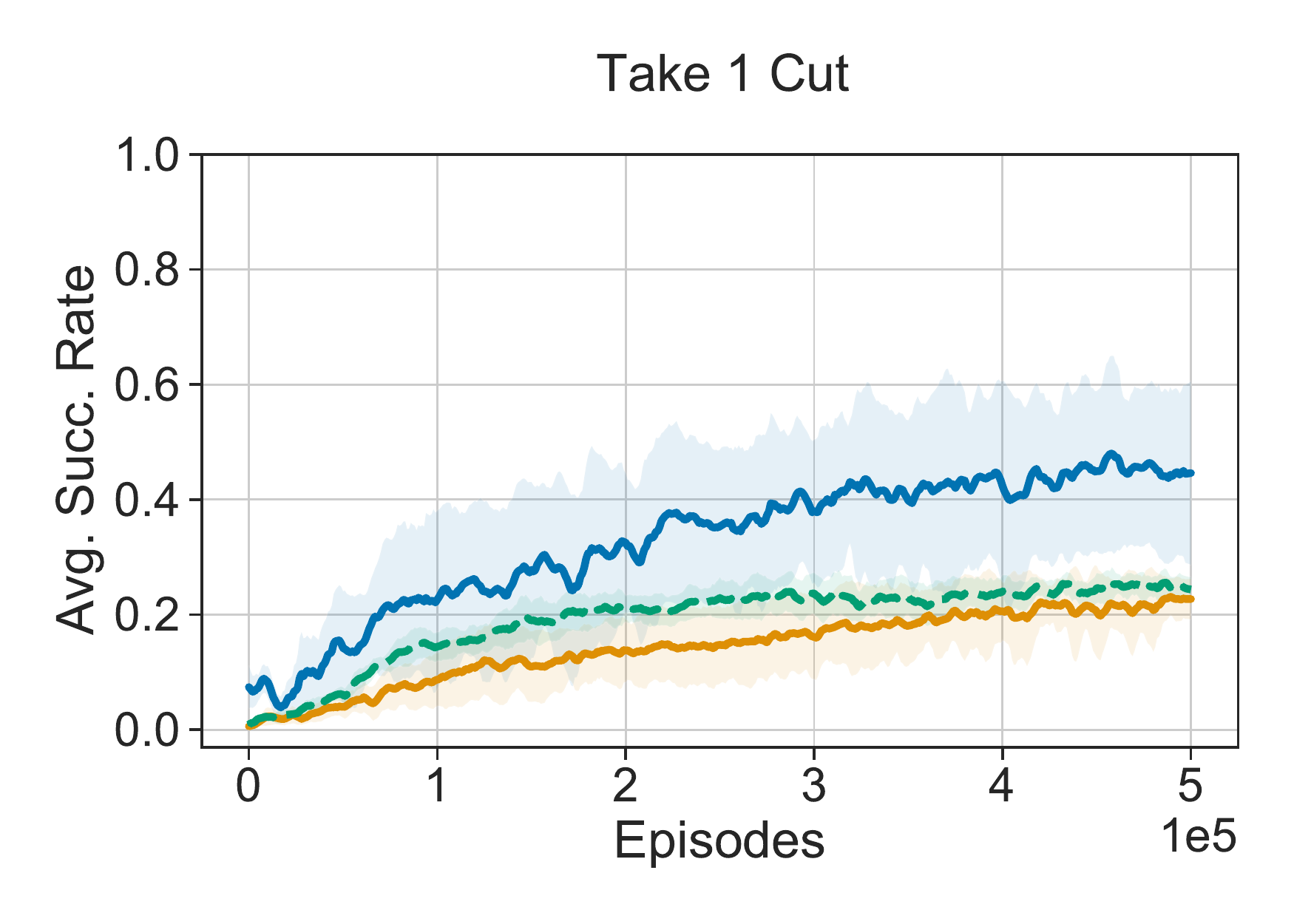}
&
\hspace{-0.6cm}\includegraphics[width=0.33\textwidth]{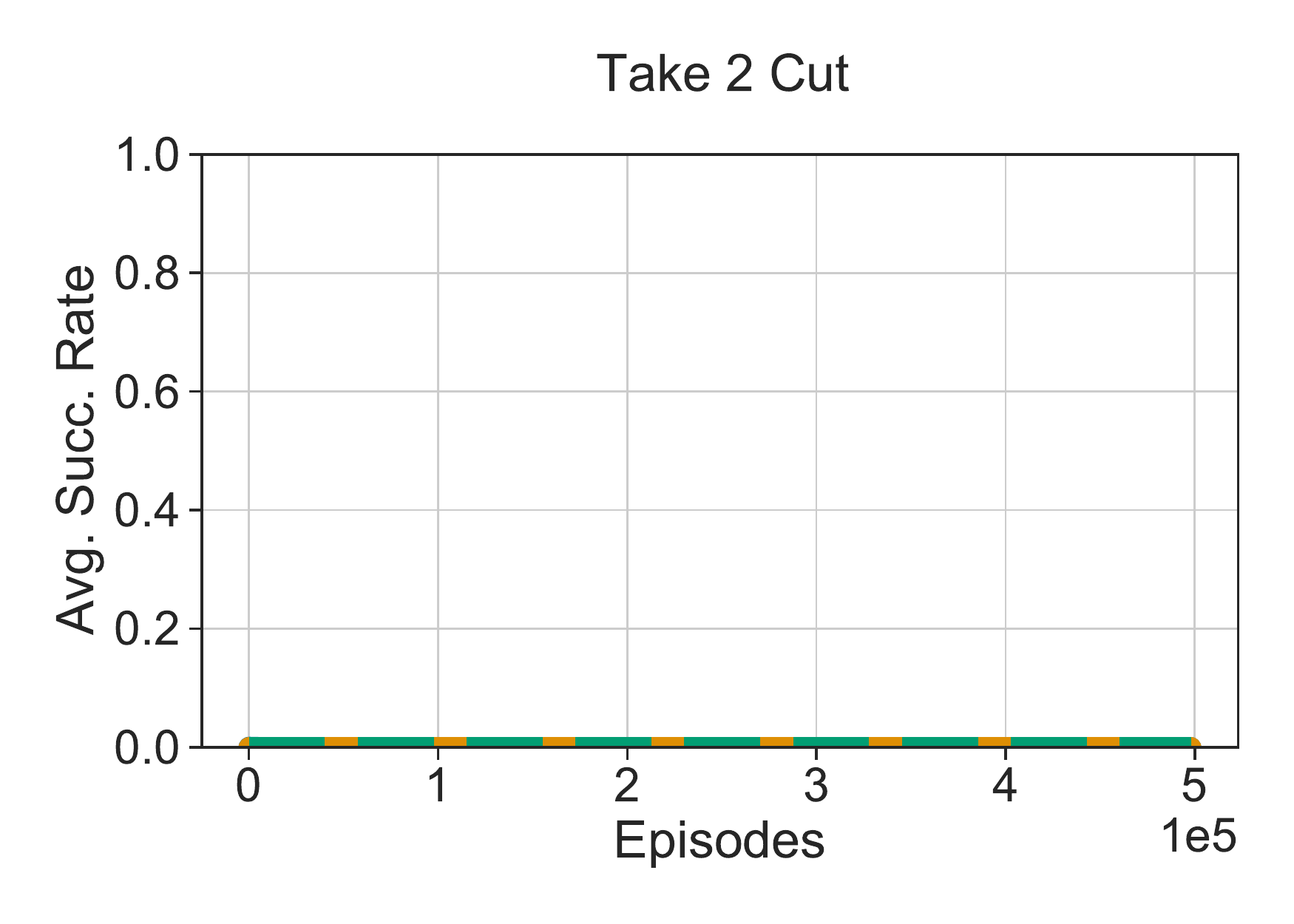}\\

\\
\multicolumn{2}{c}{\includegraphics[width=0.5\textwidth]{figures/legend.pdf}}

\end{tabular}

\caption{Success rate of \afk and baselines on \qtw.}
\label{fig:tw_score}

\end{figure*}

\begin{figure*}[t]
\centering

\begin{tabular}{cc}

\includegraphics[width=0.33\textwidth]{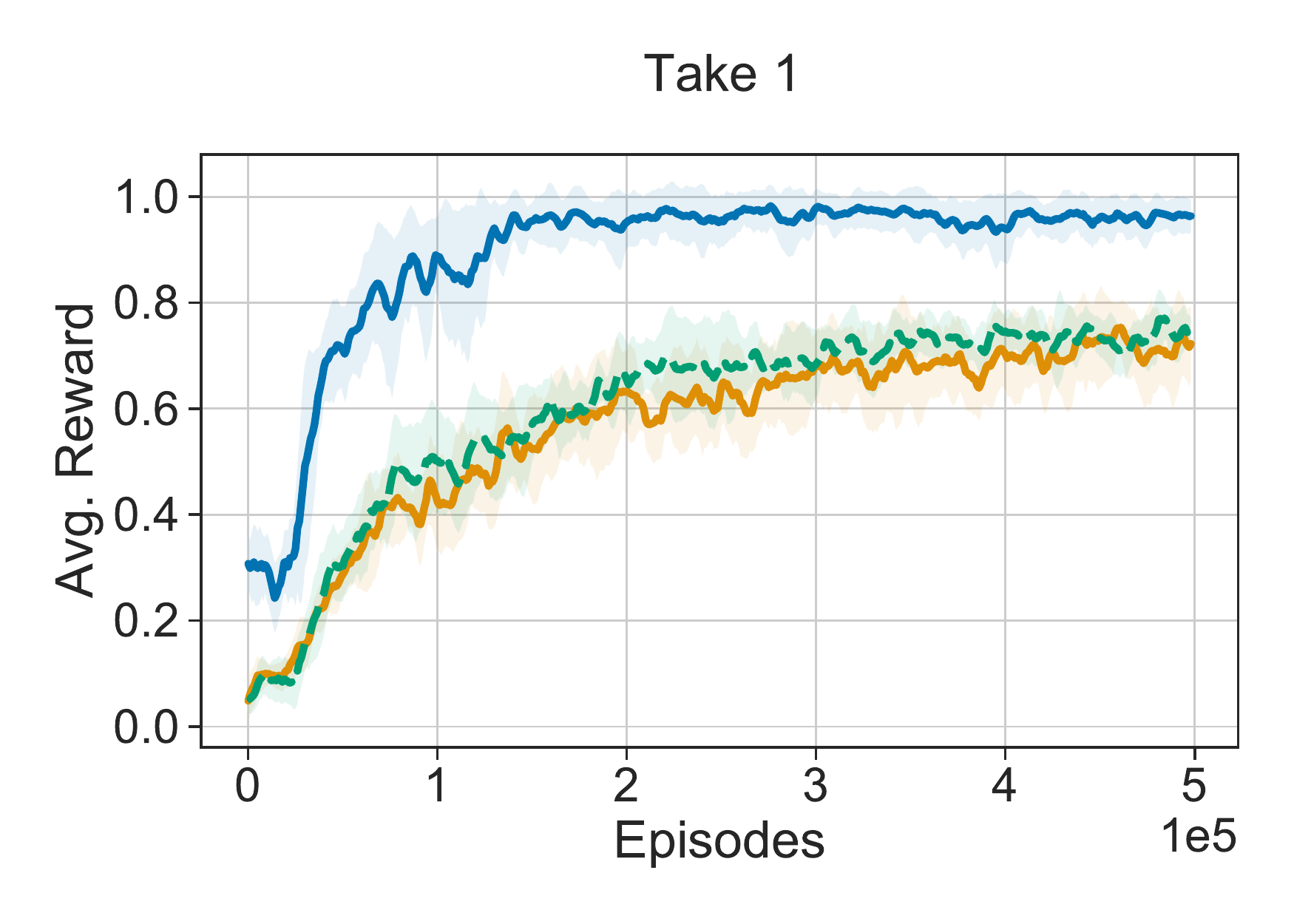}
&
\hspace{-0.6cm}\includegraphics[width=0.33\textwidth]{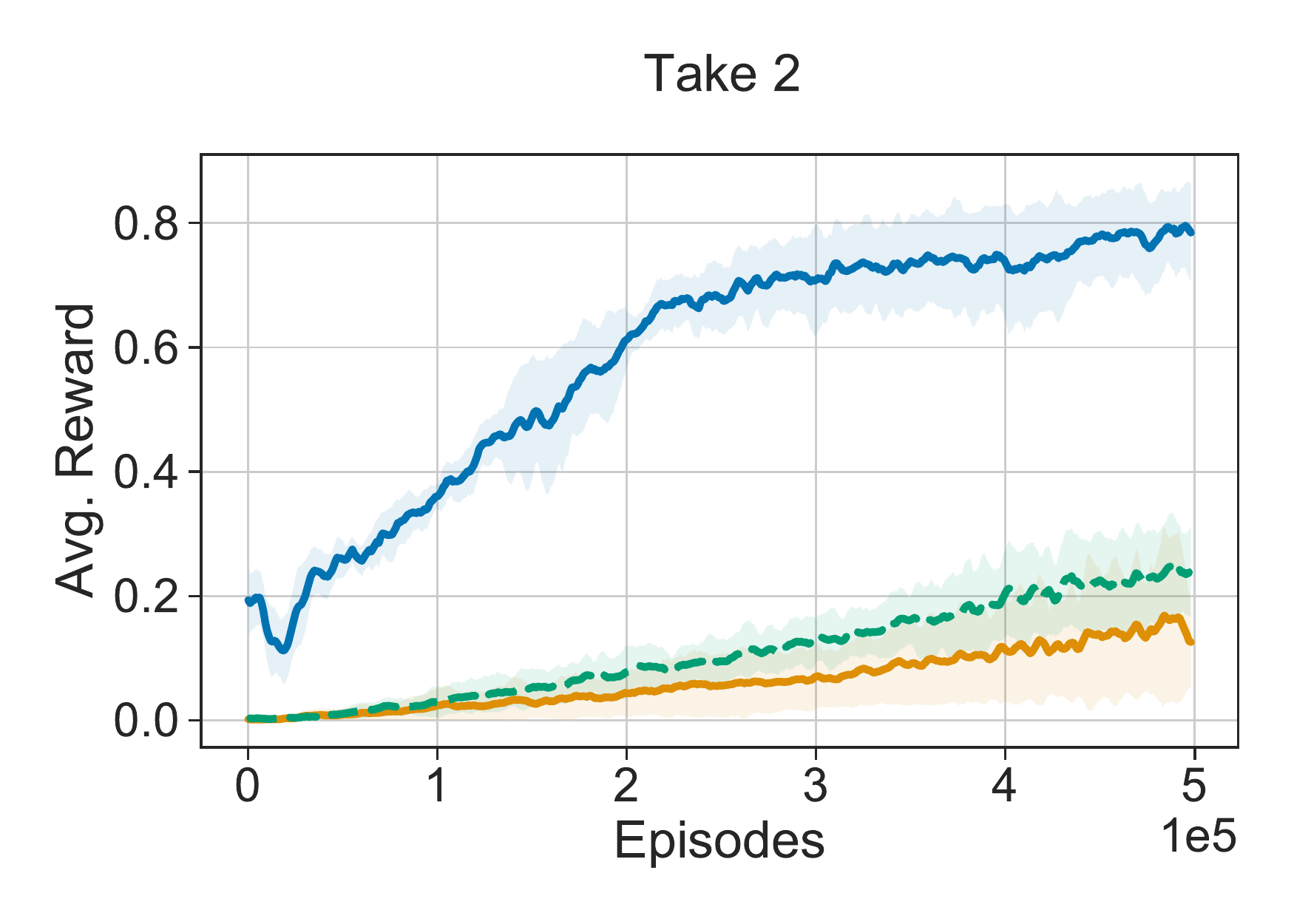}\\

\includegraphics[width=0.33\textwidth]{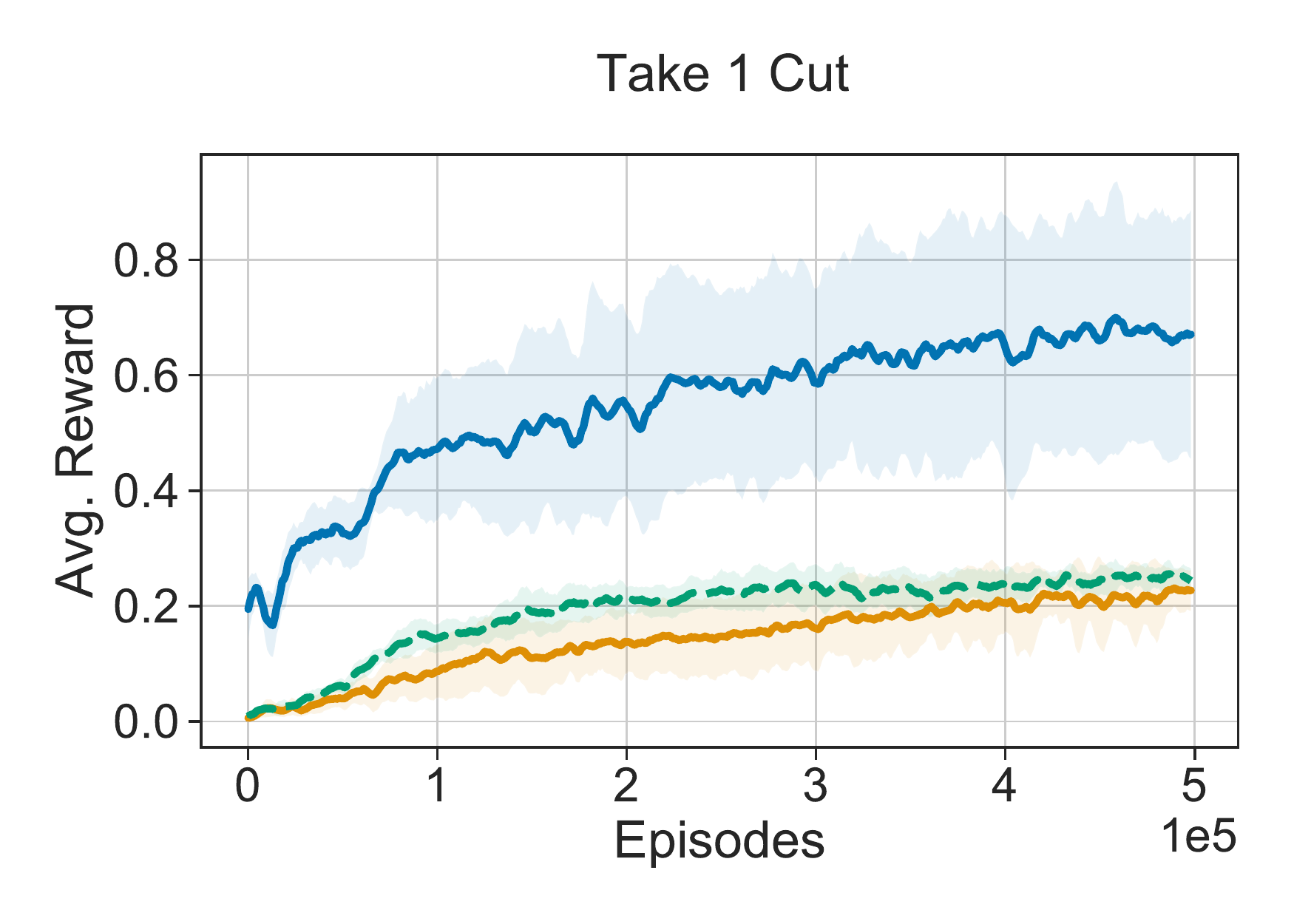}
&
\hspace{-0.6cm}\includegraphics[width=0.33\textwidth]{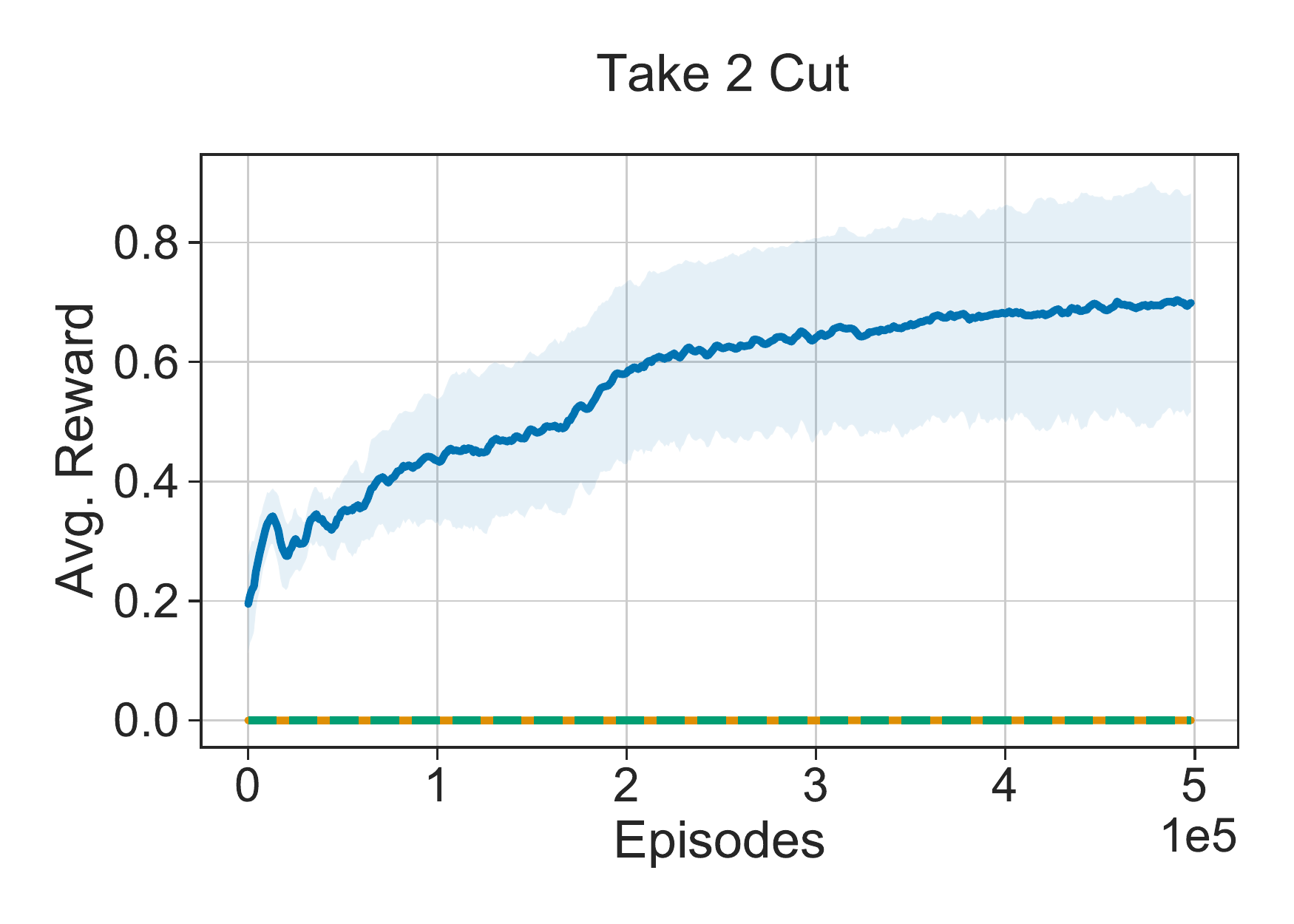}\\

\\
\multicolumn{2}{c}{\includegraphics[width=0.5\textwidth]{figures/legend.pdf}}

\end{tabular}

\caption{Training reward received by \afk and baselines on \qtw. Note, training reward is the sum of 1) reward given by the environment for solving the task; and 2) the episodic exploration bonus. }
\label{fig:tw_reward}

\end{figure*}

\begin{figure*}[t]
\centering

\begin{tabular}{cc}

\includegraphics[width=0.33\textwidth]{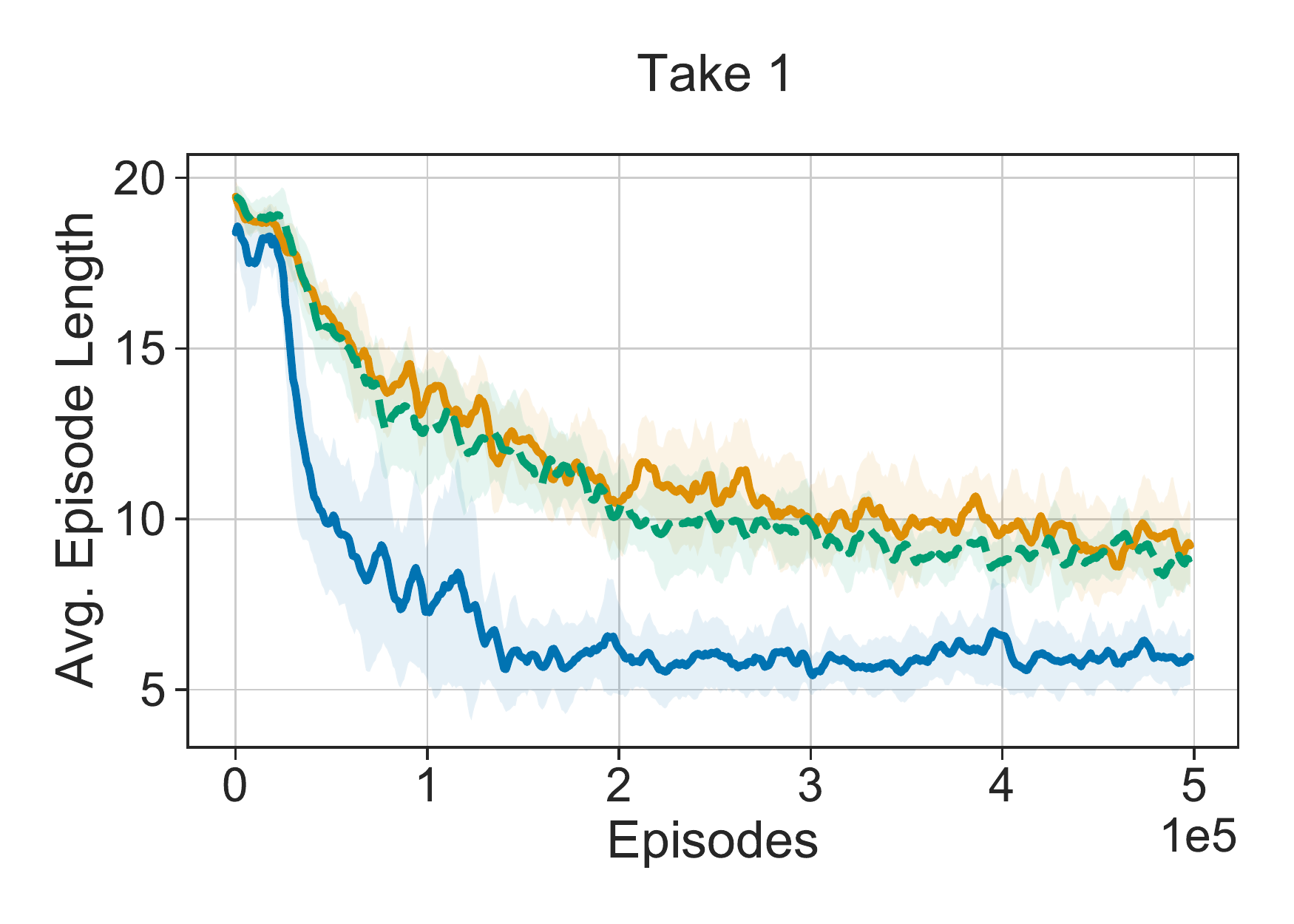}
&
\hspace{-0.6cm}\includegraphics[width=0.33\textwidth]{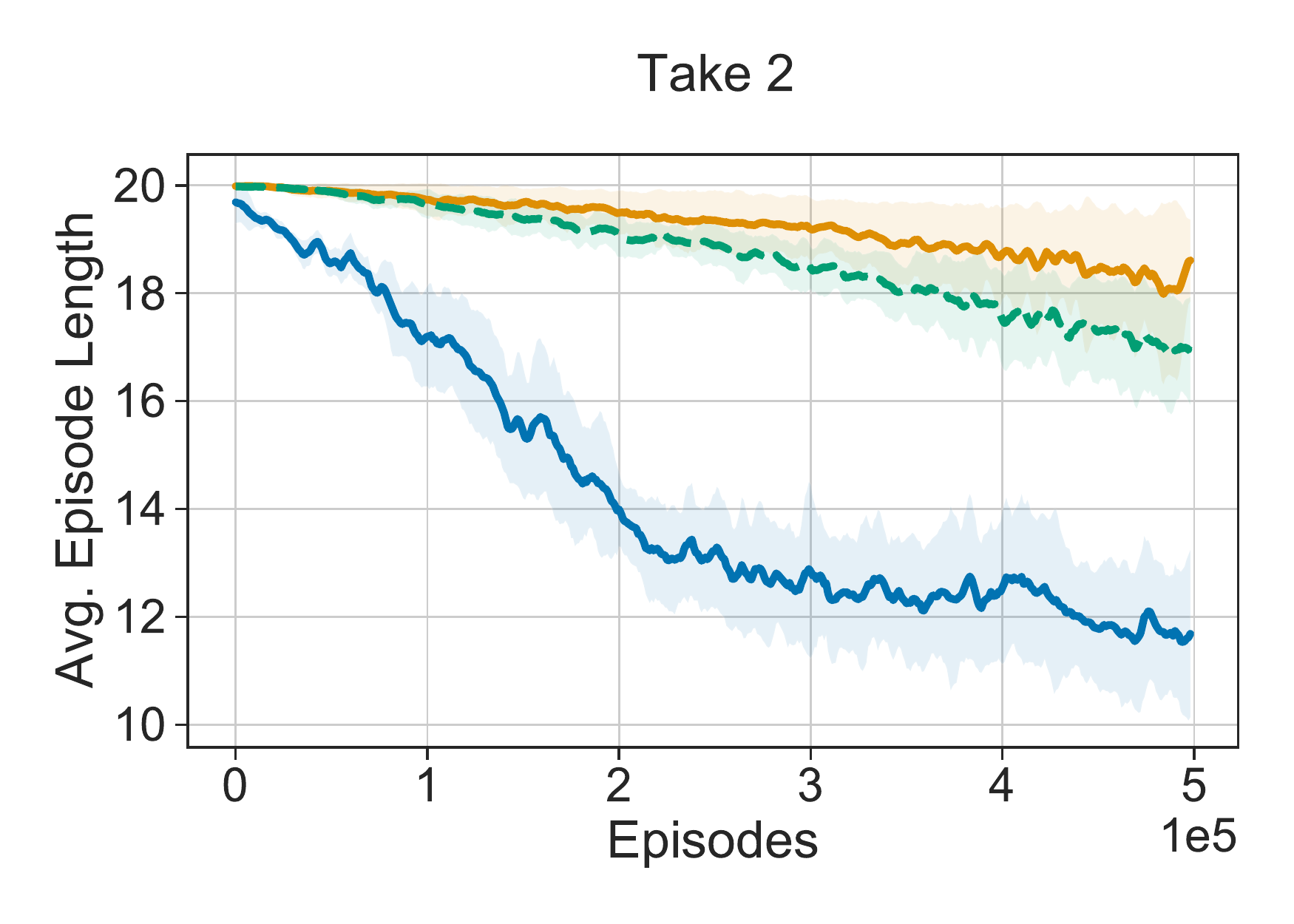}\\

\includegraphics[width=0.33\textwidth]{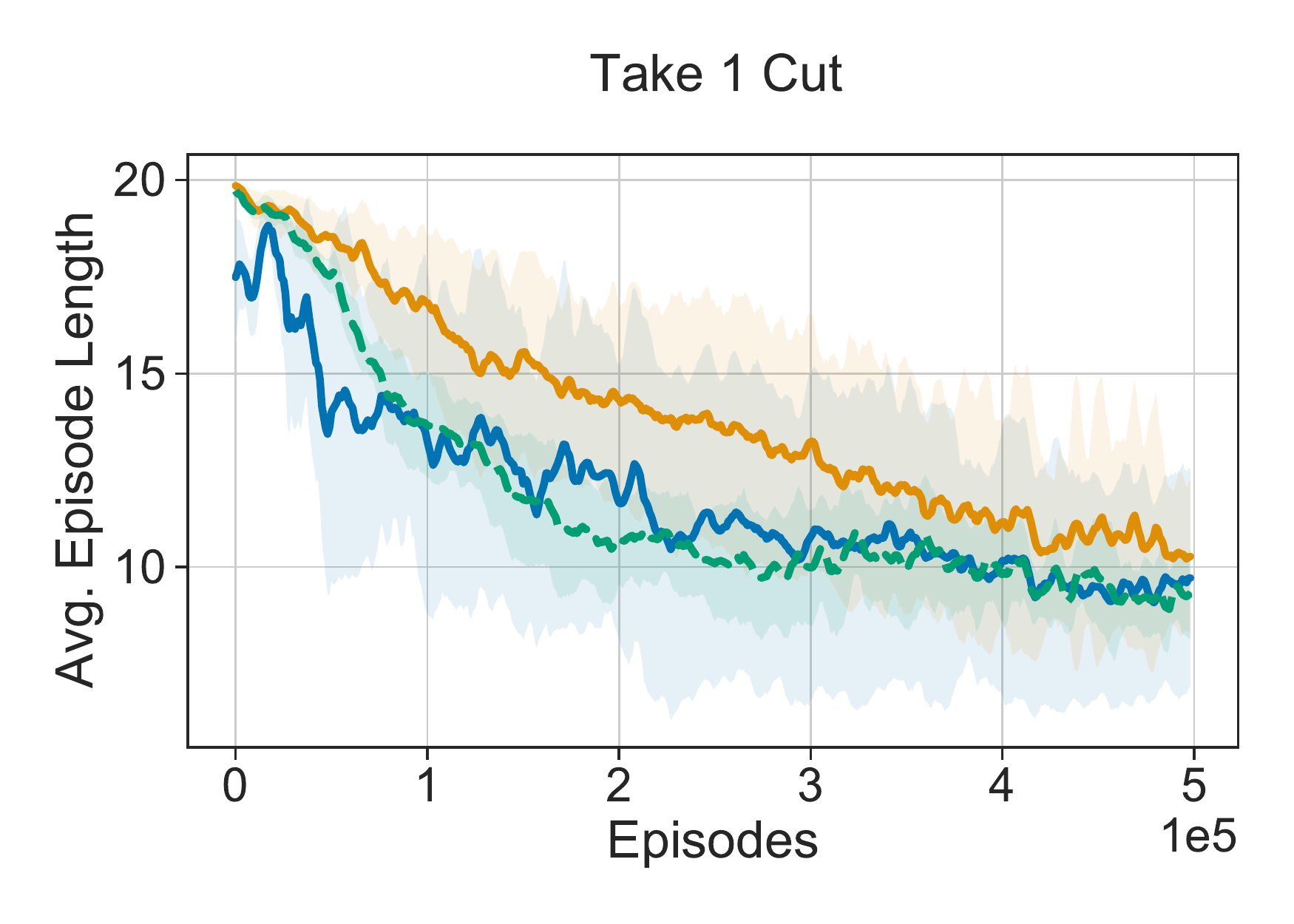}
&
\hspace{-0.6cm}\includegraphics[width=0.33\textwidth]{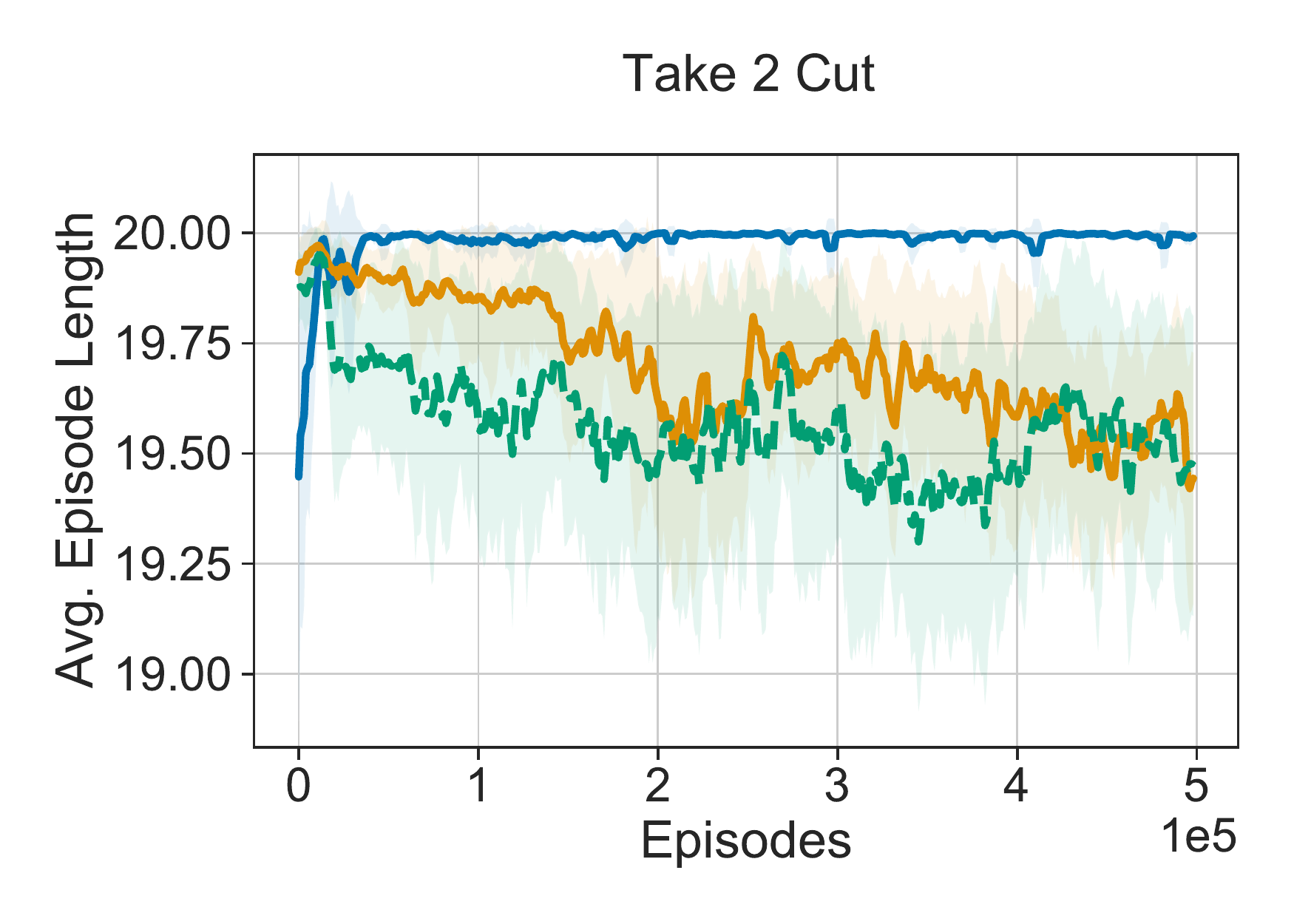}\\

\\
\multicolumn{2}{c}{\includegraphics[width=0.5\textwidth]{figures/legend.pdf}}

\end{tabular}
\caption{Steps used by AFK and baselines during training on \qtw.}
\vspace{-0.0cm}
\label{fig:tw_step}
\end{figure*}

The training curves of all \qbai and \qtw experiments in terms of success rate and episode length 
are shown in~\figref{fig:appd_plt_qbai_succ},~\figref{fig:appd_plt_qbai_len},~\figref{fig:tw_score},~\figref{fig:tw_reward}, and~\figref{fig:tw_step}.


\end{document}